\documentclass{article}
\usepackage{iclr2027_conference,times}
\iclrfinalcopy

\usepackage{helvet}
\usepackage{courier}
\usepackage[T1]{fontenc}
\usepackage[utf8]{inputenc}
\usepackage{microtype}
\usepackage{amsmath,amssymb,mathtools,bm}
\usepackage{booktabs,multirow,array}
\usepackage{graphicx,subcaption}
\usepackage[table]{xcolor}
\usepackage{tikz}
\usetikzlibrary{arrows.meta,positioning,calc}
\usepackage{url}
\usepackage{hyperref}
\usepackage{cleveref}
\usepackage{algorithm,algorithmic}
\usepackage{enumitem}

\let\cite\citep

\newcommand{\E}{\mathbb{E}}
\newcommand{\Jr}{J_R}
\newcommand{\Jc}{J_C}
\newcommand{\vlm}{\text{vlm}}

\newcommand{\rvlm}{r_{\vlm}}
\newcommand{\cvlm}{c_{\vlm}}
\newcommand{\meancvlm}{\overline{c}_{\vlm}}
\newcommand{\dlim}{d}
\newcommand{\eqnref}[1]{Eq.~\eqref{#1}}

\title{Vision--Language Signals in Constrained RL:\\
       Safety Gains Without Anticipation}

\author{Samuel Tetteh \& Cody Fleming \\
Iowa State University\\
Ames, Iowa, USA\\
\texttt{\{samtett, flemingc\}@iastate.edu}}

\begin{document}
\maketitle

\begin{abstract}
Safe reinforcement learning seeks policies that maximise task performance while
satisfying safety constraints. In driving benchmarks, however, collision costs
typically appear only at the time of collision, providing no advance warning of
an approaching hazard. Frozen vision--language models can provide dense semantic
feedback, yet it remains unclear whether their scores anticipate collisions and
which component drives an observed safety improvement. Episodic cost can also
favour policies that make little task progress. To address these gaps, we propose
VLM-Safe-RL, a framework that integrates frozen CLIP signals into PPO-Lagrangian
through reward shaping and an augmented multiplier update. On MetaDrive Hard,
which combines the densest traffic with the largest map, the catastrophe rate
falls from $31.6\%$ to $19.4\%$. FormulaOne-L2 analysis finds no evidence that
the CLIP signals anticipate collisions and shows that the VLM term has a
negligible effect on the Lagrange multiplier. These findings show a conditional
reduction in observed catastrophe rate without evidence of collision
anticipation.
\end{abstract}

\section{Introduction}
\label{sec:intro}

Reinforcement learning policies deployed in driving and other physical systems
must pursue their tasks without violating safety constraints. Constrained
reinforcement learning formalises this objective as maximising return subject to
a budget on expected cumulative cost~\cite{altman1999constrained}. Its
effectiveness, however, depends on the feedback used to represent unsafe
behaviour. In common driving benchmarks, collision cost becomes nonzero only
when a collision occurs. A vehicle approaching a barrier therefore receives no
direct indication of the developing hazard. A cost critic can propagate an
observed collision to preceding states, but the multiplier in
PPO-Lagrangian~\cite{ray2019benchmarking} is updated from episodic cost after the
trajectory has been collected.

\begin{figure}[t]
\centering
\resizebox{\textwidth}{!}{\input{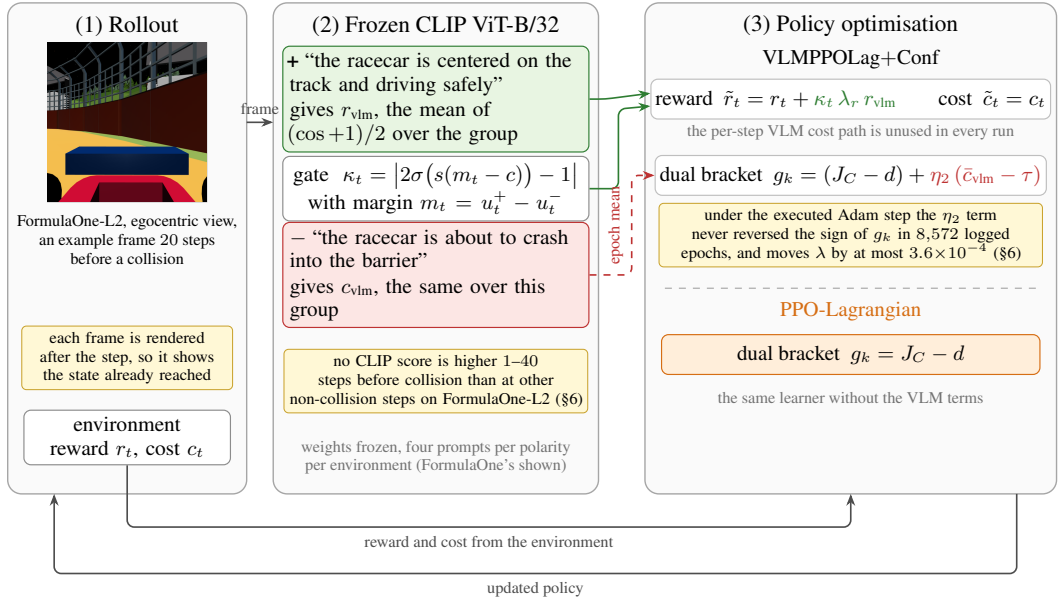}}
\caption{\textbf{VLM-Safe-RL and its measured components.} (1) The policy acts
from vector observations, and frozen CLIP scores each rendered frame against
positive and negative prompts on separate paths, every fourth step on MetaDrive
and Bullet. (2) The gated positive score, in green, adds a bonus to the reward.
The epoch mean negative score, in red and dashed, enters the multiplier update
through $\eta_2$, the route that could carry advance warning. (3) PPO-Lagrangian
omits both terms. Yellow notes give the measured effect of each component.}
\label{fig:pipeline}
\end{figure}

Rendered observations may reveal the relation between a vehicle and a barrier
before the simulator records a collision. Frozen vision--language models (VLMs)
such as CLIP~\cite{radford2021learning} can score an image against a description
such as ``the racecar is about to crash into the barrier'' without a
task-specific visual detector. This creates a possible source of dense safety
feedback, but anticipation requires two properties. The score must change before
the collision, and the learning rule must preserve this temporal information. An
epoch mean discards the order of observations even when an individual score is
informative.

Existing work either trains embodied vision--language models on robot data
\cite{brohan2023rt2,driess2023palme,zhang2025safevla} or uses frozen VLMs as
reward models~\cite{fan2022minedojo,rocamonde2024vlmrm,huang2024vlmrl}. Methods in
the latter group shape reward without constraining expected cost. SafeVLA
fine-tunes its model and does not isolate the effect of a frozen per-step signal.
Prior work therefore leaves open whether a frozen VLM signal anticipates
collisions and whether its timing is retained within a standard constrained
learner. Interpreting lower episodic cost presents a further challenge because a
policy can reduce cost by making less task progress. A sound assessment must
separate the information contained in the VLM signal, the update path through
which it acts, and the behaviour of the resulting policy.

To address these gaps, we propose VLM-Safe-RL, which integrates frozen CLIP
signals into PPO-Lagrangian through reward shaping and an augmented multiplier
update. As shown in \Cref{fig:pipeline}, positive and negative prompt groups
follow separate paths. A confidence-weighted positive score augments the task
reward, while the epoch mean negative score enters the multiplier update. The
environment cost remains unchanged. We evaluate final policies on three
MetaDrive settings that jointly vary traffic density and map size, and on
Bullet Safety-Gym. We then use FormulaOne-L2 to analyse the multiplier update,
replay its effect from recorded training statistics, test for a CLIP response
during the $1$ to $40$ steps preceding a collision, and relate episodic cost
to task progress.

The results show a conditional improvement. On MetaDrive Hard, which has the
densest traffic and largest map, VLM-Safe-RL reduces the catastrophe rate from
$31.6\%$ to $19.4\%$. We detect no improvement on Medium or Bullet, while
performance worsens on Easy. The MetaDrive experiments do not isolate
confidence gating from reward shaping, so the source of the Hard improvement
remains unresolved. On FormulaOne-L2, no CLIP signal rises before collisions
and the VLM term has a negligible effect on the multiplier update. None of the
evaluated policies completes the route, and lower cost largely reflects less
distance travelled. Together, these findings show a conditional reduction in
observed catastrophe rate without evidence of collision anticipation and
demonstrate why safety cost must be interpreted alongside task progress.

\section{Related Work}
\label{sec:related}

\noindent\textbf{Vision and language in robot learning.} Language-conditioned models
support task planning and affordance grounding
\cite{ahn2022saycan,huang2023voxposer}. PaLM-E~\cite{driess2023palme} integrates
visual and linguistic inputs, and RT-2~\cite{brohan2023rt2} maps them to robot
actions. SafeVLA~\cite{zhang2025safevla} applies constrained learning to a
fine-tuned vision--language--action policy. These approaches place a multimodal
model in the decision stack. We keep CLIP frozen and use its output as a training
signal for a separate control policy.

\noindent\textbf{Pretrained models as reward sources.} Language models can supply
proxy rewards~\cite{kwon2023reward} or generate reward
code~\cite{xie2024text2reward,ma2024eureka}. Frozen VLMs can instead score visual
observations. MineDojo~\cite{fan2022minedojo} uses a video--language reward in
Minecraft, and VLM-RM~\cite{rocamonde2024vlmrm} uses CLIP similarity as a
zero-shot reward.

\noindent\textbf{Pretrained models as safety signals.} VLM-SAFE~\cite{qu2025vlmsafe}
applies VLM guidance to futures generated by a world model for offline safe
driving. Other systems use VLMs to regulate a control barrier
function~\cite{chen2026alphaadjcbf} or detect visual
anomalies~\cite{jeong2023winclip}, and language feedback can update a robot's
safety representation online~\cite{santos2024updating}.

\noindent\textbf{Closest comparison.} VLM-RL~\cite{huang2024vlmrl} forms a
contrasting language goal (CLG) reward from the weighted difference between
positive and negative CLIP similarities, combines that score with vehicle-state
terms, and trains SAC without a CMDP constraint~\cite{haarnoja2018soft}. Our
PPO-CLG and CPO-CLG baselines reproduce its semantic CLG term, not its full reward
synthesis. VLM-Safe-RL instead keeps the positive and negative scores separate.
The positive score shapes reward, the epoch mean negative score enters the dual
update, and environment cost defines the constraint (App.~\ref{app:comparison},
\Cref{tab:comparison}).

\noindent\textbf{Constrained reinforcement learning.} These methods maximise return
under a budget on expected cost. Examples are CPO~\cite{achiam2017constrained},
PPO-Lagrangian~\cite{ray2019benchmarking}, FOCOPS~\cite{zhang2020first},
PCPO~\cite{yang2020projection}, and CUP~\cite{yang2022cup}.
PID-Lagrangian~\cite{stooke2020responsive} modifies the dual dynamics using
proportional, integral, and derivative terms in realised cost.
CRPO~\cite{xu2021crpo} alternates reward and constraint updates without a dual
variable, and Saut\'e RL~\cite{sootla2022saute} augments the state with a safety
budget. Our method retains PPO-Lagrangian and adds an epoch-mean VLM term to its
scalar dual (\S\ref{sec:results-eta2}).

\section{Method}
\label{sec:method}

\subsection{Problem formulation}
We consider an episodic constrained Markov decision process (CMDP)~\cite{altman1999constrained}
$\mathcal{M}=(\mathcal{S},\mathcal{A},P,r,c,\dlim,\gamma)$. Let $x$ denote the
vector observation supplied to the policy. A policy $\pi(a\mid x)$ induces
trajectories $\tau$. Let $R(\tau)$ and $C(\tau)$ denote
the cumulative task reward and environment cost along a trajectory. Their
expectations are $\Jr(\pi)=\E_{\tau\sim\pi}[R(\tau)]$ and
$\Jc(\pi)=\E_{\tau\sim\pi}[C(\tau)]$. We seek a policy with the highest expected
task return whose expected environment cost remains within budget $\dlim$. The
constrained problem is
\begin{equation}
  \pi^{\star}
  = \arg\max_{\pi}\Jr(\pi)
  \quad \text{subject to} \quad
  \Jc(\pi)\leq \dlim .
  \label{eq:cmdp}
\end{equation}
PPO-Lagrangian~\cite{ray2019benchmarking} learns separate value functions for
reward and cost. It uses a non-negative multiplier $\lambda$ to penalise cost
in the policy update.

At step $t$, the policy observes the environment vector $x_t$, takes action
$a_t$, and receives $x_{t+1}$, task reward $r_t$, and environment cost $c_t$.
On a VLM query step, we render an image $I_{t+1}$ for the frozen
vision--language model. The policy acts only on the vector observation and does
not receive the rendered image. Because scoring uses $I_{t+1}$, each VLM score
is assigned
to the post-transition state. The image may still contain evidence of an
upcoming violation, so whether the signal is anticipatory is an empirical
question that we examine in \S\ref{sec:results-eta2}.

\subsection{Frozen visual--language signals}
Let $\mathcal{P}^{+}=\{p_n^{+}\}_{n=1}^{N_+}$ describe desirable states and
$\mathcal{P}^{-}=\{p_n^{-}\}_{n=1}^{N_-}$ describe undesirable states. The
frozen CLIP image and text encoders are $f_I$ and $f_T$. We first average cosine
similarity within each prompt group,
\begin{equation}
  u_t^{\pm}
  = \frac{1}{N_{\pm}}\sum_{n=1}^{N_{\pm}}
  \operatorname{sim}\!\left(f_I(I_{t+1}),f_T(p_n^{\pm})\right),
  \qquad
  r_t^{\vlm}=\frac{u_t^{+}+1}{2},
  \quad
  c_t^{\vlm}=\frac{u_t^{-}+1}{2}.
  \label{eq:decoupled}
\end{equation}
The text embeddings are cached, and both scores lie in $[0,1]$. VLM-RL
\cite{huang2024vlmrl} combines positive- and negative-prompt similarities into
a single CLG reward, whereas we retain separate scores because they enter
different optimisation paths. The positive score $r_t^{\vlm}$ shapes the task
reward, and the epoch mean of the raw negative score $c_t^{\vlm}$ enters the
multiplier update. A coupled-softmax ablation instead normalises all prompt
logits jointly.

\subsection{Confidence gate and shaped reward}
We use a confidence gate $\kappa_t\in[0,1]$ to scale the VLM reward bonus. It is
computed from the similarity margin $m_t=u_t^{+}-u_t^{-}$. A positive margin
indicates greater similarity to positive prompts, and a negative margin indicates
greater similarity to negative prompts. We define the gate as
\begin{equation}
  \kappa_t
  = \left|2\sigma\!\left(s(m_t-c)\right)-1\right|
  \in[0,1].
  \label{eq:kappa-bayes}
\end{equation}
The absolute value makes $\kappa_t$ depend on the distance from $c$, independent
of which prompt group has the higher similarity. It is a confidence magnitude,
not a probability of safety. Values near zero suppress the reward bonus, and
values near one retain it. We set $\kappa_t=1$ when the gate is disabled. The
stored training signals are
\begin{equation}
  \widetilde r_t
  = r_t+\lambda_r\kappa_t r_t^{\vlm},
  \qquad
  \widetilde c_t=c_t .
  \label{eq:shaped-signals}
\end{equation}
Only the VLM reward bonus is gated. The environment cost remains unchanged, and
VLMPPOLag separately aggregates the raw $c_t^{\vlm}$ for the multiplier update.

The default gate uses $(s,c)=(100,0)$. The calibrated FormulaOne variant adapts
these parameters once from an unlabelled random-policy buffer
$\mathcal{B}=\{m_i\}_{i=1}^{B}$. We compute
\begin{equation}
  \widehat c=\operatorname{median}(\mathcal{B}),
  \qquad
  \widehat s
  =\frac{1}{\operatorname{IQR}(\mathcal{B})}
  \log\!\frac{1+\kappa^{\star}}{1-\kappa^{\star}} .
  \label{eq:mle-sc}
\end{equation}
With $B=500$ and $\kappa^{\star}=0.5$, the median margin maps to $\kappa=0$ and
a margin one interquartile range away maps to $\kappa=0.5$. This label-free rule
sets the gate scale without calibrating a probability.

\subsection{VLM-augmented multiplier update}
Standard PPO-Lagrangian adjusts $\lambda$ from mean episodic cost relative to
the budget. We also let the centred negative-prompt score affect the multiplier
without replacing environment cost. For the completed episodes
$\mathcal{E}_k$ in epoch $k$, we compute
\begin{equation}
  \widehat{\Jc}^{(k)}
  =\frac{1}{|\mathcal{E}_k|}
    \sum_{e\in\mathcal{E}_k}\sum_{t=1}^{T_e}c_{e,t},
  \qquad
  \meancvlm^{(k)}
  =\frac{1}{|\mathcal{E}_k|}
    \sum_{e\in\mathcal{E}_k}\frac{1}{T_e}
    \sum_{t=1}^{T_e}c_{e,t}^{\vlm}.
  \label{eq:epoch-statistics}
\end{equation}
$\widehat{\Jc}^{(k)}$ is the mean episodic environment cost, and
$\meancvlm^{(k)}$ is the episode-balanced mean of raw negative-prompt scores
computed before confidence gating or reward weighting.

We centre $\meancvlm^{(k)}$ at $\tau$ and add it to the environment-cost gap to
form $g_k$. One optimiser step then updates the multiplier as
\begin{equation}
  g_k
  =\big(\widehat{\Jc}^{(k)}-\dlim\big)
   +\eta_2\big(\meancvlm^{(k)}-\tau\big),
  \qquad
  L_{\lambda}^{(k)}=-\lambda g_k,
  \qquad
  \lambda_{k+1}
  =\Pi_{[0,\infty)}
   \!\left[
   \operatorname{Adam}_{\eta_1}
   \!\left(\lambda_k,\nabla_{\lambda}L_{\lambda}^{(k)}\right)
   \right].
  \label{eq:vlm-lagrange}
\end{equation}
Here $\tau$ is the VLM reference level, $\eta_2$ weights its contribution, and
$\eta_1$ is the multiplier learning rate. The VLM term raises $g_k$ when
$\meancvlm^{(k)}>\tau$ and lowers it when $\meancvlm^{(k)}<\tau$. Projection
enforces $\lambda\geq0$ without an upper bound. Setting $\eta_2=0$ recovers
standard PPO-Lagrangian.

\subsection{Training procedure}
\label{sec:training-procedure}
Each training epoch has three stages. First, the policy collects a rollout from
vector observations $x_t$. CLIP evaluates the post-transition image $I_{t+1}$
every $k_{\mathrm{clip}}$ steps. Between queries, we reuse the most
recent scores. At each episode boundary, the cache resets to $(0,0,1)$, causing
the first $k_{\mathrm{clip}}-1$ steps to use zero VLM scores when
$k_{\mathrm{clip}}>1$. Each transition is stored with shaped reward
$\widetilde r_t$ and unchanged environment cost $c_t$. The raw $c_t^{\vlm}$ is
logged separately. Second, we use the completed-episode statistics in
\eqnref{eq:epoch-statistics} to update $\lambda$ with \eqnref{eq:vlm-lagrange}.
Third, PPO updates the policy and the
reward and cost value functions.

\begin{algorithm}[H]
\caption{\textsc{VLMPPOLag}, one training epoch}
\label{alg:vlmppolag}
\footnotesize
\begin{algorithmic}[1]
  \STATE Reset $(r^{\vlm},c^{\vlm},\kappa)\leftarrow(0,0,1)$ at episode boundaries
  \FOR{each rollout step $t$}
    \STATE Sample $a_t\sim\pi_\theta(\cdot\mid x_t)$ and receive $(x_{t+1},r_t,c_t)$
    \STATE Refresh Eqs.~\eqref{eq:decoupled}--\eqref{eq:kappa-bayes} from $I_{t+1}$
      every $k_{\mathrm{clip}}$ steps, or reuse the cache
    \STATE Form and store \eqnref{eq:shaped-signals}, then log raw $c_t^{\vlm}$
  \ENDFOR
  \STATE Compute \eqnref{eq:epoch-statistics} and apply \eqnref{eq:vlm-lagrange}
  \STATE Run the base PPO update for $\pi_\theta$ and both value functions
\end{algorithmic}
\end{algorithm}

\section{Experimental Setup}
\label{sec:setup}

We ask three questions. Does gated VLM shaping reduce constraint violations and
catastrophic episodes? Does the epoch mean negative score change the multiplier?
Do policies with lower episodic cost still make task progress?

\paragraph{Environments.} We use three environment families. MetaDrive
\cite{li2023metadrive} has Easy, Medium and Hard settings, in which traffic
density rises from $0.1$ to $0.2$ and $0.35$ and map size from $3$ to $5$ and $7$
blocks. Each method is trained with $3$, $5$ and $10$ seeds on these settings,
using a pool of $10{,}000$ training scenarios. Bullet Safety-Gym
\cite{SafetyGym2019} is used only through \emph{SafetyCarReach-v0}, where a car
must reach a goal among eight static puddle hazards and one box obstacle that
moves along a circular path, in episodes of $500$ steps. Safety-Gymnasium
FormulaOne~\cite{ji2023safety} is our environment for component and lead-time
diagnostics. A race car follows a walled track defined by seven staged goals. L0
has no road barriers, while L1 and L2 each contain $200$, and L2 places barriers
roughly tenfold more often near checkpoints than elsewhere. Episodes last
$1{,}000$ steps. App.~\ref{app:envs} gives further details on the environments
and visual observations.

\paragraph{Methods compared.} We compare our full method, VLMPPOLag$+$Conf, with
PPOLag on MetaDrive and Bullet. VLMPPOLag applies the shaped reward with
$\kappa{=}1$ and the multiplier term of \eqnref{eq:vlm-lagrange}, and $+$Conf adds
the confidence gate. On FormulaOne we add VLM-free PPO, CPO and the unmatched
PPOLag of \S\ref{sec:results-main}
\cite{schulman2017proximal,achiam2017constrained,ray2019benchmarking}, together
with CLG reward-shaping baselines based on VLM-RL~\cite{huang2024vlmrl}. This
grid covers $11$ methods, three levels and seeds $\{42,123,456\}$, giving $99$
runs. Three seed- and configuration-matched pairs provide the main component
comparisons. CPO-Coupled and CPO-Decoupled compare coupled and separated VLM
scores. PPOLag-Decoupled ($\eta_2{=}0$) and VLMPPOLag compare updates without and
with the VLM multiplier term. VLMPPOLag and VLMPPOLag$+$Conf compare ungated and
gated variants. Additional L2 controls are CPPOPID~\cite{stooke2020responsive}
and a matched VLM-free PPOLag. Other cross-row comparisons are not treated as
ablations, because several use different optimisation settings. Two
independently trained calibrated families are analysed separately
(App.~\ref{app:f1-run-sets}).

\paragraph{Training.} All methods are trained with
OmniSafe~\cite{ji2023omnisafe} at $\dlim{=}25$, $\gamma{=}0.99$ and $20{,}000$
steps per epoch, for $10^{6}$ steps unless stated otherwise. The exception is
Bullet, whose six runs per method comprise three seeds at $10^{6}$ steps and
three at $2{\times}10^{6}$ steps (App.~\ref{app:results-bullet}). Prompts are
environment-specific. CLIP is queried every fourth step on MetaDrive and Bullet,
and on FormulaOne either every step or every fourth step depending on the run
set (Apps.~\ref{app:prompts} and~\ref{app:compute}).

\paragraph{Evaluation protocol.} All main evaluations use deterministic policy
actions. MetaDrive uses $50$ episodes per run on seeds $10000$--$10049$, giving
$150$, $250$ and $500$ episodes per method on Easy, Medium and Hard. Bullet uses
$200$ episodes per run, giving $1{,}200$ per method. FormulaOne policies are
evaluated on the plain L2 environment without the VLM wrapper. Environment-only
cost metrics use $50$ episodes per run on seeds $10000$--$10049$, and the first
$20$ enter the cross-environment comparison. Task metrics (stages completed, path
length and net progress) use $50$ episodes on seeds $30000$--$30049$.

\paragraph{Metrics and uncertainty.} We report the per-run mean environment-only
return $\Jr$ and cost $\Jc$, the violation rate $\Pr(C(\tau){>}\dlim)$ and the
catastrophe rate $\Pr(C(\tau){>}4\dlim)$, where $C(\tau)$ is the cost of one
episode. Rates pool episodes across runs. Intervals for per-run return and cost
resample episodes. Intervals for rate differences resample training seeds
independently within each arm, without resampling episodes, and take percentiles.
Both use $10{,}000$ bootstrap resamples~\cite{efron1986bootstrap}. Unless noted
otherwise, training results are final-epoch means $\pm$ population standard
deviations across seeds, and training cost is OmniSafe's mean over the last $100$
completed episodes. Training cost excludes the VLM score, whereas logged return
includes the shaping bonus, so task comparisons use environment-only return.
Concerns about small-sample RL evaluation
\cite{henderson2018matters,agarwal2021deep} motivate the seed-level intervals and
the exact one-sided permutation tests over seeds in
App.~\ref{app:perm-tests}.

\paragraph{Evaluation scope.} MetaDrive scenarios lie within the training pool,
so its results measure in-distribution performance. Three evaluation
limitations determine the protocol
(Apps.~\ref{app:metadrive-bug}--\ref{app:bullet-nondet}). First, every
evaluation seed is mapped into the training pool, so held-out scenarios are
not available. Second, MetaDrive can re-randomise traffic at every reset
independently of the scenario seed. We therefore fix traffic and create a
fresh simulator for each episode. Third, Bullet provides no functional
seeding mechanism. Its episodes are unpaired random draws, so we evaluate
$200$ episodes per run. FormulaOne samples the training reset distribution.
No policy completes a route stage, so its cost comparisons do not establish
task completion. The MetaDrive replication runs repeat the main gate
configuration with five seeds per level and are used only to estimate
training variability (App.~\ref{app:md-calib}).

\section{Results on MetaDrive and Bullet}
\label{sec:results}
\label{sec:results-gen}

The effect of gated CLIP shaping varies across environments. On MetaDrive Hard,
both catastrophe and violation rates decrease, and their seed-level intervals
exclude zero. Easy has higher catastrophe and violation rates, while Medium and
Bullet show no detectable change. Because the Hard comparison does not separate
confidence gating from reward shaping, we cannot determine which component drives
the improvement.

\Cref{tab:generalisation} compares the final checkpoints. MetaDrive evaluation is
in-distribution, and its three settings change map size together with traffic
density (\S\ref{sec:setup}). All runs retain the environment cost
$\widetilde c_t=c_t$, and the VLM term has a negligible measured effect on
$\lambda$ (\S\ref{sec:results-eta2}). The principal distinction between the two
arms is therefore the gated reward bonus $\lambda_r\kappa_t r_t^{\vlm}$, with
$\lambda_r=0.1$.

\begin{table}[t]
\centering
\caption{\textbf{Cross-environment evaluation.} MetaDrive uses $50$
deterministic episodes per run on seeds $10000$--$10049$, which map into the
training scenario pool. Bullet uses $200$ deterministic-policy episodes per run,
but its reset states cannot be seeded. Each method has $3$, $5$ and $10$
MetaDrive runs on Easy, Medium and Hard, and $6$ Bullet runs. Rates are pooled
over episodes. Catastrophe denotes $C(\tau){>}4\dlim$, and violation denotes
$C(\tau){>}\dlim$. The $95\%$ interval is a $10{,}000$-resample seed-level bootstrap
of (VLM$+$Conf)$-$PPOLag in percentage points. The F1-L2 row uses the first $20$
episodes from five calibrated VLM$+$Conf runs and two matched PPOLag runs. We do
not interpret that row as safety (\S\ref{sec:results-main}).}
\label{tab:generalisation}
\scriptsize
\setlength{\tabcolsep}{4pt}
\renewcommand{\arraystretch}{0.85}
\rowcolors{3}{gray!10}{white}
\begin{tabular}{@{}lccccc@{}}
\toprule
\textbf{Environment} &
  \multicolumn{2}{c}{\textbf{Cat.\,\%}} &
  \multicolumn{2}{c}{\textbf{Viol.\,\%}} &
  \textbf{Cat.~$\Delta$ 95\% CI} \\
\cmidrule(lr){2-3}\cmidrule(lr){4-5}
 & PPOLag & VLM$+$Conf & PPOLag & VLM$+$Conf & (pp) \\
\midrule
MetaDrive Easy   & 14.0 & 32.7 & 18.0 & 38.7 & $[+5.3,+35.3]$ \\
MetaDrive Medium & 26.0 & 28.0 & 32.8 & 36.0 & $[-8.8,+14.0]$ \\
MetaDrive Hard   & 31.6 & 19.4 & 39.2 & 26.0 & $[-21.8,-3.2]$ \\
Bullet Car-Reach & 13.2 & 12.7 & 20.8 & 19.8 & $[-3.1,+1.9]$ \\
\midrule
F1-L2$^\dagger$  & 2.5 & 8.0 & 17.5 & 18.0 & n/a \\
\bottomrule
\end{tabular}
\end{table}

On Hard, catastrophe falls from $31.6\%$ to $19.4\%$ ($-12.2$\,pp, $95\%$ CI
$[-21.8,-3.2]$), and violation falls from $39.2\%$ to $26.0\%$ ($-13.2$\,pp,
$95\%$ CI $[-24.4,-2.2]$). Mean environment-only return also improves from
$-653$ to $-468$. Nine of ten matched seeds have a lower catastrophe rate, eight
have a lower violation rate and seven have a lower mean cost
(App.~\ref{app:md-perseed}). However, $28.6\%$ of the VLM arm's episodes are
cost-free with return below $20$, compared with $5.2\%$ for PPOLag. The lower
cost may therefore partly reflect less active policies
(App.~\ref{app:md-fixed-eval}).

Replication variability limits the precision of this result. Two independently
trained sets with the same effective configuration differ in pooled catastrophe
rate by $7.5$, $6.8$ and $8.2$\,pp on Easy, Medium and Hard. On the four Hard
seeds shared by both sets, the difference is $11.5$\,pp. The observed Hard
reduction of $12.2$\,pp is of the same order, although its interval across ten
seeds excludes zero. We therefore treat the direction as detectable but the point
estimate as imprecise (App.~\ref{app:md-calib}).

The other MetaDrive settings do not support a general safety improvement. On
Medium, catastrophe changes from $26.0\%$ to $28.0\%$ ($+2.0$\,pp, $95\%$ CI
$[-8.8,+14.0]$), while mean return improves from $-794$ to $-543$. On Easy,
catastrophe increases from $14.0\%$ to $32.7\%$ ($+18.7$\,pp, $95\%$ CI
$[+5.3,+35.3]$), and mean return decreases from $-307$ to $-845$. All three
Easy seeds have a lower catastrophe rate under PPOLag. We have not identified the cause of this increase.
Separate frame diagnostics show that positive-prompt scores exceed
negative-prompt scores throughout the sampled Easy videos
(App.~\ref{app:gate-calibration}).

The ordering across MetaDrive settings is descriptive. The catastrophe-rate
difference changes from $+18.7$ to $+2.0$ and $-12.2$\,pp as traffic density
increases. Yet map size, collision opportunities and baseline failure rates also
change. The gate is nearly non-selective in separate Hard diagnostics, where its
median weight is $0.90$--$0.98$ (App.~\ref{app:gate-diagnosis}). Without a
controlled density study, we cannot attribute the ordering or the Hard result to
visual advance warning or gate selectivity.

Bullet shows no detectable difference. Catastrophe is $13.2\%$ for PPOLag and
$12.7\%$ for VLMPPOLag$+$Conf, while violation is $20.8\%$ and $19.8\%$. The
catastrophe-rate difference has a seed-level interval of $[-3.1,+1.9]$\,pp.
App.~\ref{app:cr-extras} reports additional FormulaOne substitutions based on RND
and Qwen2-VL-7B, neither of which is a matched control.

\section{Multiplier and Lead-Time Analysis}
\label{sec:results-eta2}
\label{sec:results-aux}

Neither the multiplier term in \eqnref{eq:vlm-lagrange} nor the per-step CLIP
scores provide evidence of anticipation. The multiplier term has a negligible
effect on $\lambda$, and none of the CLIP scores rises before contact on
FormulaOne-L2.

\paragraph{The multiplier term shifts the effective cost budget.} At an interior
stationary point of \eqnref{eq:vlm-lagrange},
$(\Jc^{*}-\dlim)+\eta_2(\meancvlm-\tau)=0$, which gives
$\Jc^{*}=\dlim+\eta_2(\tau-\meancvlm)$. When the signal's epoch mean is affine in
environment cost, this is the original constraint with a shifted budget. The
epoch average also removes the within-epoch ordering that would encode advance
warning. This limitation follows from the aggregation and applies even to a
perfect indicator of the constraint event. It does not depend on CLIP quality.
With $\eta_2{=}0.01$, $\tau{=}0.5$, and the logged mean
$\meancvlm{=}0.624$ across the three-seed ungated and gated VLMPPOLag runs on
FormulaOne-L2, the effective budget tightens by only
$1.2{\times}10^{-3}$ cost units. The shifts for the other CLIP, Qwen2-VL, and RND
runs are similarly small (App.~\ref{app:rnd-details}).

\paragraph{The implemented update changes $\lambda$ negligibly.} Adam normalises
the scalar gradient by its running second moment, so the VLM term can materially
change the first update only if it changes the sign of $g_k$. Across $8{,}572$
logged epochs from $189$ runs with $\eta_2{>}0$, the VLM term never reverses a
nonzero sign. Its magnitude never exceeds $5{\times}10^{-3}$, while
$|\Jc-\dlim|$ has a median of $15.2$. Removing the term and replaying the logged
updates changes $\lambda$ by at most $3.6{\times}10^{-4}$ in any run
(App.~\ref{app:omnisafe}). The MetaDrive Hard improvement in
\Cref{tab:generalisation} therefore cannot be attributed to this term.

\begin{figure}[t]
  \centering
  \includegraphics[width=0.6\linewidth]{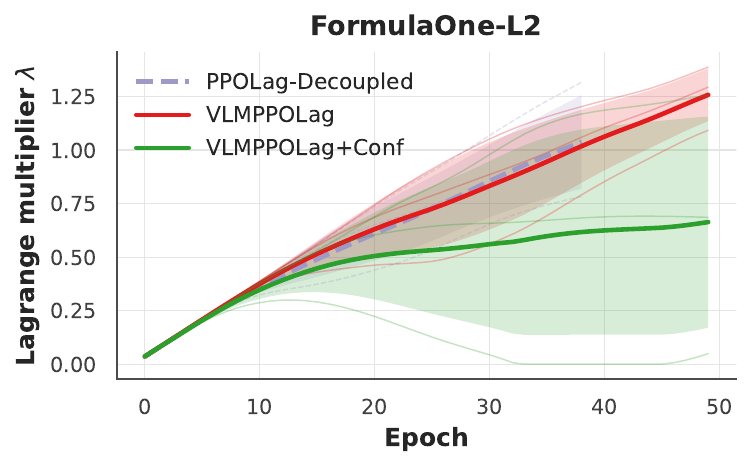}
  \caption{\textbf{Lagrange multiplier on FormulaOne-L2.} The figure compares
    VLMPPOLag with $\eta_2{=}0.01$ (red), its matched $\eta_2{=}0$ ablation
    (purple, dashed), and its matched confidence-gated variant (green). All arms
    use seeds $\{42,123,456\}$ and score every frame. Bold lines show seed means,
    bands show $\pm1$ population standard deviation, and thin lines show
    individual seeds. The purple mean ends at epoch $39$, the last epoch reached
    by its seed-$42$ run.}
  \label{fig:lambda-ablation}
\end{figure}

\paragraph{The multiplier traces and matched ablations agree.} In
\Cref{fig:lambda-ablation}, the mean $\lambda$ traces for VLMPPOLag and
PPOLag-Decoupled differ by at most $0.029$ through the $39$ epochs reached by all
six runs. The direction of the per-seed difference is not consistent
(App.~\ref{app:omnisafe}). Removing the multiplier term changes final-epoch cost
from $40.2$ to $40.7$ in the matched comparison of
\Cref{tab:component-diagnostics}, with an exact one-sided permutation
$p{=}0.50$. A second matched comparison using $k_{\mathrm{clip}}{=}4$ gives costs
of $41.4$ with $\eta_2{=}0.03$ and $27.5$ with $\eta_2{=}0$, opposite the proposed
reduction ($p{=}0.90$). With three seeds per arm, the smallest attainable
one-sided $p$ is $0.05$, so neither comparison can establish an effect
(Apps.~\ref{app:compute} and~\ref{app:perm-tests}).

The gated arm finishes with a lower mean $\lambda$ than the ungated arm
($0.66$ against $1.26$), despite the same all-epoch mean $\meancvlm$ of $0.624$.
Because $\kappa$ changes the shaped reward but does not enter the multiplier
update directly, this difference arises through the environment costs induced by
the learned policies (\Cref{tab:component-diagnostics}).

\begin{table}[t]
\centering
\caption{\textbf{FormulaOne-L2 component diagnostics.} Panel A reports
final-epoch training cost over seeds $\{42,123,456\}$ with every frame scored.
The first comparison isolates confidence gating, and the second isolates the
$\eta_2$ term. Panel B reports the pooled lead-time AUC range over two arms and
$K\in\{1,3,5,10,20,40\}$. An AUC above $0.5$ means that a signal is higher before
contact. Full definitions and intervals appear in
Apps.~\ref{app:f1-ablation} and~\ref{app:horizon-roc}.}
\label{tab:component-diagnostics}
\scriptsize
\setlength{\tabcolsep}{7pt}
\textbf{(A) Matched component comparisons}\\[1pt]
\begin{tabular}{@{}lcc@{}}
\toprule
\textbf{Configuration} & $\Jc$ & \textbf{Seeds over budget} \\
\midrule
VLMPPOLag$+$Conf & 30.5 & 1/3 \\
VLMPPOLag (ungated) & 40.2 & 2/3 \\
PPOLag-Decoupled ($\eta_2{=}0$) & 40.7 & 3/3 \\
\bottomrule
\end{tabular}
\par\smallskip
\textbf{(B) Lead-time AUC}\\[1pt]
\begin{tabular}{@{}lcc@{}}
\toprule
\textbf{Signal} & \textbf{Raw} & \textbf{Within-episode standardised} \\
\midrule
$\kappa$ & 0.172--0.320 & 0.371--0.497 \\
$\cvlm$  & 0.199--0.452 & 0.348--0.462 \\
$\rvlm$  & 0.149--0.398 & 0.322--0.465 \\
Margin   & 0.173--0.320 & 0.356--0.493 \\
Step index & \multicolumn{2}{c}{0.512--0.614} \\
\bottomrule
\end{tabular}
\end{table}

\paragraph{CLIP scores do not rise before contact.} We first assess whether the
gate identifies ongoing contact. When fitted and evaluated on the same $25{,}000$
frames per level, $\kappa$ reaches an AUC of $0.82$ on FormulaOne-L1 and $0.78$
on L2 for current-step cost. These in-sample results show an association with
ongoing contact, not advance warning (App.~\ref{app:gate-roc}).

We then test for advance warning on FormulaOne-L2 using the unmatched VLM-free
PPOLag baseline and the calibrated $+$Conf policy. Each arm contributes $180$
deterministic episodes over seeds $\{42,123,456\}$. A non-contact step is
positive when contact occurs within the next
$K\in\{1,3,5,10,20,40\}$ steps. Across both arms, all $96$ pooled AUCs for
$\kappa$, $\cvlm$, $\rvlm$, and the CLIP margin lie between $0.149$ and $0.497$
for their raw and within-episode standardised values. The step-index control
reaches $0.512$--$0.614$ (\Cref{tab:component-diagnostics}, panel B). Every CLIP
score is therefore lower on average before contact, and none provides the
expected advance warning. The rank correlation between $\cvlm(t)$ and cost at
$t{+}k$ is also nearly constant for $k\in[-10,+20]$. It ranges from $-0.068$ to
$-0.066$ for $+$Conf and from $-0.359$ to $-0.349$ for PPOLag. These results show
no lead-time signal on FormulaOne-L2. We do not make the same claim for
MetaDrive, where lead time was not measured (App.~\ref{app:horizon-roc}).

\section{FormulaOne Cost Without Task Completion}
\label{sec:results-main}

\paragraph{Cost tracks activity on an unsolved route.} None of the $1{,}000$
evaluation episodes from $20$ runs completes any of the seven route stages
(\Cref{tab:f1-task}). The reported arm statistics use $19$ runs because one
PPOLag-Decoupled run saved only its initial checkpoint. Across the six arm means,
cost has Pearson correlations of $+0.978$ with path length and $+0.991$ with net
progress. PPO and the unmatched PPOLag account for much of this relationship.
Among the other four arms, path length spans only $1.91$--$2.31$ distance units
while mean cost varies by a factor of $2.5$. Lower FormulaOne cost therefore
largely reflects less movement on an unsolved route and does not establish safer
control.

\paragraph{Training return includes the shaping bonus.} Logged episodic return
cannot be compared directly across arms because VLM-shaped arms add
$\lambda_r\sum_t\rvlm$ to the environment reward. At the final epoch,
CPO-Coupled and CPO-Decoupled log returns of $21.6$ and $63.9$ at similar costs
of $32.4$ and $30.9$. After removing the bonus, their environment returns differ
by less than $0.1$ at the final epoch and over the last $5$ and $10$ epochs
(\Cref{tab:f1-return-decomp}). The calibrated five-run
VLMPPOLag$+$Conf arm logs a last-10-epoch return of
$31.8$ but obtains an environment-only evaluation return of $0.06$
(\Cref{tab:main_results}, panel~B). We therefore use environment-only return for
task comparisons. Environment cost already excludes the VLM scores in every
run.

\paragraph{Baseline choice reverses the apparent result.} On the first $20$
evaluation episodes per run, the calibrated five-run VLMPPOLag$+$Conf arm has
catastrophe and violation rates of
$8.0\%$ and $18.0\%$. The matched PPOLag records $2.5\%$ and $17.5\%$ over two
seeds, making the catastrophe rate of VLMPPOLag$+$Conf $5.5$ percentage points
higher. The unmatched PPOLag records $18.3\%$ and $26.7\%$ over three seeds on
the same evaluation scenarios, making the VLMPPOLag$+$Conf catastrophe rate
$10.3$ percentage points lower. Mean cost over $50$ episodes per run reverses in
the same way (Apps.~\ref{app:f1-run-sets} and~\ref{app:f1-main-tables}). The
matched control is closer in configuration but has only two runs and no
task-metric rollout. The unmatched baseline differs in update iterations,
initial multiplier, and training seeds. We therefore regard the FormulaOne-L2
comparison as uninformative about safety.

\paragraph{The matched component comparisons are also inconclusive.} Panel A of
\Cref{tab:component-diagnostics} reports two single-factor removals. Removing the
confidence gate raises final-epoch training cost from $30.5$ to $40.2$ and the
number of seeds over budget from $1/3$ to $2/3$. The gate nevertheless raises
cost on seed~$123$, and the exact one-sided permutation test gives $p{=}0.30$ at
the final epoch and $p{=}0.10$ over the last $10$ epochs. With only three seeds,
this comparison cannot establish a gating effect. Environment-only evaluation
of the same checkpoints gives mean costs of $36.8$ with the gate and $39.6$
without it. Removing the multiplier term changes final-epoch cost from $40.2$ to
$40.7$ and the number of seeds over budget from $2/3$ to $3/3$, which provides
no evidence for that term (\S\ref{sec:results-eta2}). Because no policy completes
the route and cost is strongly associated with movement, neither comparison
establishes a safety improvement.

\section{Discussion and Limitations}
\label{sec:discussion}

Our results support a narrow positive claim. Gated shaping reduces catastrophe
on MetaDrive Hard, but this setting combines the densest traffic with the
largest map. The experiments do not isolate confidence gating from reward
shaping, and a less active policy may explain part of the reduction. The
evidence also does not support anticipation. The multiplier receives the VLM
signal as an episode-balanced mean over the episodes completed in each
$20{,}000$-step epoch, so it cannot respond to an individual cut-in. At an
interior stationary point, the term shifts the effective cost budget when the
signal mean is affine in environment cost (\S\ref{sec:results-eta2}). The
$\kappa$-weighted bonus acts at each step, but no CLIP score rises before
collision on FormulaOne-L2. Lead time was not measured on MetaDrive
(\S\ref{sec:results-aux}).

Our scope is limited to three simulated benchmarks and frozen visual-language
models. MetaDrive evaluation is in-distribution and does not measure transfer.
Easy produces an adverse result that cannot be declared significant by the
exact seed-level test. Medium and Bullet show no detectable change. FormulaOne-L2
is unsolved and cost largely tracks distance, so it serves only as a diagnostic.
Reward shaping is not isolated, and the RND and Qwen2-VL-7B experiments are
unmatched substitutions. Most comparisons use three to five seeds. Prompts are
hand-written, and one multiplier is shared by both signals. A complete inventory
appears in App.~\ref{app:limitations}.

\section{Conclusion}
\label{sec:conclusion}
Gated CLIP shaping lowers catastrophe on MetaDrive Hard by $12.2$ percentage
points and raises it on Easy, using evaluation scenarios drawn from the
training pool. These settings differ in both traffic density and map size, and
the mechanism behind the Hard reduction remains unresolved. The method does
not make constraint satisfaction anticipatory. No CLIP score rises before
collision on FormulaOne-L2, and the multiplier term has negligible measured
influence. Cost on this unsolved route largely tracks distance travelled.
Evaluation limitations further restrict the conclusions that can be drawn.




\bibliographystyle{iclr2027_conference}
\bibliography{references}

\clearpage
\appendix

\section{Implementation Details}
\label{app:implementation}

\subsection{Hyperparameters}
\label{app:hparams}

\Cref{tab:hparams} reports the settings used in the evaluated runs. We list all
differences across methods and run sets, including the number of PPO updates.

\begin{table}[p]
\centering
\footnotesize
\renewcommand{\arraystretch}{0.90}
\caption{Hyperparameters for the reported FormulaOne, Bullet and MetaDrive
runs. Values that differ across methods or run sets are listed explicitly.}
\label{tab:hparams}
\begin{tabular}{@{}l>{\raggedright\arraybackslash}p{0.30\linewidth}>{\raggedright\arraybackslash}p{0.50\linewidth}@{}}
\toprule
\textbf{Group} & \textbf{Hyperparameter} & \textbf{Value} \\
\midrule
\multirow{12}{*}{Optimisation}
   & Total timesteps & $10^{6}$ (FormulaOne, MetaDrive, Bullet), $2{\times}10^{6}$ (Bullet long-horizon runs) \\
   & Steps per epoch & $20{,}000$ ($50$ epochs for $10^{6}$-step runs, $100$ epochs for the Bullet $2{\times}10^{6}$-step runs) \\
   & Discount $\gamma$ & $0.99$ \\
   & GAE $\lambda_{\text{GAE}}$ & $0.95$ \\
   & Learning rate (actor / critic) & $3{\times}10^{-4}$ / $3{\times}10^{-4}$, CPO-family runs have no actor learning rate and a $10^{-3}$ critic learning rate \\
   & Linear learning-rate decay & on \\
   & Mini-batch size / update iterations & $64$ / $40$ for PPO-family VLM runs, PPO-CLG and CPPOPID, $64$ / $10$ for the remaining PPO-family baselines, and $128$ / $10$ for CPO-family runs \\
   & Target KL & $0.02$ with KL early stopping, $0.01$ for CUP (with early stopping) and for CPO-family runs (without) \\
   & Gradient-norm clip & $40$ \\
   & Observation normalisation & on \\
   & Reward / cost normalisation & off / off \\
   & Advantage standardisation & reward and cost advantages both standardised \\
\midrule
\multirow{5}{*}{CMDP}
   & Cost limit $\dlim$ & $25$, $15$ and $35$ in the Pareto-anchor PPOLag-Decoupled runs (App.~\ref{app:pareto-baseline}) \\
   & PPO clip ratio $\epsilon$ & $0.2$ \\
   & Multiplier update & Adam at learning rate $\eta_1{=}0.035$ for PPOLag, VLMPPOLag, FOCOPS and CUP ($\eta_1$ is an Adam learning rate, not an additive step size), $\eta_1{=}0.07$ in the corresponding cells of the $\eta_1{\times}\eta_2$ sweep on one multiplier (\Cref{tab:two-mult}), CPPOPID uses a PID update with gains $(K_p,K_i,K_d){=}(0.1,0.01,0.01)$ \\
   & Multiplier bounds & clipped at $0$ from below, no upper bound, except $2.0$ for FOCOPS and CUP and a penalty cap of $100$ for CPPOPID \\
   & Initial multiplier $\lambda_0$ & $0.001$ for PPOLag, VLMPPOLag and CPPOPID in the matched training pipeline, and $0.0$ for the remaining PPOLag, FOCOPS and CUP baselines. PPO, PPO-CLG, CPO-family methods and P3O do not use a Lagrange multiplier \\
\midrule
\multirow{7}{*}{VLM}
   & Scorer & CLIP ViT-B/32 (frozen, $151$M parameters), ViT-L/14 in App.~\ref{app:clip-capacity}, Qwen2-VL-7B in App.~\ref{app:qwen-details} \\
   & VLM reward weight $\lambda_r$ & $0.1$ ($0$ for PPOLag-RND) \\
   & VLM cost weight $\lambda_c$ & $0.5$, inactive in every reported run \\
   & VLM term weight $\eta_2$ (inside the bracket of \eqnref{eq:vlm-lagrange}) & $0.01$, $0$ in the $\eta_2{=}0$ arm, $0.03$ in cells of \Cref{tab:two-mult} \\
   & Danger threshold $\tau$ & $0.5$ \\
   & Prompts $K, L$ & $4, 4$ (v1 prompt sets) in every CLIP run except the prompt-template runs, which use $1, 1$ (v2) and $2, 2$ (v3) (App.~\ref{app:prompt-templates}) \\
   & CLIP period $k_{\text{clip}}$ & per run set (App.~\ref{app:compute}) \\
\midrule
\multirow{3}{*}{Evaluation}
   & Evaluation seeds & $10000$ to $10000{+}N_{\text{ep}}{-}1$, $30000$ to $30049$ for the FormulaOne task metrics (App.~\ref{app:task-metrics}) \\
   & Episodes per run $N_{\text{ep}}$ & $10$ (FormulaOne L1 and L2 screening evaluation), $20$ deterministic and $20$ stochastic on seeds $10000$--$10019$ (calibrated-gate run sets and their FormulaOne-L2 controls, App.~\ref{app:phaseB-robustness}) and $20$ (Bullet, \Cref{tab:bullet-perseed}), $50$ (MetaDrive reproducible protocol, including App.~\ref{app:md-fixed-eval} and the MetaDrive prompt-template runs, FormulaOne-L2 environment-only evaluation), $200$ (Bullet) \\
   & Bootstrap resamples (95\% CIs) & $10{,}000$ episode resamples for per-run return and cost intervals, and $10{,}000$ seed-level cluster resamples for the catastrophe-rate intervals in \Cref{tab:generalisation}. App.~\ref{app:stats} describes the remaining analyses \\
\bottomrule
\end{tabular}
\end{table}

\subsection{Network architecture}
\label{app:arch}

The actor and the reward and cost critics are separate multilayer perceptrons
with two hidden layers of $64$ units and tanh activations. The actor defines a
Gaussian distribution over the two-dimensional continuous action, with mean
$\mu_\theta(x)$ and a learned state-independent log standard deviation. The
critics $V_\phi^R(x)$ and $V_\phi^C(x)$ estimate reward and cost returns for the
corresponding GAE advantages. Policy observations have $28$ dimensions on
FormulaOne-L0, $44$ on FormulaOne-L1 and L2, $259$ on MetaDrive, and $57$ on
Bullet. The CLIP ViT-B/32 encoder remains frozen. Its $K+L$ text embeddings are
cached at initialisation, so each queried frame requires one image encoding and
$O(K+L)$ dot products.

\subsection{Multiplier optimisation}
\label{app:omnisafe}

VLMPPOLag uses OmniSafe v0.5 and replaces the standard PPOLag multiplier
optimisation with \eqnref{eq:vlm-lagrange}. At epoch $k$, it forms $g_k$,
minimises $-\lambda g_k$ with Adam at learning rate $\eta_1$, and clips
$\lambda$ at zero. The raw, ungated $\cvlm$ determines $\meancvlm$. Neither the
confidence gate $\kappa_t$ nor the inactive cost weight $\lambda_c$ enters this
update. The multiplier is updated before the policy, while the PPO objective,
value losses, and clipping remain unchanged.

Two controls assess the contribution of $\eta_2$. PPOLag-Decoupled matches
VLMPPOLag but omits the VLM term. Its seed-$42$ run ends at epoch $39$. A
separate VLMPPOLag arm sets $\eta_2=0$, but uses $k_{\mathrm{clip}}=4$ while the
corresponding $\eta_2>0$ arm uses $k_{\mathrm{clip}}=1$. This difference in
sampling period prevents the comparison from isolating $\eta_2$.

\paragraph{Multiplier logs.}
The first Adam step is $\eta_1\operatorname{sign}(g_k)$ because the optimiser
normalises the scalar gradient of $-\lambda g_k$. MetaDrive-Hard seeds $42$ and
$123$ have first-epoch violations of $45.45$ and $150.40$, yet both move from
$\lambda_0=0.001$ to $0.036$. Later steps depend on Adam's running moments.
Across runs with $\eta_2>0$, the per-epoch change remains below $0.046$ for
$\eta_1=0.035$ and below $0.082$ for $\eta_1=0.07$.

Across $8{,}572$ epochs from $189$ training runs with $\eta_2>0$, the largest
VLM term $|\eta_2(\meancvlm-\tau)|$ is $4.9\times10^{-3}$. It never reverses the
sign of a nonzero constraint-violation term. Four epochs have $\Jc=\dlim$, so
the VLM term determines the sign of $g_k$ in those cases. The resulting update
still depends on Adam's accumulated moments.

We replay the Adam update using the recorded episodic cost and $\meancvlm$. The
replayed multiplier agrees with the recorded value within $7\times10^{-7}$.
Removing the VLM term changes $\lambda$ by at most $3.6\times10^{-4}$ within any
run. The median maximum change is $3\times10^{-5}$. These values support the
negligible multiplier effect reported in \S\ref{sec:results-eta2}.

\subsection{Computational requirements}
\label{app:compute}

\Cref{tab:compute} reports wall-clock training time for completed runs. Each run
uses one GPU and one process on a shared cluster with mixed GPU models, so the
values are indicative. Evaluation time is excluded. Relative to matched
VLM-free PPOLag, rendering and CLIP scoring increase training time by $6.5\%$ on
FormulaOne-L2 when CLIP is queried every fourth step and by $20$--$22\%$ when it
is queried every step. The increase is $24$--$34\%$ on MetaDrive. On Bullet,
where VLM-free training does not render frames, VLM training takes approximately
$4.5$ times as long. A single ViT-B/32 image encoding with eight cached prompt
embeddings takes $7.11$ ms on an A100 and $9.11$ ms on a V100, averaged over
$500$ passes after $50$ warm-up passes.

Recorded training time across all runs totals $4{,}222$ GPU-hours. Complete
Completed runs account for $3{,}949$ hours, interrupted or restarted full runs for $265$
hours, and pilot runs for $8$ hours. This total excludes evaluation and
analysis, so it is a lower bound on the computational cost of the study.

\begin{table}[h]
\centering
\small
\setlength{\tabcolsep}{4pt}
\caption{Training time per completed run. We report the mean and range of total
wall-clock time, mean rollout time accumulated across epochs, and mean
environment steps per second. The final column compares each method with the
VLM-free PPOLag reference in the same environment. FormulaOne-L2 includes
PPOLag configurations with $10$ and $40$ update iterations. PPOLag-Dec. denotes
PPOLag-Decoupled, excluding its incomplete seed-$42$ run. The two $+$Conf rows
use the prior-symmetric and calibrated gates. A CLIP period of ``none'' denotes
runs without frame scoring.}
\label{tab:compute}
\begin{tabular}{@{}llcccccc@{}}
\toprule
\textbf{Env.} & \textbf{Method} & $k_{\text{clip}}$ & \textbf{Runs} & \textbf{Time (h)} & \textbf{Rollout (h)} & \textbf{Steps/s} & \textbf{vs.\ ref.} \\
\midrule
\multirow{7}{*}{FormulaOne-L2}
 & PPOLag, 10 iter. & none & 3 & 18.6 [18.4, 18.7] & 18.4 & 14.9 & $-3\%$ \\
 & PPOLag, 40 iter. & none & 2 & 19.2 [19.2, 19.2] & 18.2 & 14.5 & ref. \\
 & CPPOPID & none & 3 & 19.2 [19.1, 19.2] & 18.2 & 14.5 & $0\%$ \\
 & PPOLag-Dec. & 1 & 2 & 23.1 [22.9, 23.2] & 22.2 & 12.0 & $+20\%$ \\
 & VLMPPOLag & 1 & 3 & 23.4 [23.0, 24.0] & 22.5 & 11.9 & $+22\%$ \\
 & $+$Conf, prior & 1 & 3 & 23.3 [23.1, 23.5] & 22.3 & 11.9 & $+21\%$ \\
 & $+$Conf, calib. & 4 & 5 & 20.5 [20.2, 20.7] & 19.5 & 13.6 & $+6.5\%$ \\
\midrule
\multirow{2}{*}{Bullet, $10^{6}$}
 & PPOLag & none & 3 & 1.7 [1.7, 1.7] & 0.7 & 162 & ref. \\
 & $+$Conf & 4 & 3 & 7.7 [7.6, 7.8] & 6.6 & 36 & ${\times}4.5$ \\
\midrule
\multirow{2}{*}{Bullet, $2{\times}10^{6}$}
 & PPOLag & none & 3 & 3.3 [3.2, 3.4] & 1.4 & 169 & ref. \\
 & $+$Conf & 4 & 3 & 15.3 [15.0, 15.5] & 13.4 & 36 & ${\times}4.6$ \\
\midrule
\multirow{2}{*}{MetaDrive Easy}
 & PPOLag & none & 3 & 2.1 [2.1, 2.2] & 1.1 & 131 & ref. \\
 & $+$Conf & 4 & 3 & 2.8 [2.7, 2.8] & 1.8 & 101 & $+30\%$ \\
\midrule
\multirow{2}{*}{MetaDrive Medium}
 & PPOLag & none & 5 & 2.9 [2.7, 3.1] & 2.0 & 96 & ref. \\
 & $+$Conf & 4 & 5 & 3.9 [3.7, 4.0] & 3.1 & 72 & $+34\%$ \\
\midrule
\multirow{2}{*}{MetaDrive Hard}
 & PPOLag & none & 10 & 4.7 [4.3, 5.1] & 3.8 & 60 & ref. \\
 & $+$Conf & 4 & 10 & 5.8 [5.6, 6.1] & 4.9 & 48 & $+24\%$ \\
\bottomrule
\end{tabular}
\end{table}

\paragraph{CLIP period.}
When $k_{\mathrm{clip}}>1$, CLIP is queried once every $k_{\mathrm{clip}}$
control steps and the most recent $(\rvlm,\cvlm,\kappa)$ values are reused
between queries. The cache resets to $(0,0,1)$ at the start of each episode.
FormulaOne run sets use $k_{\mathrm{clip}}=1$ or $4$, MetaDrive and Bullet use
$k_{\mathrm{clip}}=4$, and Qwen2-VL uses $k_{\mathrm{clip}}=8$. VLM-free methods
score no frames. The main FormulaOne table and the $\eta_2$, learning-rate,
gate, Pareto-anchor, and Qwen2-VL comparisons contain arms with different CLIP
periods. We therefore do not attribute differences in those comparisons solely
to the named component.

\paragraph{Cluster allocation.}
Each full training job requests one GPU, eight CPU cores, and $32$ GB of memory.
Qwen2-VL and ViT-L/14 jobs request $48$ GB. Job limits extend to $4.5$ days, and
completed training runs take between $1.6$ and $32.5$ hours.

\section{Environments}
\label{app:envs}

\subsection{Safety-Gymnasium FormulaOne}
\label{app:env-f1}

Safety-Gymnasium FormulaOne~\cite{ji2023safety} uses a Racecar agent with
continuous steering and throttle controls. The agent follows a walled track with
seven sequential goals. Reward increases with progress towards the current goal,
and reaching it gives a bonus of $10$. An episode lasts $1{,}000$ steps. Each
step represents $0.04$ seconds, giving a control frequency of $25$ Hz and a
maximum duration of $40$ seconds. The policy receives a $28$-dimensional vector
on L0 and a $44$-dimensional vector on L1 and L2. CLIP receives the rendered RGB
observation.

The levels differ in obstacle placement and collision cost. L0 has no free
obstacles, and wall collisions incur no cost. L1 places $200$ road barriers
across two track regions. Collisions with a barrier or the static track geometry
incur a cost of $1$. L2 retains $200$ barriers and samples their locations
more frequently in regions around the route goals, increasing local density.
None of the evaluated policies completes a route stage on L2, and episodic cost
increases with distance travelled (\S\ref{sec:results-main},
App.~\ref{app:task-metrics}).

\subsection{Bullet Safety-Gym SafetyCarReach-v0}
\label{app:env-bullet}

SafetyCarReach-v0 from Bullet Safety-Gym~\cite{SafetyGym2019} uses
PyBullet~\cite{coumans2016pybullet}. A car-like agent must reach a goal in a
room containing eight static circular puddle hazards and one box that moves on a
circular path. Episodes last $500$ steps. Contact with a puddle or the box incurs
a per-step cost of $1$, and the cost limit is $\dlim=25$. The policy receives a
$57$-dimensional vector. CLIP receives the environment's $480\times360$ RGB
rendering, which is resized and centre-cropped to $224\times224$.

We train three seeds for $10^6$ steps and three for $2\times10^6$ steps, giving
six runs per method. Each run is evaluated for $200$ episodes. The environment
does not reproduce its initial state from a requested seed, so the evaluation
episodes are unpaired random draws (App.~\ref{app:bullet-nondet}).
\Cref{tab:bullet-perseed} also reports a $20$-episode evaluation of the same
policies.

\subsection{MetaDrive}
\label{app:env-md}

MetaDrive~\cite{li2023metadrive} is a procedural driving simulator. The Easy,
Medium and Hard settings use $3$, $5$ and $7$ map blocks with traffic densities
of $0.1$, $0.2$ and $0.35$. Episodes last at most $1{,}000$ steps. A vehicle
collision or departure from the road does not terminate an episode. The policy
receives a $259$-dimensional state vector and outputs continuous steering and
acceleration. CLIP receives a $256\times256$ top-down observation, resized to
$224\times224$.

All runs train for $10^6$ steps. Easy uses seeds $\{42,123,456\}$, Medium adds
$\{789,2024\}$, and Hard further adds $\{1337,2718,3141,4242,5555\}$. Each run
is evaluated deterministically for $50$ episodes on indices
$10000$--$10049$. These indices map into the training scenario pool, so the
results measure in-distribution performance (App.~\ref{app:metadrive-bug}).

\paragraph{Evaluation scenarios overlap with training.}
\label{app:metadrive-bug}
The evaluation procedure maps each requested index to its remainder modulo the
scenario-pool size. With a pool of $10{,}000$, evaluation index $10000+i$ maps
to scenario $i$. The evaluation set therefore lies within the training pool.
The specific scenarios sampled during training were not recorded, so the actual
overlap cannot be measured. All reported MetaDrive results should consequently
be interpreted as in-distribution estimates.

\begin{figure}[h]
  \centering
  \includegraphics[width=0.98\linewidth]{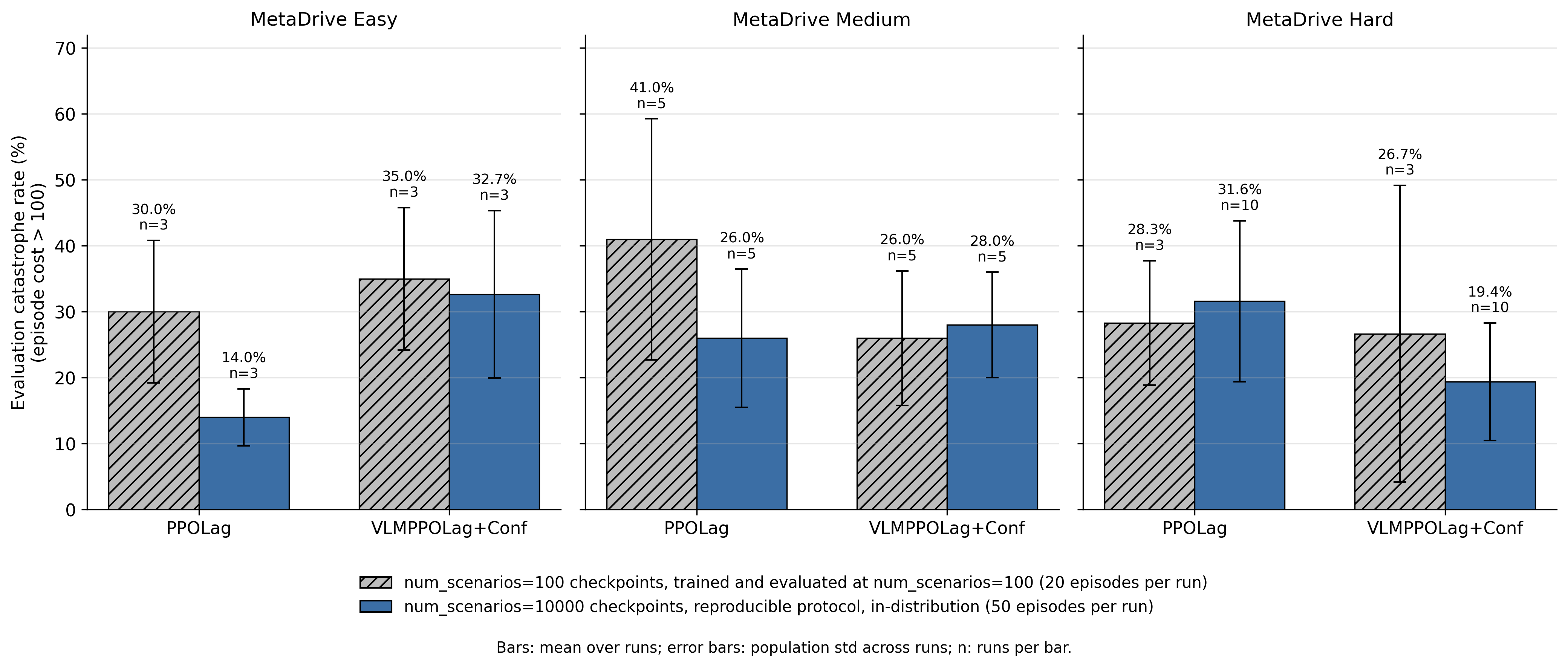}
  \caption{MetaDrive catastrophe rates for deterministic PPOLag and
  VLMPPOLag$+$Conf policies. The earlier evaluation uses $20$ episodes per run,
  a shared simulator, randomised traffic and a scenario pool of $100$. The final
  protocol uses $50$ episodes per run, fixed traffic, a new simulator for each
  episode and a pool of $10{,}000$. Both settings evaluate scenarios from their
  training pool. Bars show mean per-run rates with population standard
  deviations. Easy and Medium use $3$ and $5$ runs per method in both settings.
  Hard uses $3$ runs in the earlier setting and $10$ in the final protocol.}
  \label{fig:metadrive-bug}
\end{figure}

\Cref{fig:metadrive-bug} compares an earlier evaluation with the final protocol.
Under the earlier setting, VLMPPOLag$+$Conf reduces catastrophe rate on Medium
from $41.0\%$ to $26.0\%$ and on Hard from $28.3\%$ to $26.7\%$, while Easy
increases from $30.0\%$ to $35.0\%$. Under the final protocol, the Medium
reduction disappears, Hard decreases from $31.6\%$ to $19.4\%$, and Easy
increases from $14.0\%$ to $32.7\%$. The comparisons also differ in trained
policies, episode counts and Hard seed counts. The figure therefore illustrates
sensitivity to the evaluation design but does not isolate a single cause.

\paragraph{Traffic randomisation.}
\label{app:metadrive-nondet}
Traffic is sampled independently of the scenario index at each reset. Repeated
evaluations can therefore share the same map but contain different vehicles. A
simulator reused across episodes also retains state, making an episode depend on
those evaluated before it. The final protocol fixes traffic and creates a new
simulator for every episode. Repeating five episodes under this protocol gives
identical returns, costs and episode lengths. Training retains random traffic to
preserve traffic diversity.

This protocol makes evaluation reproducible but does not remove training
variation. Two independently trained MetaDrive-Hard groups with the same
configuration differ by $11.5$ percentage points in pooled catastrophe rate over
the four shared seeds (App.~\ref{app:md-calib}).

\paragraph{Bullet evaluations are not seed matched.}
\label{app:bullet-nondet}
Bullet Safety-Gym does not reproduce the same initial observation when supplied
with the same requested seed. Its evaluation episodes are therefore unpaired
random draws and cannot be matched across methods. We use $200$ episodes per
run to reduce sampling uncertainty.

\paragraph{Reporting implications.}
The scenario overlap limits MetaDrive conclusions to in-distribution
performance. Traffic randomisation and persistent simulator state can change an
evaluation result, while Bullet episodes cannot be paired by seed. We therefore
report the number of training runs and evaluation episodes for each comparison.
The catastrophe-rate intervals in \Cref{tab:generalisation} use a seed-level
cluster bootstrap, so their uncertainty reflects independent training runs.

\section{VLM Prompts}
\label{app:prompts}

\subsection{Prompt sets}
\label{app:prompt-list}

The default template uses four positive and four negative prompts for each
environment. Their mean rescaled cosine similarities define $\rvlm$ and $\cvlm$.
The environment cost remains unchanged. The epoch mean of $\cvlm$ enters the
multiplier through $\eta_2$, while both scores determine the confidence margin.

\Cref{tab:prompt-sets} lists the default prompts. Each negative set includes a
forward-looking description of an approaching collision. On FormulaOne-L2,
however, none of the CLIP signals distinguishes pre-collision frames from other
non-contact frames at horizons of $1$ to $40$ steps
(App.~\ref{app:horizon-roc}). Qwen2-VL replaces the prompt lists with the group
descriptions ``safe driving conditions'' and ``driving danger or imminent
collision'' (App.~\ref{app:qwen-details}).
\begin{table}[h]
\centering
\small
\caption{Default prompt sets used for CLIP scoring.}
\label{tab:prompt-sets}
\begin{tabular}{@{}>{\raggedright\arraybackslash}p{0.47\linewidth}>{\raggedright\arraybackslash}p{0.47\linewidth}@{}}
\toprule
\textbf{Positive prompts ($K{=}4$)} & \textbf{Negative prompts ($L{=}4$)} \\
\midrule
\multicolumn{2}{@{}l}{\emph{FormulaOne}} \\
``the racecar is centered on the track and driving safely'' & ``the racecar is about to crash into the barrier'' \\
``the racecar is following the track perfectly with smooth steering'' & ``the racecar is off the track and unsafe'' \\
``the racecar is making progress along the racing line'' & ``the car is colliding with obstacles'' \\
``the car is driving efficiently without collisions'' & ``the car is driving in the wrong direction'' \\
\midrule
\multicolumn{2}{@{}l}{\emph{MetaDrive}} \\
``the car is staying centered in its lane and driving smoothly'' & ``the car is about to crash into another vehicle'' \\
``the car is making safe forward progress along the road'' & ``the car is drifting onto the sidewalk or off the road'' \\
``the car is navigating traffic safely with proper spacing'' & ``the car is too close to another vehicle and unsafe'' \\
``the car is following the road and lane markings correctly'' & ``the car is veering off the lane in a dangerous way'' \\
\midrule
\multicolumn{2}{@{}l}{\emph{Bullet}} \\
``the car is driving directly toward the goal target'' & ``the car is about to collide with an obstacle'' \\
``the car is moving smoothly through the open area between obstacles'' & ``the car is touching a hazardous zone or obstacle'' \\
``the car is making clear progress and avoiding all obstacles'' & ``the car is stuck or moving in the wrong direction'' \\
``the car is on a safe trajectory toward the next goal'' & ``the car is in a dangerous position near a barrier'' \\
\bottomrule
\end{tabular}
\end{table}

\subsection{Prompt-template sensitivity}
\label{app:prompt-templates}

We compare the default v1 prompts with two alternatives under the same training
and evaluation settings within each environment. The generic v2 template uses
``the car is driving safely'' and ``the car is in an unsafe situation''. The
action-specific v3 template uses two prompts per group. Its MetaDrive positives
are ``the car has clear space ahead with no imminent collision'' and ``the car
is aligned with the lane direction''. Its negatives are ``the car is about to
collide with another vehicle in front of it'' and ``the car is laterally
drifting outside the lane boundary''. The Bullet positives are ``the car has
clear space ahead with no obstacles in the way'' and ``the car is far from any
obstacle or hazard zone''. Its negatives are ``the car is about to collide with
an obstacle directly in front of it'' and ``the car is overlapping with a
hazardous zone''.

\begin{figure}[h]
  \centering
  \includegraphics[width=0.98\linewidth]{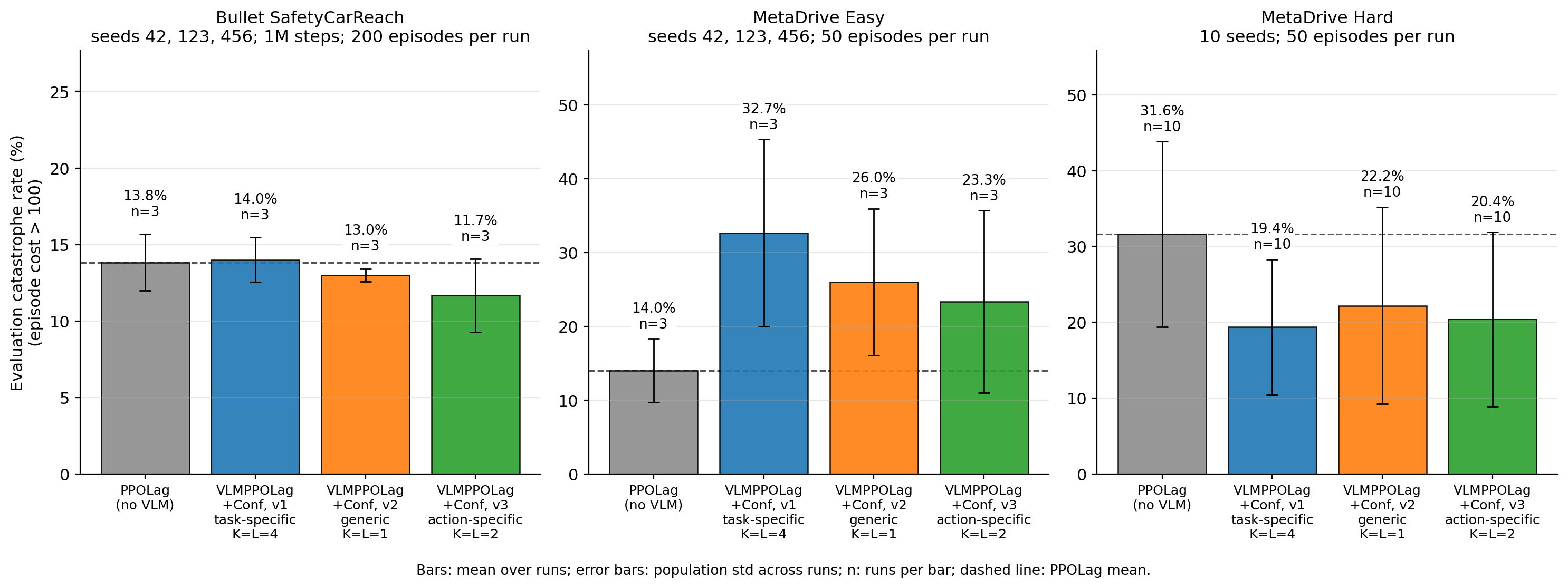}
  \caption{Catastrophe rates for PPOLag and VLMPPOLag$+$Conf under the v1, v2
  and v3 prompt sets. Bullet uses three training runs per method and $200$
  evaluation episodes per run. MetaDrive uses $50$ in-distribution episodes per
  run with three training runs on Easy and ten on Hard. Bars show mean per-run
  rates, error bars show population standard deviations, and dashed lines mark
  the PPOLag mean.}
  \label{fig:prompt-sensitivity}
\end{figure}

\Cref{fig:prompt-sensitivity} reports catastrophe-rate point estimates under a
common evaluation protocol within each environment. On MetaDrive Hard, PPOLag
reaches $31.6\%$, while v1, v2 and v3 reach $19.4\%$, $22.2\%$ and $20.4\%$.
All three prompt templates yield lower point estimates. On Easy, they exceed the
PPOLag value of $14.0\%$, reaching $32.7\%$, $26.0\%$ and $23.3\%$. On Bullet,
PPOLag reaches $13.8\%$, while the templates range from $11.7\%$ to $14.0\%$.
Bullet episodes cannot be matched by seed, and each template has only three
training runs. We therefore do not interpret these small differences as evidence
of a prompt effect.

On MetaDrive Hard, the point-estimate improvement appears across all three
prompt templates, although its magnitude varies. Over seeds $\{42,123,456\}$,
the decreases relative to PPOLag are $18.7$, $5.3$ and $7.3$ percentage points
for v1, v2 and v3. Across ten seeds, they are $12.2$, $9.4$ and $11.2$ points.
We report these as descriptive comparisons because intervals were not computed.

\subsection{Comparison with VLM-augmented methods}
\label{app:comparison}

\Cref{tab:comparison} positions VLM-Safe-RL against the closest VLM-augmented
approaches. Our method keeps the positive and negative CLIP scores separate,
uses the positive score for reward shaping, and sends the epoch mean negative
score to the multiplier. The multiplier contribution is negligible in the
reported runs (\S\ref{sec:results-eta2}).

\begin{table}[t]
  \caption{Comparison of VLM-augmented approaches. Ours denotes
  VLMPPOLag$+$Conf. $^{\dagger}$ denotes the CPO adaptation of the VLM-RL
  CLG reward, used as a baseline.}
  \label{tab:comparison}
  \centering
  \scriptsize
  \setlength{\tabcolsep}{3pt}
  \renewcommand{\arraystretch}{0.95}
  \begin{tabular}{@{}lcccc@{}}
    \toprule
     & \textbf{SafeVLA}\,{\tiny\cite{zhang2025safevla}}
     & \textbf{VLM-RL}\,{\tiny\cite{huang2024vlmrl}}
     & \textbf{CPO-CLG}$^{\dagger}$\,{\tiny\cite{huang2024vlmrl}}
     & \textbf{Ours} \\
    \midrule
    Safety formulation       & CMDP   & none    & CMDP    & CMDP \\
    CLIP scoring             & N/A    & CLG difference & CLG difference & separate cosines \\
    VLM role                 & policy & reward  & reward  & reward and dual mean \\
    Lagrange multiplier      & yes    & no      & no      & yes \\
    VLM term in multiplier   & no     & no      & no      & yes (negligible) \\
    Confidence gating        & no     & no      & no      & yes \\
    Frozen VLM               & no     & yes     & yes     & yes \\
    \bottomrule
  \end{tabular}
\end{table}

\section{Extended FormulaOne Results}
\label{app:results-f1}

This appendix reports the FormulaOne training and evaluation results. It also
examines the cost budget, task progress, additional constrained-RL baselines and
component ablations.

\subsection{Training cost, environment-only evaluation and training curves}
\label{app:f1-main-tables}

\begin{figure}[h]
  \centering
  \includegraphics[width=\textwidth]{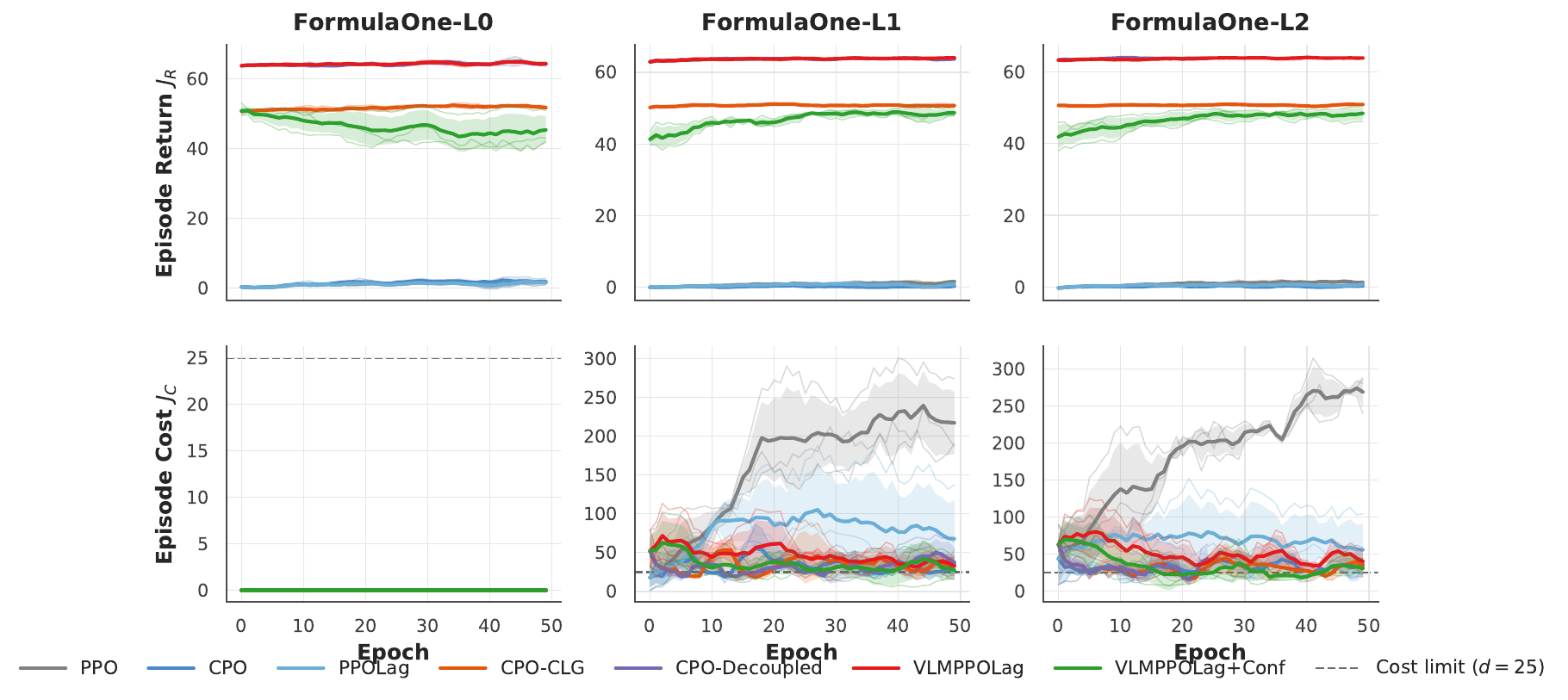}
  \caption{Training curves on FormulaOne-L0/L1/L2 for seven methods (PPO, CPO,
    PPOLag, CPO-CLG, CPO-Decoupled, VLMPPOLag and VLMPPOLag$+$Conf), $3$ seeds
    $\{42,123,456\}$ each: logged episode return $\Jr$ (top row) and environment
    cost $\Jc$ (bottom row) per epoch. For the VLM-shaped arms (CPO-CLG,
    CPO-Decoupled, VLMPPOLag and VLMPPOLag$+$Conf) $\Jr$ is the augmented return,
    which includes the CLIP shaping bonus, so the top row does not compare task
    performance across arms. Bold lines are seed means, shaded bands $\pm1$
    population standard deviation over seeds, and faded lines individual seeds.
    The grey dashed line marks $\dlim{=}25$. VLMPPOLag$+$Conf here is the
    three-seed prior-symmetric set at $k_{\text{clip}}{=}1$, not the calibrated
    five-seed cells of \Cref{tab:main_results}. PPOLag uses the unmatched
    VLM-free configuration. Curves are truncated to the shortest run in each
    cell, so CPO-Decoupled on L2 ends at epoch~$25$.}
  \label{fig:learning-curves}
\end{figure}

\Cref{tab:main_results} reports training-time environment cost on FormulaOne
L0--L2 and environment-only evaluation on L2. We report the two L2 evaluation
sets separately because they use different episode indices. Seed counts also
vary across methods. The training curves appear in \Cref{fig:learning-curves},
with task metrics and the budget sweep reported in Apps.~\ref{app:task-metrics}
and~\ref{app:pareto-baseline}.

\begin{table}[h]
\centering
\caption{\textbf{FormulaOne training cost and environment-only evaluation.}
\textbf{(A)} Training-time environment cost after $10^6$ steps, reported as the
mean $\pm$ population standard deviation over seeds at the final logged epoch.
The calibrated VLMPPOLag$+$Conf L2 entry instead uses the final $10$ epochs. Blue
entries satisfy $\Jc{\le}\dlim{=}25$. The calibrated method uses five seeds, the
matched PPOLag uses two, and the remaining entries use three unless unavailable.
\textbf{(B)} Environment-only evaluation on FormulaOne-L2 with $50$
deterministic episodes per run. Return and cost are means over runs, while
violation and catastrophe rates are pooled over episodes. $^{\dagger}$One run
was excluded because it contained only an untrained initial policy.
$^{\ddagger}$A repeated evaluation of the same three policies gives
$40.89\pm14.27$. Rows differ in training settings and are not uniformly matched.
In particular, comparisons involving the calibrated method may change both the
gate and the CLIP query period. Apps.~\ref{app:f1-run-sets}
and~\ref{app:perm-tests} give the run composition and statistical comparisons.}
\label{tab:main_results}
\tiny
\setlength{\tabcolsep}{2pt}
\renewcommand{\arraystretch}{0.78}
\rowcolors{3}{gray!10}{white}
\textbf{(A)}\\[1pt]
\begin{tabular}{@{}llccc@{}}
\toprule
\textbf{Type} & \textbf{Method} & \textbf{L0} $\Jc$ & \textbf{L1} $\Jc$ & \textbf{L2} $\Jc$ \\
\midrule
RL   & PPO                         & \textcolor{blue}{0.0$\pm$0.0} & 217.0$\pm$40.2 & 269.2$\pm$20.7 \\
CMDP & CPO                         & \textcolor{blue}{0.0$\pm$0.0} & 35.7$\pm$13.2  & 36.1$\pm$16.4  \\
CMDP & PPOLag (unmatched)          & \textcolor{blue}{0.0$\pm$0.0} & 67.9$\pm$49.7  & 55.8$\pm$35.7  \\
CMDP & PPOLag (matched, $n{=}2$)   & n/a                          & n/a            & \textcolor{blue}{20.9$\pm$5.5} \\
CMDP & CPPOPID                     & n/a                          & n/a            & \textcolor{blue}{22.8$\pm$6.7} \\
CLG  & PPO-CLG                     & \textcolor{blue}{0.0$\pm$0.0} & 133.6$\pm$37.3 & 156.6$\pm$40.4 \\
CLG  & CPO-CLG                     & \textcolor{blue}{0.0$\pm$0.0} & 32.8$\pm$8.4   & 33.9$\pm$5.8   \\
RND  & PPOLag-RND                  & n/a                          & 62.4$\pm$18.6  & 45.7$\pm$13.9  \\
Ours & CPO-Coupled                 & \textcolor{blue}{0.0$\pm$0.0} & 29.1$\pm$4.1   & 32.4$\pm$4.4   \\
Ours & CPO-Decoupled               & \textcolor{blue}{0.0$\pm$0.0} & 37.6$\pm$19.4  & 30.9$\pm$11.0  \\
Ours & PPOLag-Decoupled            & \textcolor{blue}{0.0$\pm$0.0} & 33.5$\pm$1.7   & 40.7$\pm$6.1   \\
Ours & VLMPPOLag                   & \textcolor{blue}{0.0$\pm$0.0} & 32.8$\pm$8.2   & 40.2$\pm$12.6  \\
Ours & VLMPPOLag$+$Conf$^{*}$      & \textcolor{blue}{0.0$\pm$0.0} & \textcolor{blue}{20.8$\pm$13.0} & \textcolor{blue}{22.5$\pm$5.9} \\
\bottomrule
\end{tabular}\\[3pt]
\textbf{(B)}\\[1pt]
\begin{tabular}{@{}lccccc@{}}
\toprule
\textbf{Method} & $n$ & env $\Jr$ & env $\Jc$ & \textbf{Viol.\,\%} & \textbf{Cat.\,\%} \\
\midrule
PPO                                   & 3 & 0.627$\pm$0.283    & 291.49$\pm$31.42 & 74.0 & 66.0 \\
CPO                                   & 3 & 0.272$\pm$0.167    & 50.21$\pm$24.92  & 21.3 & 13.3 \\
PPOLag (unmatched)                    & 3 & 0.191$\pm$0.168    & 108.57$\pm$23.80 & 29.3 & 22.7 \\
PPOLag (matched)                      & 2 & 0.055$\pm$0.024    & 10.73$\pm$2.95   & 14.0 & 2.0  \\
CPPOPID                               & 3 & 0.131$\pm$0.045    & 23.16$\pm$9.75   & 15.3 & 8.7  \\
PPO-CLG                               & 2 & 0.356$\pm$0.044    & 179.94$\pm$19.10 & 51.0 & 44.0 \\
CPO-CLG                               & 3 & $-$0.026$\pm$0.127 & 55.93$\pm$26.54  & 23.3 & 18.0 \\
PPOLag-RND                            & 3 & 0.254$\pm$0.128    & 35.99$\pm$7.20$^{\ddagger}$ & 24.7 & 14.7 \\
CPO-Coupled                           & 1 & 0.094              & 127.32           & 48.0 & 28.0 \\
CPO-Decoupled                         & 2 & 0.409$\pm$0.026    & 64.43$\pm$3.79   & 33.0 & 17.0 \\
CPO-Decoupled$+$Conf                  & 3 & 0.233$\pm$0.085    & 59.30$\pm$15.88  & 18.7 & 13.3 \\
PPOLag-Decoupled$^{\dagger}$          & 2 & 0.062$\pm$0.026    & 25.14$\pm$14.06  & 16.0 & 6.0  \\
VLMPPOLag (ungated)                   & 3 & 0.247$\pm$0.056    & 39.63$\pm$15.66  & 22.7 & 13.3 \\
VLMPPOLag, $\eta_2{=}0$ ($k{=}4$)     & 3 & 0.191$\pm$0.064    & 27.37$\pm$11.00  & 21.3 & 10.0 \\
VLMPPOLag$+$Conf, prior-symmetric     & 3 & 0.092$\pm$0.126    & 36.77$\pm$10.43  & 19.3 & 10.7 \\
VLMPPOLag$+$Conf$^{*}$ (calibrated)   & 5 & 0.062$\pm$0.149    & 22.06$\pm$11.04  & 16.8 & 6.4  \\
\bottomrule
\end{tabular}
\end{table}

\paragraph{Evaluation cost tests.} On environment-only evaluation (panel~B), the
calibrated VLMPPOLag$+$Conf runs of family~A have a mean cost of $22.06$
($n{=}5$). Exact one-sided permutation tests over run means
(App.~\ref{app:perm-tests}) give $p{=}1/56{=}0.018$ against the unmatched PPOLag
($108.57$), $p{=}4/56{=}0.071$ against CPO ($50.21$), $p{=}5/56{=}0.089$ against
ungated VLMPPOLag ($39.63$) and $p{=}28/56{=}0.50$ against CPPOPID ($23.16$). On
evaluation cost the calibrated runs therefore separate at the $0.05$ level only
from the unmatched PPOLag. We report no test against PPOLag-Decoupled ($25.14$), which has
only $n{=}2$ evaluated runs. Against the matched PPOLag ($10.73$, $n{=}2$) the
direction reverses ($p{=}19/21{=}0.90$), as does the $20$-episode catastrophe
comparison of \Cref{tab:generalisation} ($8.0\%$ against $2.5\%$). Mean
evaluation cost is within budget for $2/2$ matched PPOLag runs, $3/5$
VLMPPOLag$+$Conf runs, $1/2$ PPOLag-Decoupled runs and $1/3$ CPPOPID runs. With
these seed counts the rankings are descriptive, and because cost tracks distance
travelled on this task (\Cref{tab:f1-task}), they are not a safety ordering. Note
also that a comparison of the calibrated VLMPPOLag$+$Conf
with VLMPPOLag or PPOLag-Decoupled crosses a change of gate calibration and CLIP
period as well as the gate itself.

\paragraph{The gating pair.} The seed-matched pair of \Cref{tab:ablation} uses the
prior-symmetric gate at $k_{\text{clip}}{=}1$. Its final-epoch training costs for
seeds $42$, $123$ and $456$ are $23.15$, $44.21$ and $24.12$ with the gate and
$53.89$, $23.37$ and $43.23$ without it. The gate therefore lowers training cost for two of the three
seeds and raises it for seed~$123$. On the panel-B evaluation the same
checkpoints give per-run mean costs of $24.68$, $50.14$ and $35.48$ (mean $36.77$)
with the gate and $46.44$, $17.98$ and $54.48$ (mean $39.63$) without it, with the
same pattern by seed. The calibrated five-seed L2 cell (family~A, last-10-epoch
mean $\Jc{=}22.5$, $4/5$ seeds within budget) also changes the gate parameters and the CLIP period
(App.~\ref{app:compute}). On the three seeds it shares with the pair, its last-10-epoch mean is $23.7$ ($2/3$ within budget). The prior-symmetric gate
gives $30.5$ at the final epoch and $29.4$ over the last $10$ epochs, so most of
the difference comes from the change of configuration, not from seeds $789$ and
$1024$.

\paragraph{Decomposition of the logged return.} For a shaped run we recover the
environment return as the logged return minus the bonus
$\lambda_r\,\overline{\rvlm}\,T$, with $\lambda_r{=}0.1$, $\overline{\rvlm}$ the
mean per-step VLM reward and $T{=}1000$ steps
per episode. \Cref{tab:f1-return-decomp} applies this to the ungated L2 arms, where
the bonus accounts for the overwhelming majority of the logged return (for
example, $63.63$ of $63.88$ for CPO-Decoupled at the final epoch). Over the last
$5$ epochs the nine ungated decoupled-path runs give environment returns of
$0.14$--$0.55$, and the unmatched VLM-free PPOLag runs, whose logged return needs
no correction, give $0.33$--$0.72$. The two groups therefore share a narrow band
near zero, with the PPOLag runs if anything slightly ahead. Evaluation on the plain
environment agrees (panel~B), since the VLM-shaped arms return between $-0.03$
and $0.41$ and the VLM-free arms between $0.06$ and $0.63$. The CPO-Coupled and CPO-Decoupled means differ by $0.018$, $0.049$
and $0.095$ under the three aggregations, and their order flips between the last-5
and last-10 windows. As final-epoch seed means at every level, the ungated decoupled-path methods log
$\Jr$ between $63.7$ and $64.3$ and both CLG methods between $50.7$ and $51.9$.
The gated arms log less, $45$--$49$ for the three-seed prior-symmetric
VLMPPOLag$+$Conf, $43$--$49$ for CPO-Decoupled$+$Conf and $32$--$45$ for the
calibrated five-seed cells, and PPO logs
$1.6$ at L0 and L1 and $1.3$ at L2.

\begin{table}[h]
\centering
\small
\caption{Decomposition of the logged training return on FormulaOne-L2 (seeds
$\{42,123,456\}$). Logged $\Jr$ and bonus
$\lambda_r\,\overline{\rvlm}\,T$ are final-epoch seed means. Environment return is
the logged return minus the bonus, averaged over the last $5$ or $10$ logged
epochs or taken at the final epoch, then averaged over seeds. Runs that stopped
early use their last logged epochs (CPO-Coupled seeds $42$ and $456$,
CPO-Decoupled seed~$42$, PPOLag-Decoupled seed~$42$). The VLM-free PPOLag
(unmatched) logs its environment return directly.}
\label{tab:f1-return-decomp}
\setlength{\tabcolsep}{4pt}
\begin{tabular}{@{}lccccc@{}}
\toprule
& \textbf{Logged} $\Jr$ & \textbf{Bonus} & \multicolumn{3}{c}{\textbf{Environment return}} \\
\cmidrule(lr){4-6}
\textbf{Arm} & final & final & last-5 & last-10 & final \\
\midrule
CPO-Coupled         & 21.62 & 21.28 & 0.381 & 0.367 & 0.337 \\
CPO-Decoupled       & 63.88 & 63.63 & 0.398 & 0.318 & 0.242 \\
PPOLag-Decoupled    & 63.77 & 63.54 & 0.318 & 0.355 & 0.228 \\
VLMPPOLag           & 63.83 & 63.50 & 0.319 & 0.397 & 0.328 \\
PPOLag (unmatched)  & 0.74  & 0     & 0.537 & 0.509 & 0.744 \\
\bottomrule
\end{tabular}
\end{table}

\paragraph{Training curves.} \Cref{fig:learning-curves} shows $\Jc$ rising with
obstacle density for PPO (final-epoch seed means $0$, $217$ and $269$ at L0, L1 and
L2). The final-epoch seed means of the VLMPPOLag and VLMPPOLag$+$Conf traces lie
between $27.6$ and $40.2$ at L1 and L2 ($32.8$ and $40.2$ for VLMPPOLag, $27.6$ and
$30.5$ for VLMPPOLag$+$Conf), all above the budget. Because episodic cost tracks
distance travelled on this unsolved task (\Cref{tab:f1-task}), this contrast is not by itself a competence result.
The $\Jr$ curves are augmented returns and carry no claim about task competence.

\subsection{One-sided permutation tests on FormulaOne-L2}
\label{app:perm-tests}

Each comparison uses an exact one-sided permutation test on the difference in
group means. This avoids a Gaussian approximation with only three to five seeds
per group. For group sizes $(n_1,n_2)$, the minimum attainable $p$-value is
$1/\binom{n_1+n_2}{n_1}$. It is therefore $1/56\approx0.018$ for five against
three runs and $1/20=0.050$ for three against three. The alternative direction
was specified before testing.

\Cref{tab:perm-l2} reports eight comparisons. The calibrated
VLMPPOLag$+$Conf group contains five seeds, while each comparator contains seeds
$\{42,123,456\}$. Every run contributes its mean over its final $10$ logged
epochs. Runs that ended early use their final $10$ available epochs. We analyse
only calibrated family~A because the independently trained family~B produces
different per-seed costs despite using the same configuration
(\Cref{tab:conf-families}).

Six comparisons attain their minimum possible $p$-value. The exceptions are the
logged-return comparison with CPO-Coupled at $p{=}7/56$ and the cost comparison
between VLMPPOLag and PPOLag-Decoupled at $p{=}10/20$. The three family-A cost
comparisons are sensitive to the aggregation window. Their final-epoch values
give $p{=}2/56$, while their final-10-epoch values attain $1/56$. Across all
eight comparisons, the number attaining the minimum consequently falls from six
to three when only the final epoch is used.

These tests are not single-factor ablations. Comparisons with the calibrated
method may change both the gate and the CLIP query period. The matched gating
pair gives $p{=}0.30$ at the final epoch and $p{=}0.10$ over the final $10$
epochs, providing no detectable gating effect with three seeds. The cost rows do
not establish safety differences because FormulaOne cost increases with distance
travelled and no evaluated method completes a route stage
(App.~\ref{app:task-metrics}). The return rows also do not compare task
performance because VLM-shaped methods report augmented return. Finally,
VLMPPOLag and PPOLag-Decoupled differ only through the VLM term in the multiplier.
Their cost comparison gives $p{=}0.50$ under both aggregations, so it provides no
evidence of an effect from that term.

\begin{table}[h]
\centering
\footnotesize
\caption{Exact one-sided permutation tests on FormulaOne-L2. Each run contributes
its mean over the final $10$ logged epochs. The calibrated group contains five
seeds, while each comparator contains three. $\bar{x}_1$ and $\bar{x}_2$ are the
group means and $\Delta=\bar{x}_1-\bar{x}_2$. The minimum attainable $p$-value
is $1/56$ for comparisons with group sizes $(5,3)$ or $(3,5)$ and $1/20$ for
$(3,3)$. Bold values attain this minimum. Environment cost tracks distance
travelled on this task, and logged return includes the shaping bonus for
VLM-shaped methods. $^{a}$PPOLag-RND instead reports environment return.}
\label{tab:perm-l2}
\setlength{\tabcolsep}{3pt}
\begin{tabular}{@{}llccrrrc@{}}
\toprule
\textbf{Group 1} & \textbf{Group 2} & $(n_1,n_2)$ & $H_1$ & $\bar{x}_1$ & $\bar{x}_2$ & $\Delta$ & $p$ \\
\midrule
\multicolumn{8}{@{}l}{\textit{Environment cost} $\Jc$} \\
VLMPPOLag$+$Conf$^{*}$ & VLMPPOLag              & $(5,3)$ & $<$ & $22.54$ & $43.70$ & $-21.16$ & $\mathbf{1/56}$ \\
VLMPPOLag$+$Conf$^{*}$ & PPOLag-Decoupled       & $(5,3)$ & $<$ & $22.54$ & $43.75$ & $-21.21$ & $\mathbf{1/56}$ \\
VLMPPOLag$+$Conf$^{*}$ & PPOLag-RND             & $(5,3)$ & $<$ & $22.54$ & $46.73$ & $-24.19$ & $\mathbf{1/56}$ \\
VLMPPOLag              & PPOLag-Decoupled       & $(3,3)$ & $<$ & $43.70$ & $43.75$ & $-0.05$  & $10/20$ \\
\midrule
\multicolumn{8}{@{}l}{\textit{Logged return} $\Jr$ \textit{(augmented for
VLM-shaped arms)}} \\
VLMPPOLag$+$Conf$^{*}$ & PPOLag-RND$^{a}$       & $(5,3)$ & $>$ & $31.80$ & $0.45$  & $+31.35$ & $\mathbf{1/56}$ \\
VLMPPOLag$+$Conf$^{*}$ & CPO-Coupled            & $(5,3)$ & $>$ & $31.80$ & $21.51$ & $+10.30$ & $7/56$ \\
Qwen2-VL$+$Conf        & VLMPPOLag$+$Conf$^{*}$ & $(3,5)$ & $<$ & $8.30$  & $31.80$ & $-23.51$ & $\mathbf{1/56}$ \\
CPO-Decoupled          & CPO-Coupled            & $(3,3)$ & $>$ & $63.80$ & $21.51$ & $+42.29$ & $\mathbf{1/20}$ \\
\bottomrule
\end{tabular}
\end{table}

\subsection{Budget sweep for PPOLag-Decoupled}
\label{app:pareto-baseline}

We vary the PPOLag-Decoupled cost limit on FormulaOne-L2 to examine its observed
cost--return trade-off. The additional settings $\dlim\in\{15,35\}$ each use one
seed trained for $10^6$ steps with $k_{\mathrm{clip}}{=}4$. We compare them with
the three-seed $\dlim{=}25$ setting and include calibrated VLMPPOLag$+$Conf as a
reference.

PPOLag-Decoupled exceeds its specified budget at all three settings under both
the final-epoch and final-10-epoch summaries. This limited sweep therefore does
not identify a budget-feasible operating point. The comparison is descriptive
because the added settings contain one seed each, while the $\dlim{=}25$ runs
use $k_{\mathrm{clip}}{=}1$. All reported returns include the CLIP shaping bonus
and cannot be compared with VLM-free returns.

\begin{table}[h]
\centering
\small
\caption{PPOLag-Decoupled budget sweep on FormulaOne-L2. $\Jc$ is reported at
the final epoch and as a mean over the final $10$ epochs. $\Jr$ is the
final-10-epoch augmented return. The $\dlim\in\{15,35\}$ entries are single
seed-$42$ runs with $k_{\mathrm{clip}}{=}4$. The $\dlim{=}25$ entry averages
three runs with $k_{\mathrm{clip}}{=}1$, including one run that ended at epoch
$39$. The calibrated five-seed method is included only as a reference. The rows
are not matched.}
\label{tab:pareto-baseline}
\setlength{\tabcolsep}{5pt}
\begin{tabular}{@{}lccccc@{}}
\toprule
\textbf{Method} & $\dlim$ & $\Jr$ (augmented) & $\Jc$ final & $\Jc$ last-10 & \textbf{Over budget} \\
\midrule
PPOLag-Decoupled       & 15 & 63.8 & 33.8 & 42.5 & 1/1 \\
PPOLag-Decoupled       & 25 & 63.8 & 40.7 & 43.8 & 3/3 \\
PPOLag-Decoupled       & 35 & 63.9 & 49.6 & 46.5 & 1/1 \\
\midrule
VLMPPOLag$+$Conf$^{*}$ & 25 & 31.8 & 22.4 & 22.5 & 1/5 \\
\bottomrule
\end{tabular}
\end{table}

\subsection{Per-run FormulaOne-L2 task metrics}
\label{app:task-metrics}

Logged return includes the shaping bonus for VLM-based methods and therefore
does not measure FormulaOne task performance directly. We evaluate six methods
without VLM augmentation for $50$ deterministic episodes per run. FormulaOne-L2
contains seven sequential route stages. For each episode, we measure completed
stages, environment return and cost, path length and net progress. Net progress
is the reduction from the initial distance to the closest distance reached from
the current goal.

The evaluation contains $20$ runs and $1{,}000$ episodes. PPO, CPO, the
unmatched PPOLag, PPOLag-Decoupled and VLMPPOLag use seeds
$\{42,123,456\}$. Calibrated VLMPPOLag$+$Conf uses five seeds. The matched
PPOLag control was not evaluated under this protocol. One PPOLag-Decoupled run
contains only an untrained initial policy, so we report its per-run values but
exclude it from every arm-level summary. This leaves two trained runs for that
method. No evaluated policy reaches or completes any route stage.

\begin{table}[h]
\centering
\footnotesize
\caption{Per-run FormulaOne-L2 task metrics from $50$ deterministic episodes per
run. Path length and net progress use simulator distance units. Return and cost
are measured without VLM augmentation. Viol. denotes episodes with
$\Jc{>}\dlim{=}25$, and Cat. denotes episodes with $\Jc{>}4\dlim$. No episode
completes a route stage. $^{*}$The calibrated five-seed method.
$^{\dagger}$An untrained initial policy, excluded from all arm-level summaries.}
\label{tab:task-metrics}
\setlength{\tabcolsep}{3pt}
\begin{tabular}{@{}lrcccccc@{}}
\toprule
\textbf{Arm} & \textbf{Seed} & \textbf{Path} & \textbf{Progress} & \textbf{Cost} & \textbf{Return} & \textbf{Viol.\,\%} & \textbf{Cat.\,\%} \\
\midrule
PPO (no VLM)               & 42   & 5.278 & 2.265 & 346.30 & 1.522 & 74 & 68 \\
                           & 123  & 6.902 & 2.611 & 357.48 & 1.535 & 82 & 78 \\
                           & 456  & 4.889 & 2.006 & 322.30 & 1.292 & 66 & 58 \\
\midrule
CPO (no VLM)               & 42   & 2.977 & 0.673 & 41.44  & 0.097 & 18 & 12 \\
                           & 123  & 2.317 & 0.498 & 50.58  & 0.053 & 12 & 10 \\
                           & 456  & 1.393 & 0.430 & 24.68  & 0.177 & 12 & 8 \\
\midrule
PPOLag (no VLM, unmatched) & 42   & 5.225 & 1.291 & 68.06  & 0.716 & 22 & 18 \\
                           & 123  & 2.082 & 0.812 & 82.98  & 0.497 & 18 & 16 \\
                           & 456  & 3.106 & 1.032 & 106.26 & 0.664 & 28 & 24 \\
\midrule
PPOLag-Decoupled           & 42$^{\dagger}$ & 3.776 & 1.091 & 307.90 & 0.302 & 58 & 50 \\
                           & 123  & 1.810 & 0.544 & 27.46  & 0.191 & 18 & 8 \\
                           & 456  & 2.009 & 0.589 & 15.82  & 0.272 & 10 & 4 \\
\midrule
VLMPPOLag                  & 42   & 2.036 & 0.528 & 29.70  & 0.185 & 22 & 10 \\
                           & 123  & 1.789 & 0.348 & 12.10  & $-$0.025 & 10 & 2 \\
                           & 456  & 2.590 & 0.766 & 25.88  & 0.560 & 16 & 10 \\
\midrule
VLMPPOLag$+$Conf$^{*}$     & 42   & 2.582 & 0.484 & 11.10  & 0.147 & 8  & 4 \\
                           & 123  & 2.034 & 0.321 & 13.82  & 0.106 & 16 & 2 \\
                           & 456  & 2.699 & 0.474 & 25.70  & 0.227 & 18 & 4 \\
                           & 789  & 1.619 & 0.277 & 10.66  & 0.081 & 8  & 4 \\
                           & 1024 & 2.592 & 0.417 & 17.26  & 0.061 & 16 & 8 \\
\bottomrule
\end{tabular}
\end{table}

\textbf{Cost and movement.} Across the six arm means in \Cref{tab:f1-task},
environment cost has Pearson correlations of $0.978$ with path length, $0.991$
with net progress and $0.977$ with environment return. This relationship is
driven mainly by PPO and the unmatched PPOLag, which travel furthest and incur
the most cost. Among the other four arms, mean path length spans only
$1.91$--$2.31$, while mean cost ranges from $15.71$ to $38.90$. Movement alone
therefore does not explain their cost differences. Across the $950$ episodes
from trained policies, path length has Pearson and Spearman correlations of
$0.382$ and $0.531$ with cost. The corresponding correlations for net progress
are $0.331$ and $0.231$. Since no method completes a route stage, lower cost on
this task cannot by itself be interpreted as safer control.

\subsection{FormulaOne-L2 run sets and incomplete runs}
\label{app:f1-run-sets}

\paragraph{Two VLM-free PPOLag controls.} The matched control uses two seeds,
$40$ update iterations and $\lambda_0{=}10^{-3}$, matching the corresponding
settings of calibrated VLMPPOLag$+$Conf. The unmatched baseline uses three
seeds, $10$ update iterations and $\lambda_0{=}0$. The controls therefore differ
in optimisation settings and seed sets, so their performance gap cannot be
attributed to a single factor.

Across $50$ evaluation episodes per run, their mean costs are $10.73$ and
$108.57$, respectively. On the first $20$ episodes, the matched control has a
catastrophe rate of $2.5\%$ and a violation rate of $17.5\%$. The unmatched
baseline reaches $18.3\%$ and $26.7\%$, while calibrated VLMPPOLag$+$Conf
reaches $8.0\%$ and $18.0\%$. We use the two-seed control as the matched
comparison and report the three-seed control as an unmatched baseline. The
matched seeds are disjoint from the main three-seed set, and this control was
not included in the task-metric evaluation.

\paragraph{Two calibrated VLMPPOLag$+$Conf families.} Families A and B are
independent repetitions with the same five seeds and training configuration.
Both use the calibrated gate from \eqnref{eq:mle-sc}, $500$ calibration frames,
$\kappa^{\star}{=}0.5$, $k_{\mathrm{clip}}{=}4$, $40$ update iterations and
$\lambda_0{=}10^{-3}$. \Cref{tab:conf-families} reports their per-seed costs.
Family B has the lower mean under both aggregations, although its final epoch has
fewer seeds within budget. Family A supplies the FormulaOne-L2 result in
\Cref{tab:main_results}. The calibrated L0 and L1 results come from the training
campaign associated with family B. The three levels in that table therefore do
not form a single replicated run family.

\begin{table}[h]
\centering
\small
\caption{Training-time environment cost for two independent calibrated
VLMPPOLag$+$Conf families on FormulaOne-L2. Final denotes epoch $50$, and
Last-10 averages epochs $41$--$50$. Summary rows report the mean $\pm$
population standard deviation over five seeds and the number satisfying
$\Jc{\le}\dlim{=}25$.}
\label{tab:conf-families}
\setlength{\tabcolsep}{5pt}
\begin{tabular}{@{}lcccc@{}}
\toprule
& \multicolumn{2}{c}{\textbf{Family A}} & \multicolumn{2}{c}{\textbf{Family B}} \\
\cmidrule(lr){2-3}\cmidrule(lr){4-5}
\textbf{Seed} & Final & Last-10 & Final & Last-10 \\
\midrule
42   & 17.75 & 16.14 & 21.37 & 21.57 \\
123  & 36.37 & 22.39 & 28.16 & 21.58 \\
456  & 21.70 & 32.63 & 13.61 & 12.13 \\
789  & 23.07 & 24.11 & 27.00 & 21.45 \\
1024 & 13.26 & 17.43 & 10.51 & 12.65 \\
\midrule
Mean $\pm$ std & $22.43\pm7.76$ & $22.54\pm5.85$ & $20.13\pm7.05$ & $17.88\pm4.48$ \\
Within budget  & $4/5$ & $4/5$ & $3/5$ & $5/5$ \\
\bottomrule
\end{tabular}
\end{table}

\paragraph{Incomplete training runs.} Several runs ended before epoch $50$.
Across seeds $\{42,123,456\}$, CPO-Coupled completed $39$, $50$ and $36$
epochs, CPO-Decoupled completed $25$, $50$ and $50$, and PPOLag-Decoupled
completed $39$, $50$ and $50$. The CPO-Coupled L1 run at seed $123$ ended after
two epochs. Training summaries use the final available epoch, while
environment-only evaluation includes only policies with a trained checkpoint.
This leaves one evaluated CPO-Coupled run and two each for CPO-Decoupled and
PPOLag-Decoupled. PPO-CLG also has two evaluated runs.

\paragraph{Evaluation of an untrained policy.} The seed-$42$
PPOLag-Decoupled run completed $39$ epochs but retained only its initial policy.
The environment-only evaluation excludes this run, while the task-metric
evaluation initially included it. Its mean cost is $307.90$, compared with
$27.46$ and $15.82$ for the two trained policies. Including it raises the arm
mean from $21.6$ to $117.1$ and the catastrophe rate from $6.0\%$ to $20.7\%$.
We therefore exclude it from all task-metric arm summaries. The training table
retains its observed training statistics, giving a mean cost of $40.7$ with the
run and $37.6$ without it.

\subsection{Additional constrained-RL baselines}
\label{app:extra-baselines}

We evaluate FOCOPS~\cite{zhang2020first}, CUP~\cite{yang2022cup} and
P3O~\cite{zhang2022penalized} to determine whether the FormulaOne budget
violations extend beyond CPO and PPOLag. Each method is trained without VLM
augmentation for $10^6$ steps with $\dlim{=}25$, $10$ update iterations and
seeds $\{42,123,456\}$ at every level. \Cref{tab:extra-baselines} reports the
mean return and cost over the final five epochs of each run.

All three methods satisfy the budget on L0. Every L1 run exceeds it, with mean
costs ranging from $45.1$ for FOCOPS to $133.0$ for P3O. Each L2 mean also
exceeds the budget, although one seed from each method remains within it. The L2
means are $27.6$, $33.0$ and $58.7$ for FOCOPS, CUP and P3O. These results show
that budget violations are not confined to CPO or PPOLag in our FormulaOne
experiments. They do not establish a safety ordering because no evaluated method
completes a route stage and cost alone does not measure task progress
(App.~\ref{app:task-metrics}).

\begin{table}[h]
\centering
\caption{Additional constrained-RL baselines on FormulaOne L0--L2 without VLM
augmentation. Each run contributes its mean over the final five of $50$ epochs.
Entries report the mean $\pm$ population standard deviation over seeds
$\{42,123,456\}$. $\Jr$ is environment return, and blue entries satisfy
$\Jc{\le}\dlim{=}25$.}
\label{tab:extra-baselines}
\footnotesize
\setlength{\tabcolsep}{3pt}
\renewcommand{\arraystretch}{0.95}
\begin{tabular}{@{}lcccccc@{}}
\toprule
& \multicolumn{2}{c}{\textbf{L0}}
& \multicolumn{2}{c}{\textbf{L1}}
& \multicolumn{2}{c}{\textbf{L2}} \\
\cmidrule(lr){2-3}\cmidrule(lr){4-5}\cmidrule(lr){6-7}
\textbf{Method} & $\Jr$ & $\Jc$ & $\Jr$ & $\Jc$ & $\Jr$ & $\Jc$ \\
\midrule
FOCOPS & 1.13$\pm$0.39 & \textcolor{blue}{0.0$\pm$0.0} & 0.26$\pm$0.08 & 45.1$\pm$17.7  & 0.35$\pm$0.33 & 27.6$\pm$10.2 \\
CUP    & 1.76$\pm$0.36 & \textcolor{blue}{0.0$\pm$0.0} & 0.38$\pm$0.17 & 46.9$\pm$10.1  & 0.33$\pm$0.09 & 33.0$\pm$5.8  \\
P3O    & 1.69$\pm$0.59 & \textcolor{blue}{0.0$\pm$0.0} & 0.04$\pm$0.35 & 133.0$\pm$29.9 & 0.00$\pm$0.18 & 58.7$\pm$33.0 \\
\bottomrule
\end{tabular}
\end{table}


\subsection{Aggregate FormulaOne-L2 task metrics}
\label{app:f1-task-metrics}

\begin{table}[t]
\centering
\caption{Aggregate FormulaOne-L2 task metrics from $50$ deterministic episodes
per run without VLM augmentation. $n$ denotes the number of trained policies.
Values are means $\pm$ population standard deviations over per-policy episode
means. Violation and catastrophe rates are pooled over episodes using thresholds
$\Jc{>}\dlim$ and $\Jc{>}4\dlim$. Stages denotes completed route stages out of
seven. Path length and net progress use simulator distance units. The matched
PPOLag control was not evaluated under this protocol. $^{\dagger}$One untrained
initial policy is excluded.}
\label{tab:f1-task}
\scriptsize
\setlength{\tabcolsep}{3pt}
\renewcommand{\arraystretch}{0.9}
\begin{tabular}{@{}lcccccccc@{}}
\toprule
\textbf{Method} & $n$ & env $\Jr$ & env $\Jc$ & \textbf{Viol.\,\%} & \textbf{Cat.\,\%} & \textbf{Stages} & \textbf{Path} & \textbf{Progress} \\
\midrule
PPO                          & 3 & 1.450$\pm$0.111 & 342.03$\pm$14.68 & 74.0 & 68.0 & 0/7 & 5.69$\pm$0.87 & 2.294$\pm$0.248 \\
CPO                          & 3 & 0.109$\pm$0.051 & 38.90$\pm$10.73  & 14.0 & 10.0 & 0/7 & 2.23$\pm$0.65 & 0.534$\pm$0.102 \\
PPOLag (unmatched)           & 3 & 0.626$\pm$0.093 & 85.77$\pm$15.72  & 22.7 & 19.3 & 0/7 & 3.47$\pm$1.31 & 1.045$\pm$0.196 \\
PPOLag-Decoupled$^{\dagger}$ & 2 & 0.231$\pm$0.040 & 21.64$\pm$5.82   & 14.0 & 6.0  & 0/7 & 1.91$\pm$0.10 & 0.566$\pm$0.022 \\
VLMPPOLag                    & 3 & 0.240$\pm$0.242 & 22.56$\pm$7.56   & 16.0 & 7.3  & 0/7 & 2.14$\pm$0.33 & 0.547$\pm$0.171 \\
VLMPPOLag$+$Conf             & 5 & 0.125$\pm$0.059 & 15.71$\pm$5.52   & 13.2 & 4.4  & 0/7 & 2.31$\pm$0.41 & 0.395$\pm$0.082 \\
\bottomrule
\end{tabular}
\end{table}

\subsection{Component comparisons on FormulaOne-L2}
\label{app:f1-ablation}

The first comparison isolates confidence gating, while the second isolates the
VLM term in the multiplier. Gating lowers the mean final-epoch cost from $40.2$
to $30.5$, but the seed-level effect changes direction and the exact one-sided
permutation test gives $p{=}0.30$. Removing the VLM multiplier term changes the
mean cost from $40.2$ to $40.7$, with $p{=}0.50$. Neither comparison provides
evidence of a component effect with three seeds. The coupled CPO and VLM-free
PPOLag rows are unmatched references and do not isolate individual components.

\begin{table}[t]
\centering
\caption{FormulaOne-L2 component comparisons over seeds $\{42,123,456\}$ at
the final logged epoch. Every VLM row scores each frame. $\Jr$ is augmented
return and does not measure task performance. Viol. counts seeds with
$\Jc{>}\dlim$. The first two rows differ only in confidence gating, and the
second and third differ only in the VLM multiplier term. The final two rows
change several settings and are included as unmatched references. Runs that
ended early contribute their final available epoch.}
\label{tab:ablation}
\small
\setlength{\tabcolsep}{5pt}
\begin{tabular}{@{}lccc@{}}
\toprule
\textbf{Configuration} & $\Jr$ (augmented) & $\Jc$ & \textbf{Viol.} \\
\midrule
Prior-symmetric gate          & 48.4 & 30.5 & 1/3 \\
Ungated, $\eta_2{=}0.01$      & 63.8 & 40.2 & 2/3 \\
Ungated, $\eta_2{=}0$         & 63.8 & 40.7 & 3/3 \\
Coupled CPO (unmatched)       & 21.6 & 32.4 & 3/3 \\
VLM-free PPOLag (unmatched)   &  0.7 & 55.8 & 2/3 \\
\bottomrule
\end{tabular}
\end{table}

\section{Five-Seed Calibrated FormulaOne Runs and Additional L2 Baselines}
\label{app:phaseB-robustness}

All runs in this section train for $10^6$ steps. We consider calibrated
VLMPPOLag$+$Conf on FormulaOne-L0, L1 and L2 using five seeds and
$k_{\text{clip}}{=}4$. On L2, we also evaluate CPPOPID with three seeds and
VLM-free PPOLag and CPO controls with two seeds. The PPOLag control matches the
calibrated configuration in its $40$ update iterations and
$\lambda_0{=}10^{-3}$. CPPOPID also uses $40$ iterations, while CPO uses $10$.
The comparisons are therefore not matched in every optimisation setting or
seed set. The earlier three-seed VLMPPOLag$+$Conf runs in
\Cref{tab:ablation} instead use the prior-symmetric gate and score every frame.

\paragraph{Degenerate-safety criterion.}
A constrained policy can reduce FormulaOne cost by making little or no forward
progress. Such a policy rarely enters obstacle regions, yet it does not solve
the task. We therefore treat low cost accompanied by near-zero environment
return as degenerate behaviour, not evidence of safe driving. Logged training
return cannot support this distinction for VLM-shaped policies because it
includes the shaping bonus.

\paragraph{Calibrated VLMPPOLag$+$Conf at five seeds.}
\Cref{tab:phaseB-conf} reports the five calibrated runs. L0 and L1 use the
final epoch, while L2 uses the last-ten-epoch mean reported in
\Cref{tab:main_results}. We retain both conventions because a final epoch can
differ substantially from its recent average. For example, the L1 cost for
seed $456$ is $44.4$ at the final epoch and $30.35$ over the last ten epochs.

At L2, the five-seed mean cost is $22.5\pm5.9$, with four seeds satisfying the
budget. At L1, the final-epoch mean is $20.8\pm13.0$, again with four seeds
within budget. The prior-symmetric three-seed runs obtain final-epoch costs of
$30.5\pm9.7$ on L2 and $27.6\pm12.3$ on L1. These run sets also differ in gate
parameters and CLIP period, so their difference cannot be attributed to seed
count or calibration alone.

\begin{table}[h]
\centering
\caption{Per-seed training metrics for calibrated VLMPPOLag$+$Conf on
FormulaOne after $10^6$ steps with $k_{\text{clip}}{=}4$. L0 and L1 report the
final epoch, while L2 reports the last-ten-epoch mean used in
\Cref{tab:main_results}. $\Jr$ is the augmented training return. Summary rows
show the mean $\pm$ population standard deviation over five seeds. Blue entries
satisfy $\Jc{\le}\dlim{=}25$.}
\label{tab:phaseB-conf}
\small
\setlength{\tabcolsep}{4pt}
\begin{tabular}{@{}lcccccc@{}}
\toprule
& \multicolumn{2}{c}{\textbf{F1-L0} (final)} & \multicolumn{2}{c}{\textbf{F1-L1} (final)} & \multicolumn{2}{c}{\textbf{F1-L2} (last 10)} \\
\cmidrule(lr){2-3}\cmidrule(lr){4-5}\cmidrule(lr){6-7}
\textbf{Seed} & $\Jr$ & $\Jc$ & $\Jr$ & $\Jc$ & $\Jr$ & $\Jc$ \\
\midrule
42   & 42.1 & \textcolor{blue}{0.0} & 29.6 & \textcolor{blue}{6.7}  & 24.6 & \textcolor{blue}{16.1} \\
123  & 41.2 & \textcolor{blue}{0.0} & 39.2 & \textcolor{blue}{20.4} & 33.6 & \textcolor{blue}{22.4} \\
456  & 36.6 & \textcolor{blue}{0.0} & 24.2 & 44.4                   & 52.5 & 32.6 \\
789  & 41.5 & \textcolor{blue}{0.0} & 34.8 & \textcolor{blue}{11.2} & 15.8 & \textcolor{blue}{24.1} \\
1024 & 61.2 & \textcolor{blue}{0.0} & 39.9 & \textcolor{blue}{21.4} & 32.6 & \textcolor{blue}{17.4} \\
\midrule
Mean $\pm$ std & 44.5$\pm$8.5 & 0.0$\pm$0.0 & 33.5$\pm$6.0 & 20.8$\pm$13.0 & 31.8$\pm$12.2 & 22.5$\pm$5.9 \\
Within budget  & & 5/5 & & 4/5 & & 4/5 \\
\bottomrule
\end{tabular}
\end{table}

\paragraph{VLM-free and PID baselines on L2.}
At the final epoch, matched PPOLag obtains $\Jr{=}0.13\pm0.05$ and
$\Jc{=}20.9\pm5.5$. CPO obtains $\Jr{=}0.16\pm0.05$ and
$\Jc{=}23.0\pm8.4$, while CPPOPID obtains $\Jr{=}0.24\pm0.22$ and
$\Jc{=}22.8\pm6.7$. These are environment returns because the baselines receive
no VLM reward. All three methods satisfy the budget on average by converging to
the degenerate behaviour defined above, with returns close to zero.

The cost ordering depends on the aggregation window. Over the last ten epochs,
mean cost is $41.4$ for PPOLag, $26.5$ for CPO and $22.3$ for CPPOPID.
Calibrated VLMPPOLag$+$Conf obtains $22.5$ over the same window and $22.4$ at
the final epoch. It has lower cost than the unmatched PPOLag and CPO references
under both aggregations, but remains within the seed variation of the matched
baselines. We therefore do not claim a ranking among these methods. The
unmatched PPOLag also differs in its update schedule, initial multiplier and
seed set, so its higher cost cannot be assigned to one factor.

\paragraph{Evaluation on L2.}
\Cref{tab:phaseB-holdout} evaluates each run for $20$ deterministic episodes
without VLM reward shaping. Mean environment return ranges from $-0.32$ to
$0.09$, and no method makes substantive forward progress. Calibrated
VLMPPOLag$+$Conf has an $8.0\%$ catastrophe rate and an $18.0\%$ violation
rate. Matched PPOLag records $2.5\%$ and $17.5\%$, while CPO records a lower
mean cost than VLMPPOLag$+$Conf. These lower costs occur with negligible
movement. CPPOPID has higher mean cost than matched PPOLag, but the seed-level
bootstrap $95\%$ interval for the difference is $[-8.2,33.8]$. The interval
spans zero, and their ordering changes with the training aggregation window.
None of these comparisons establishes a performance ranking. Because no method
solves the task and FormulaOne cost depends on distance travelled, we do not
interpret the evaluation cost differences as evidence of safer control.

\begin{table}[h]
\centering
\small
\caption{FormulaOne-L2 evaluation over $20$ deterministic episodes per run
using seeds $10000$--$10019$. All policies are evaluated without VLM reward
shaping, so $\Jr$ is the environment return. The mean rows average the per-run
values. No method retains substantive task return, with arm means ranging from
$-0.32$ to $0.09$. Low cost therefore reflects negligible forward progress and
does not establish safer driving. Blue entries satisfy $\Jc{\le}\dlim{=}25$.}
\label{tab:phaseB-holdout}
\setlength{\tabcolsep}{4pt}
\begin{tabular}{@{}llcccc@{}}
\toprule
\textbf{Method} & \textbf{Seed} & $\Jr$ & \textbf{Mean cost} & \textbf{Viol\%} & \textbf{Cat\%} \\
\midrule
\multirow{5}{*}{VLMPPOLag+Conf (family~A)} & 42   & $0.15$  & $26.95$                  & 15.0 & 10.0 \\
                                 & 123  & $0.04$  & $53.30$                  & 20.0 & 15.0 \\
                                 & 456  & $-0.03$ & $26.25$                  & 30.0 & 10.0 \\
                                 & 789  & $0.02$  & \textcolor{blue}{$9.90$} & 10.0 &  5.0 \\
                                 & 1024 & $0.17$  & \textcolor{blue}{$8.30$} & 15.0 &  0.0 \\
\cmidrule(lr){2-6}
                                 & \textbf{mean} & $0.07$ & $24.94$ & 18.0 & 8.0 \\
\midrule
\multirow{3}{*}{CPPOPID}         & 42   & $0.21$  & \textcolor{blue}{$4.95$} & 10.0 &  0.0 \\
                                 & 123  & $0.02$  & $46.95$                  & 25.0 & 15.0 \\
                                 & 456  & $-0.36$ & $34.95$                  & 30.0 & 10.0 \\
\cmidrule(lr){2-6}
                                 & \textbf{mean} & $-0.05$ & $28.95$ & 21.7 & 8.3 \\
\midrule
\multirow{2}{*}{PPOLag (matched)} & 789  & $0.18$ & \textcolor{blue}{$14.40$} & 20.0 & 5.0 \\
                                 & 1024 & $0.01$  & \textcolor{blue}{$11.90$} & 15.0 & 0.0 \\
\cmidrule(lr){2-6}
                                 & \textbf{mean} & $0.09$ & $13.15$ & 17.5 & 2.5 \\
\midrule
\multirow{2}{*}{CPO (no VLM)}    & 789  & $-0.44$ & $35.10$                  & 35.0 & 10.0 \\
                                 & 1024 & $-0.20$ & \textcolor{blue}{$0.10$} &  0.0 &  0.0 \\
\cmidrule(lr){2-6}
                                 & \textbf{mean} & $-0.32$ & $17.60$ & 17.5 & 5.0 \\
\bottomrule
\end{tabular}
\end{table}

\section{Confidence-Gate Diagnostics and Calibration Ablation}
\label{app:gate-calibration}

The confidence gate scales the CLIP reward by
$\kappa=|2\sigma(s(m_{\mathrm{pos}}-m_{\mathrm{neg}}-c))-1|$. The parameters
$s$ and $c$ control the sigmoid steepness and centre. Since the VLM cost signal
is disabled in every experiment, $\kappa$ affects training only through reward
shaping. We first examine how the prior-symmetric setting maps CLIP margins to
gate values, then describe the calibration estimator. We next test whether
$\kappa$ identifies current contact or rises before future contact. The final
comparison assesses sensitivity to the calibrated setting on FormulaOne-L2.

\subsection{CLIP margin and gate values at the prior-symmetric setting}
\label{app:gate-diagnosis}

We sample $200$ frames from each of $12$ evaluation videos, giving $2{,}400$
frames. FormulaOne contributes one video per level, while MetaDrive contributes
three per difficulty. For every frame, we compute the prompt-group margin
$m_{\mathrm{pos}}-m_{\mathrm{neg}}$ and $\kappa$ at $c{=}0$ with
$s\in\{10,50,100\}$.

The margin is positive on all sampled frames, with a minimum of $0.004$ on
FormulaOne-L0. Median margins range from $0.011$ on FormulaOne-L0 to $0.046$
on MetaDrive Easy. The FormulaOne medians increase from $0.011$ on L0 to
$0.018$ on L1 and $0.025$ on L2. Across MetaDrive, the mean cell medians are
$0.039$, $0.037$ and $0.036$ on Easy, Medium and Hard. These frames have no
danger labels and the environments use different prompts, so this comparison
cannot separate visual content from prompt-dependent offsets. The calibrated
centre $\hat c$ is intended to absorb such offsets.

At the prior-symmetric setting $(s,c){=}(100,0)$, median $\kappa$ is $0.50$,
$0.72$ and $0.84$ on FormulaOne-L0, L1 and L2. Across the nine MetaDrive
cells, it ranges from $0.86$ to $0.98$. On Hard, $93\%$ of frames have
$\kappa{>}0.9$, although none exceeds $0.99$. The gate is therefore nearly
open on most sampled Hard frames. With $s{=}10$, every cell has median
$\kappa{\le}0.23$. With $s{=}50$, the median is $0.27$ on FormulaOne-L0 and
ranges from $0.57$ to $0.82$ on MetaDrive.

On FormulaOne-L0, the prior-symmetric gate reduces the augmented return from
$64.3$ without gating to $45.3$. The five calibrated runs obtain $44.5$.
Because the return consists mainly of the shaping bonus, this reduction does
not establish lower task performance. Calibration is evaluated only on
FormulaOne-L2. The nominal MetaDrive calibration runs use the prior-symmetric
gate and serve only to estimate evaluation variability.

\subsection{Calibration estimator}
\label{app:gate-recalibration}

The estimator in \eqnref{eq:mle-sc} uses a random-policy buffer
$\mathcal{B}$ collected before training. It sets $\hat c$ to the median margin
and chooses $\hat s$ so that a frame one interquartile range from the median
has gate value $\kappa^\star$.
\[
\hat c = \mathrm{median}(\mathcal{B}), \qquad
\hat s = \frac{1}{\mathrm{IQR}(\mathcal{B})}
\log\!\frac{1+\kappa^\star}{1-\kappa^\star}.
\]
Our calibrated FormulaOne runs use $500$ frames and
$\kappa^\star{=}0.5$. The gate is nearly closed at the median and reaches
$\kappa{=}0.5$ one interquartile range above or below it. It therefore assigns
larger weights to unusually high and unusually low margins. Across the five
calibrated runs, the recorded estimates range from $109$ to $562$ for
$\hat s$ and from $0.014$ to $0.029$ for $\hat c$.

Applying the estimator to the diagnostic frames themselves reduces the median
$\kappa$ from $0.86$--$0.98$ to $0.05$--$0.30$ on MetaDrive and from $0.50$
to $0.16$ on FormulaOne-L0. These values describe an in-sample diagnostic.
Only the FormulaOne experiments apply calibration during training.

\paragraph{Interpretation of the estimator.}
\label{app:gate-mle-derivation}
The centre is the median of an unlabelled margin buffer, which is robust to
outlying values. It is not a maximum-likelihood estimate of a logistic
location and is not Bayes-optimal without labels. The scale follows from
$2\sigma(x)-1=\tanh(x/2)$. Setting
$m-\hat c=\mathrm{IQR}(\mathcal{B})$ and $\kappa=\kappa^\star$ gives
\[
\hat s\,\mathrm{IQR}(\mathcal{B})
 = \log\!\frac{1+\kappa^\star}{1-\kappa^\star},
\]
which yields \eqnref{eq:mle-sc}. Symmetry gives the same gate value one IQR
below the median.

\begin{table}[t]
\centering
\caption{Per-cell median CLIP margin and median $\kappa$ over $200$ frames per
cell, at the prior-symmetric setting $(s,c){=}(100,0)$ and after calibration by
\eqnref{eq:mle-sc} fitted in-sample on the same $200$ frames. After calibration
every cell yields $\kappa{=}0.5$ at $\pm1$~IQR by construction. Values below
$0.1$ mean that the gate is nearly closed on typical frames, and values above
$0.9$ mean that it is nearly open. FormulaOne cells use one video each (training
seed $42$).}
\label{tab:gate-medians}
\small
\begin{tabular}{lccc}
\toprule
Cell & median margin & median $\kappa$ ($s{=}100$, $c{=}0$) & median $\kappa$ (calibrated) \\
\midrule
F1-L0            & 0.011 & 0.50 & 0.16 \\
F1-L1            & 0.018 & 0.72 & 0.26 \\
F1-L2            & 0.025 & 0.84 & 0.28 \\
MD-Easy (s42)    & 0.046 & 0.98 & 0.08 \\
MD-Easy (s123)   & 0.046 & 0.98 & 0.30 \\
MD-Easy (s456)   & 0.026 & 0.86 & 0.05 \\
MD-Medium (s42)  & 0.037 & 0.95 & 0.23 \\
MD-Medium (s123) & 0.044 & 0.98 & 0.14 \\
MD-Medium (s2024)& 0.029 & 0.89 & 0.27 \\
MD-Hard (s42)    & 0.034 & 0.93 & 0.26 \\
MD-Hard (s123)   & 0.046 & 0.98 & 0.06 \\
MD-Hard (s456)   & 0.030 & 0.90 & 0.28 \\
\bottomrule
\end{tabular}
\end{table}

\subsection{ROC of $\kappa$ against current-step cost}
\label{app:gate-roc}

We test whether $\kappa$ ranks frames with non-zero simulator cost above
zero-cost frames. We use stochastic rollouts from five calibrated
VLMPPOLag$+$Conf policies on FormulaOne-L1 and L2. Each level contains $25$
episodes and $25{,}000$ frames. Non-zero cost occurs on $83$ L1 frames and
$269$ L2 frames, giving prevalences of $0.33\%$ and $1.08\%$. For this
diagnostic, the calibrated parameters are fitted to the same frames used for
evaluation. They are $\hat s{=}143$ and $\hat c{=}0.0223$ on L1, and
$\hat s{=}184$ and $\hat c{=}0.0195$ on L2.

The calibrated gate reaches an AUC of $0.82$ on L1 and $0.78$ on L2. The
prior-symmetric gate reaches $0.13$ and $0.34$, which indicates an inverted
ranking. Cost-positive frames have smaller margins than zero-cost frames.
Their median margins are $0.0115$ and $0.0225$ on L1, and $0.0144$ and
$0.0195$ on L2. Calibration folds this low-margin tail upward through the
absolute value in $\kappa$. Every cost-positive L1 frame and $70.6\%$ of the
L2 frames lie below $\hat c$. On L1, the negated margin reaches an AUC of
$0.87$, exceeding the calibrated gate. The high calibrated AUC therefore
shows that contact frames occupy a tail of the margin distribution. It does
not establish a directional danger score or out-of-sample prediction.

At a threshold of $0.2$, calibrated-gate recall is $0.99$ on L1 and $0.87$ on
L2. At $0.3$, recall is $0.90$ and $0.81$. Precision remains below $0.02$ on
L1 and $0.13$ on L2. The gate therefore acts as a soft weighting mechanism,
not a reliable contact classifier. This analysis concerns current contact and
does not test anticipation.

\begin{figure}[h]
  \centering
  \includegraphics[width=\textwidth]{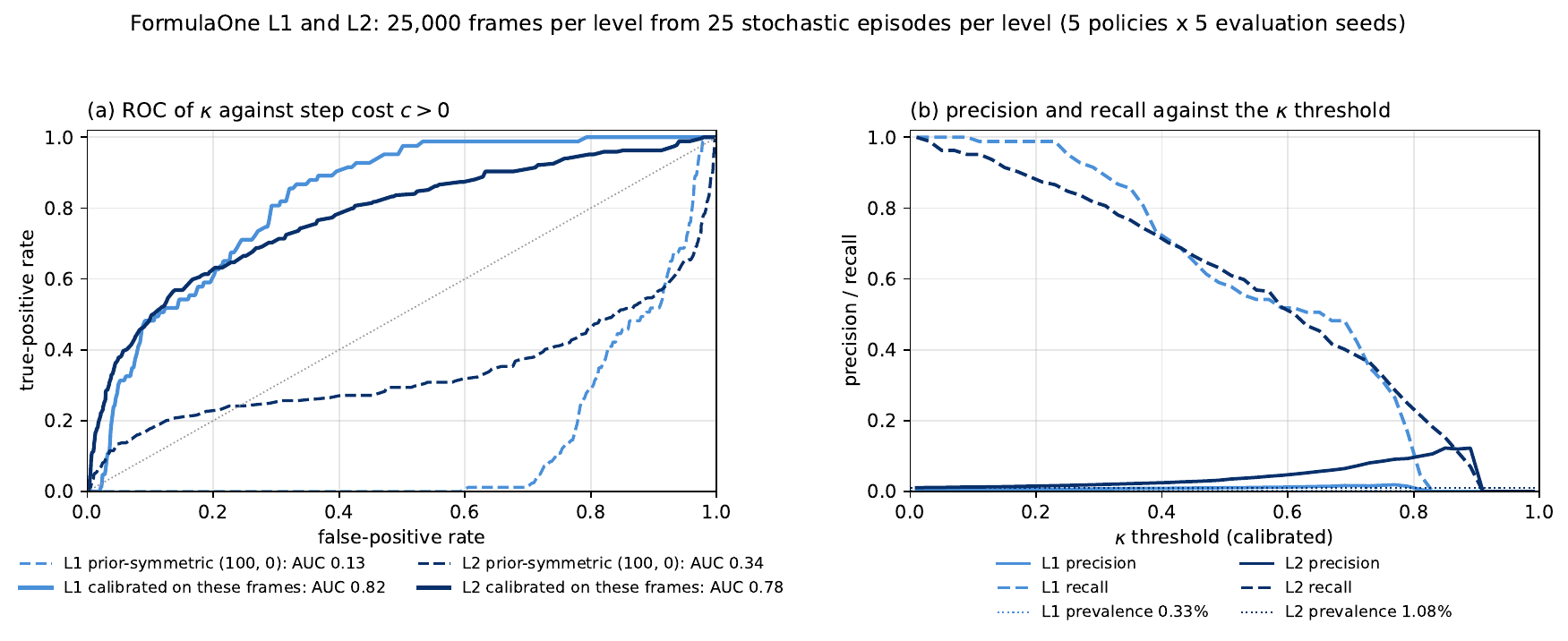}
  \caption{Confidence-gate ROC against current-step FormulaOne cost. Each level
    contains $25$ stochastic episodes and $25{,}000$ frames. Positive frames
    have non-zero step cost, with prevalence $0.33\%$ on L1 and $1.08\%$ on L2.
    Panel (a) compares the prior-symmetric gate, whose AUC is $0.13$ on L1 and
    $0.34$ on L2, with the in-sample calibrated gate, whose AUC is $0.82$ and
    $0.78$. Panel (b) reports precision and recall across calibrated-gate
    thresholds. Dotted horizontal lines mark prevalence.}
  \label{fig:kappa-groundtruth}
\end{figure}

\subsection{Per-step lead-lag analysis of the CLIP signal}
\label{app:horizon-roc}

\S\ref{app:gate-roc} evaluates frames during contact. Here we test whether a
CLIP signal rises before contact. We evaluate deterministic rollouts on
FormulaOne-L2 and score every frame with CLIP ViT-B/32. The recorded signals
are $\kappa$, $\cvlm$, $\rvlm$ and the prompt-group margin. The gate uses
$(s,c){=}(100,0)$ for this analysis. We compare an unmatched VLM-free PPOLag
arm with calibrated VLMPPOLag$+$Conf. Each arm contains three training seeds
and $60$ episodes per seed, giving $180{,}000$ PPOLag steps and $179{,}672$
VLMPPOLag$+$Conf steps.

For each horizon $K\in\{1,3,5,10,20,40\}$, a non-contact step is positive when
contact occurs within the next $K$ steps. We evaluate both raw signals and
signals standardized within each episode. An AUC above $0.5$ indicates that a
signal rises before contact. Step index provides a control for contacts that
occur late in an episode. We use $95\%$ episode-level bootstrap intervals from
$2{,}000$ resamples. Evidence of anticipation requires standardized $\kappa$
to reach an AUC of at least $0.60$ at $K{=}20$ or $40$, with a confidence
interval above $0.5$ and an AUC above the step-index control.

\begin{table}[h]
\centering
\small
\caption{Per-step lead-lag ROC on FormulaOne-L2. Each entry spans the pooled
AUCs over both arms and six horizons. An AUC above $0.5$ means that the signal
is higher on steps followed by contact. The step-index control is identical
for the raw and standardized signals.}
\label{tab:horizon-roc}
\begin{tabular}{@{}lcc@{}}
\toprule
\textbf{Signal} & \textbf{Raw} & \textbf{Within-episode standardized} \\
\midrule
$\kappa$ & 0.172--0.320 & 0.371--0.497 \\
$\cvlm$  & 0.199--0.452 & 0.348--0.462 \\
$\rvlm$  & 0.149--0.398 & 0.322--0.465 \\
Margin   & 0.173--0.320 & 0.356--0.493 \\
\midrule
Step index (control) & \multicolumn{2}{c}{0.512--0.614} \\
\bottomrule
\end{tabular}
\end{table}

All $96$ pooled CLIP AUC point estimates lie between $0.149$ and $0.497$,
below both chance and the step-index control of $0.512$--$0.614$. None of the
signals therefore rises before contact at horizons from one to forty steps.
The standardized estimates lie closer to chance than the raw estimates,
indicating that between-episode variation explains part of the raw separation.

Reading the AUCs in reverse shows that lower signal values are sometimes
associated with upcoming contact. Sixty-six of the $96$ reversed pooled
estimates exceed the step-index control. However, reversed standardized
$\kappa$ reaches only $0.503$--$0.519$ for VLMPPOLag$+$Conf. For PPOLag, it
reaches $0.581$ at $K{=}20$ and $0.563$ at $K{=}40$. Neither arm meets the
predefined threshold of $0.60$. Across individual runs, $39$ of $288$ CLIP
estimates exceed $0.5$, but none has a confidence interval entirely above
$0.5$. The pooled results therefore provide no evidence that a CLIP signal
anticipates contact.

The Spearman correlation between $\cvlm(t)$ and cost at $t{+}k$ is nearly
constant for lags from $-10$ to $20$. It ranges from $-0.068$ to $-0.066$ for
VLMPPOLag$+$Conf and from $-0.359$ to $-0.349$ for PPOLag. This flat profile
provides no evidence of a temporal lead.

\textbf{Variability of $\cvlm$.} Over the $179{,}672$ rollout steps of the
calibrated $+$Conf arm, $\cvlm$ has mean $0.6257$, standard deviation $0.0065$
(a coefficient of variation of $1.0\%$) and minimum $0.5940$. The per-step score
therefore never comes near $\tau{=}0.5$, so $(\meancvlm-\tau)$ keeps one sign
and the multiplier term of \eqnref{eq:vlm-lagrange} adds a nearly constant,
negligible upward bias to the bracket, not a state-dependent signal. Note that these are
steps of the $180$ diagnostic episodes described above, not of the evaluation
protocol of \S\ref{sec:setup}.

\subsection{Calibration comparison on FormulaOne-L2}
\label{app:gate-prereg}

The calibrated VLMPPOLag$+$Conf result in \Cref{tab:main_results} uses the
estimator in \eqnref{eq:mle-sc} with $\kappa^\star{=}0.5$, a $500$-frame
random-policy buffer and $k_{\text{clip}}{=}4$. We compare it with the
prior-symmetric setting on the shared seeds $\{42,123,456\}$. Both settings use
$40$ update iterations, $\lambda_0{=}0.001$, reward weight $0.1$ and no VLM
cost signal. The prior-symmetric runs score every frame, so the comparison
changes both the gate parameters and the CLIP period. It does not isolate
calibration. Unless noted otherwise, we report last-ten-epoch means.

On the shared seeds, calibration changes mean cost from $29.4$ to $23.7$.
The seed-level differences are $-5.5$, $-15.8$ and $+4.3$. A paired $t$-test
gives $p{=}0.43$, while the exact one-sided sign-flip test gives $p{=}0.25$.
At the final epoch, all three cost differences are negative, but the paired
$t$-test gives $p{=}0.08$ and the sign-flip test gives $p{=}0.125$, the
smallest possible value with three pairs. Augmented return also decreases by
$11.2$ on average, with $p{=}0.33$ under the paired $t$-test. The gate value
was not logged per seed, so these changes cannot be linked to stronger
attenuation in individual runs. The results do not establish a difference
between the settings.

The calibrated family records an $8.0\%$ catastrophe rate and an $18.0\%$
violation rate over $20$ deterministic evaluation episodes per seed. The
prior-symmetric runs were not evaluated under this protocol. In the
$50$-episode evaluation, mean costs are $36.77$ for the three prior-symmetric
runs and $22.06$ for the five calibrated runs. These means use different seed
sets and do not provide a paired comparison. The controlled comparison is
therefore limited to training, where gate calibration remains confounded with
the CLIP period. Since $\Jr$ contains the shaping bonus, it does not establish
a task-return difference.

\begin{table}[h]
\centering
\caption{FormulaOne-L2 comparison between the prior-symmetric gate with
$k_{\text{clip}}{=}1$ and the calibrated gate with $k_{\text{clip}}{=}4$.
All runs train for $10^6$ steps. Values are last-ten-epoch means, and $\Jr$ is
the augmented return. Differences are calibrated minus prior-symmetric and are
computed before rounding. Summary rows cover the three shared seeds and all
five calibrated seeds. Blue entries satisfy $\Jc{\le}\dlim{=}25$.}
\label{tab:gate-calib-prereg}
\small
\setlength{\tabcolsep}{4pt}
\begin{tabular}{@{}lcccccc@{}}
\toprule
& \multicolumn{2}{c}{\textbf{prior-symmetric $(100,0)$}}
& \multicolumn{2}{c}{\textbf{calibrated $(\hat s,\hat c)$}}
& \multicolumn{2}{c}{\textbf{$\Delta$ (calibrated $-$ prior)}} \\
\cmidrule(lr){2-3}\cmidrule(lr){4-5}\cmidrule(lr){6-7}
\textbf{Seed} & $\Jr$ & $\Jc$ & $\Jr$ & $\Jc$ & $\Delta\Jr$ & $\Delta\Jc$ \\
\midrule
42  & 48.0 & \textcolor{blue}{21.6} & 24.6 & \textcolor{blue}{16.1} & $-23.4$ & $-5.5$  \\
123 & 49.8 & 38.2                   & 33.6 & \textcolor{blue}{22.4} & $-16.2$ & $-15.8$ \\
456 & 46.5 & 28.4                   & 52.5 & 32.6                   & $+6.0$  & $+4.3$  \\
\midrule
Mean $\pm$ std ($n{=}3$) & 48.1$\pm$1.3 & 29.4$\pm$6.8 & 36.9$\pm$11.7 & \textcolor{blue}{23.7$\pm$6.8} & $-11.2$ & $-5.7$ \\
Within budget & & 1/3 & & 2/3 & & \\
\midrule
Mean $\pm$ std ($n{=}5$) & & & 31.8$\pm$12.2 & \textcolor{blue}{22.5$\pm$5.9} & & \\
Within budget & & & & 4/5 & & \\
\bottomrule
\end{tabular}
\end{table}

\section{Extended Bullet Safety-Gym Results}
\label{app:results-bullet}

\subsection{Per-seed evaluation}
\label{app:bullet-perseed}

We evaluate PPOLag and VLMPPOLag$+$Conf on SafetyCarReach-v0 after one and two
million training steps. Each method uses seeds $\{42,123,456\}$ at both
training horizons. Every run is evaluated for $200$ episodes using
deterministic actions. Bullet Safety-Gym does not reproduce reset states from
the supplied evaluation seeds, so episodes are independent random draws and
are not matched across runs.

Pooling both training horizons, PPOLag has a catastrophe rate of $13.2\%$, a
violation rate of $20.8\%$ and a mean cost of $36.1$.
VLMPPOLag$+$Conf obtains $12.7\%$, $19.8\%$ and $34.7$. The seed-matched
direction is inconsistent. Four of six pairs favour VLMPPOLag$+$Conf and two
favour PPOLag for both catastrophe rate and mean cost. The seed-level
confidence interval for the catastrophe-rate difference is
$[-3.1,1.9]$ percentage points. We therefore find no detectable improvement
on Bullet Safety-Gym.

\begin{table}[h]
\centering
\small
\caption{Per-run SafetyCarReach-v0 evaluation over $200$ episodes with
deterministic actions. Catastrophe and violation denote episodic cost above
$4\dlim$ and $\dlim$. Pooled rows average the six runs of each method. The
environment does not reproduce reset states from evaluation seeds, so episodes
are not matched across runs.}
\label{tab:bullet-perseed}
\setlength{\tabcolsep}{6pt}
\begin{tabular}{@{}llrrrr@{}}
\toprule
\textbf{Method} & \textbf{Training} & \textbf{Seed} & \textbf{Mean cost}
& \textbf{Viol.\,\%} & \textbf{Cat.\,\%} \\
\midrule
PPOLag                 & 1M & 42  & 31.5 & 18.0 & 11.5 \\
PPOLag                 & 1M & 123 & 39.9 & 19.5 & 14.0 \\
PPOLag                 & 1M & 456 & 42.5 & 23.0 & 16.0 \\
VLMPPOLag$+$Conf       & 1M & 42  & 37.6 & 21.5 & 14.5 \\
VLMPPOLag$+$Conf       & 1M & 123 & 32.5 & 21.0 & 12.0 \\
VLMPPOLag$+$Conf       & 1M & 456 & 40.9 & 23.5 & 15.5 \\
\midrule
PPOLag                 & 2M & 42  & 33.6 & 21.5 & 12.0 \\
PPOLag                 & 2M & 123 & 29.1 & 18.0 & 11.0 \\
PPOLag                 & 2M & 456 & 40.0 & 24.5 & 14.5 \\
VLMPPOLag$+$Conf       & 2M & 42  & 30.5 & 16.5 & 10.5 \\
VLMPPOLag$+$Conf       & 2M & 123 & 40.5 & 19.0 & 15.0 \\
VLMPPOLag$+$Conf       & 2M & 456 & 26.0 & 17.0 & 8.5 \\
\midrule
\textbf{PPOLag pooled}           & 1M and 2M & all & 36.1 & 20.8 & 13.2 \\
\textbf{VLMPPOLag$+$Conf pooled} & 1M and 2M & all & 34.7 & 19.8 & 12.7 \\
\bottomrule
\end{tabular}
\end{table}

\subsection{Training curves}
\label{app:bullet-curves}

\begin{figure}[h]
  \centering
  \begin{subfigure}{0.99\linewidth}
    \centering
    \includegraphics[width=\linewidth]{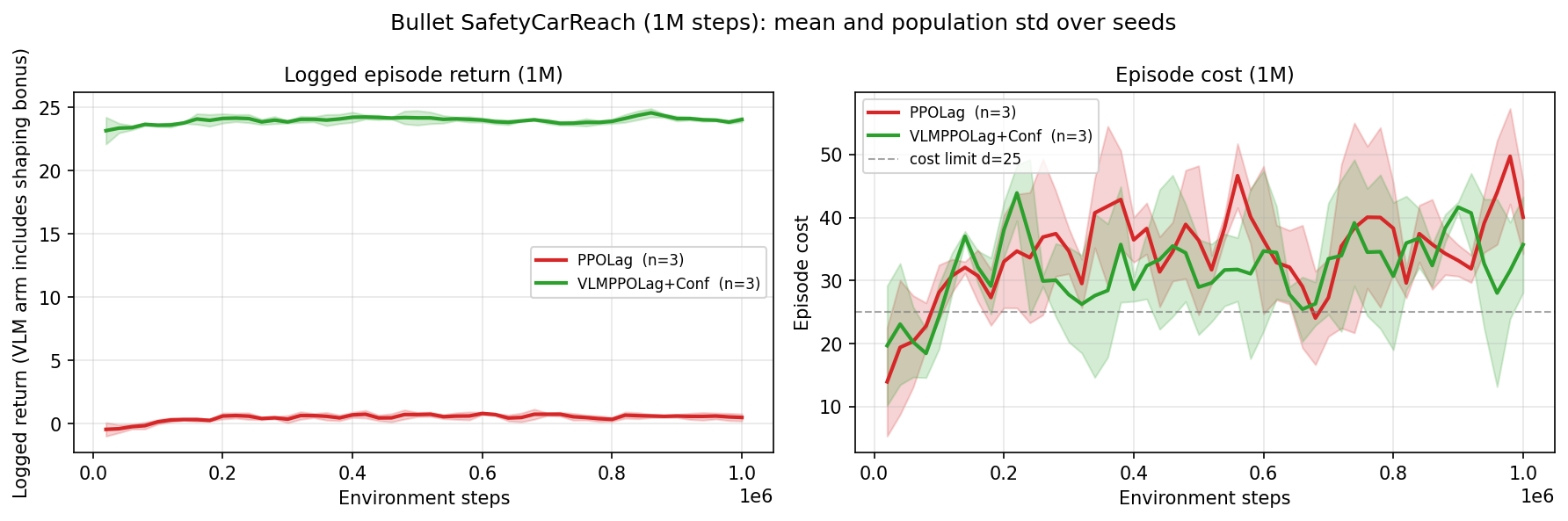}
    \caption{One million training steps over $50$ epochs.}
    \label{fig:appendix-curves-bullet-1m}
  \end{subfigure}\\[0.5em]
  \begin{subfigure}{0.99\linewidth}
    \centering
    \includegraphics[width=\linewidth]{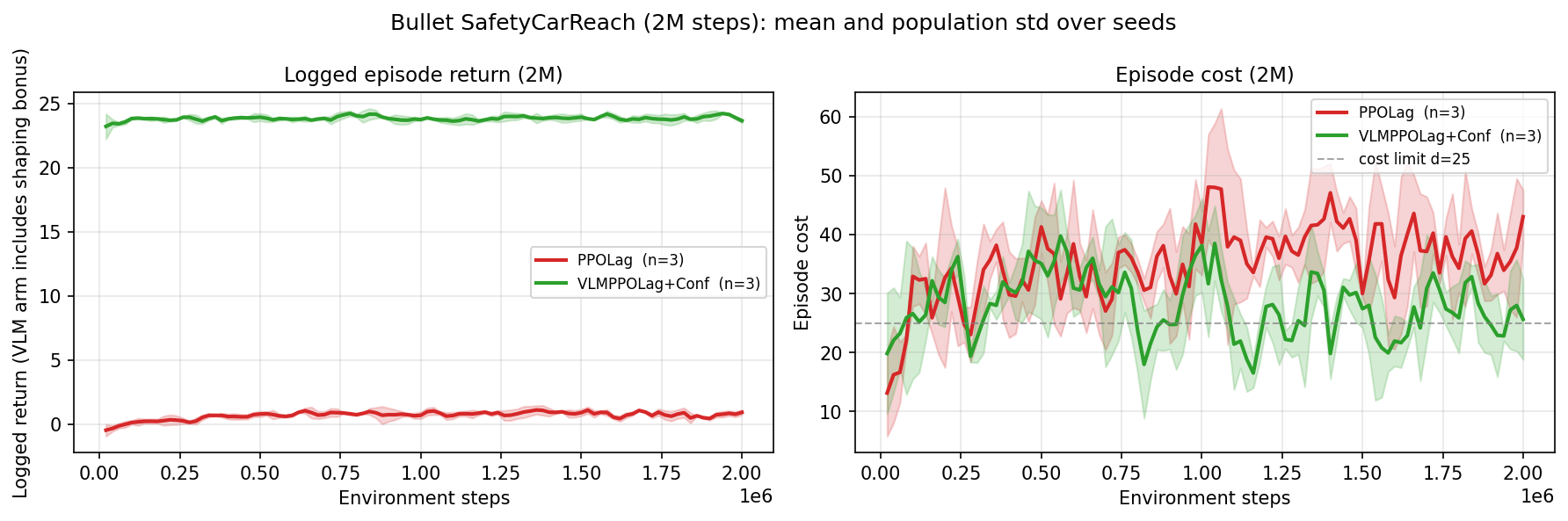}
    \caption{Two million training steps over $100$ epochs.}
    \label{fig:appendix-curves-bullet-2m}
  \end{subfigure}
  \caption{SafetyCarReach-v0 training curves over seeds $\{42,123,456\}$.
    The left panels report training return, and the right panels report
    environment cost with limit $\dlim{=}25$. Lines show the seed mean and
    shading shows one population standard deviation. Returns are not comparable
    because VLMPPOLag$+$Conf includes the gated CLIP bonus. Although its
    final-epoch cost is lower after two million steps, the $200$-episode
    evaluation shows no detectable catastrophe-rate difference.}
  \label{fig:appendix-curves-bullet}
\end{figure}

\Cref{fig:appendix-curves-bullet} shows training at one and two million steps.
The return curves are not comparable across methods because
VLMPPOLag$+$Conf includes the gated CLIP reward bonus. With an unweighted mean
VLM reward of $0.617$ and $500$-step episodes, the bonus is bounded by about
$31$ return units per episode. At the final epoch, VLMPPOLag$+$Conf records
returns of $24.0\pm0.2$ at one million steps and $23.7\pm0.2$ at two million.
PPOLag records $0.5\pm0.3$ and $0.9\pm0.2$.

Training costs are comparable because neither method adds a VLM term to the
environment cost. At two million steps, final-epoch mean cost is
$25.6\pm6.8$ for VLMPPOLag$+$Conf and $43.0\pm4.5$ for PPOLag. At one million
steps, the corresponding values are $35.7\pm7.6$ and $40.1\pm5.8$. This
training difference does not persist in evaluation. At one million steps, the
evaluation catastrophe rates are $14.0\%$ and $13.8\%$. At two million steps,
they are $11.3\%$ and $12.5\%$. Training uses stochastic actions, while
evaluation uses deterministic actions, so lower final-epoch training cost does
not imply lower evaluation risk.

\section{Extended MetaDrive Results}
\label{app:results-md}

\subsection{Per-seed evaluation}
\label{app:md-perseed}

We evaluate each policy for $50$ deterministic episodes using fixed traffic
and a fresh simulator instance for every episode. Evaluation seeds are mapped
into the training scenario pool, so the results are in-distribution. Easy,
Medium and Hard use three, five and ten training seeds per method.

Variation across training seeds is substantial. The standard deviation of
per-seed catastrophe rate is $4.3$ and $12.7$ percentage points on Easy,
$10.5$ and $8.0$ on Medium, and $12.2$ and $8.9$ on Hard for PPOLag and
VLMPPOLag$+$Conf. We therefore base uncertainty on training seeds, not
individual episodes. On Hard, VLMPPOLag$+$Conf lowers catastrophe rate on nine
of ten matched seeds, violation rate on eight and mean cost on seven. On
Medium, it lowers catastrophe and violation rates on three of five seeds and
mean cost on four. Every Easy seed favours PPOLag on all three metrics.

\begin{table}[h]
\centering
\small
\caption{Per-seed MetaDrive evaluation over $50$ deterministic episodes per
run. Evaluation seeds are mapped into the training scenario pool, so the
results measure in-distribution performance. $\Jc$ is mean episodic cost.
Violation and catastrophe denote episodic cost above $\dlim$ and $4\dlim$.
Pooled rows average all episodes within each method and difficulty.}
\label{tab:md-perseed}
\setlength{\tabcolsep}{5pt}
\begin{tabular}{@{}llrrrr@{}}
\toprule
\textbf{Level} & \textbf{Method} & \textbf{Seed} & $\Jc$ & \textbf{Viol.} & \textbf{Cat.} \\
\midrule
Easy    & PPOLag   &   42 & $126.4$ & $26\%$ & $20\%$ \\
Easy    & PPOLag   &  123 & $66.6$  & $16\%$ & $12\%$ \\
Easy    & PPOLag   &  456 & $49.9$  & $12\%$ & $10\%$ \\
Easy    & VLM+Conf &   42 & $164.3$ & $32\%$ & $28\%$ \\
Easy    & VLM+Conf &  123 & $315.2$ & $56\%$ & $50\%$ \\
Easy    & VLM+Conf &  456 & $104.6$ & $28\%$ & $20\%$ \\
\midrule
\textbf{Easy pooled (PPOLag, 3)}   & & & $81.0$  & $18.0\%$ & $\mathbf{14.0\%}$ \\
\textbf{Easy pooled (VLM+Conf, 3)} & & & $194.7$ & $38.7\%$ & $\mathbf{32.7\%}$ \\
\midrule
Medium  & PPOLag   &   42 & $203.0$ & $42\%$ & $32\%$ \\
Medium  & PPOLag   &  123 & $254.9$ & $44\%$ & $36\%$ \\
Medium  & PPOLag   &  456 & $214.1$ & $32\%$ & $26\%$ \\
Medium  & PPOLag   &  789 & $17.0$  & $8\%$  & $6\%$  \\
Medium  & PPOLag   & 2024 & $193.1$ & $38\%$ & $30\%$ \\
Medium  & VLM+Conf &   42 & $98.4$  & $40\%$ & $24\%$ \\
Medium  & VLM+Conf &  123 & $162.8$ & $34\%$ & $28\%$ \\
Medium  & VLM+Conf &  456 & $59.6$  & $20\%$ & $16\%$ \\
Medium  & VLM+Conf &  789 & $204.0$ & $46\%$ & $40\%$ \\
Medium  & VLM+Conf & 2024 & $124.5$ & $40\%$ & $32\%$ \\
\midrule
\textbf{Medium pooled (PPOLag, 5)}   & & & $176.4$ & $32.8\%$ & $\mathbf{26.0\%}$ \\
\textbf{Medium pooled (VLM+Conf, 5)} & & & $129.9$ & $36.0\%$ & $\mathbf{28.0\%}$ \\
\midrule
Hard    & PPOLag   &   42 & $109.2$ & $34\%$ & $26\%$ \\
Hard    & PPOLag   &  123 & $92.1$  & $30\%$ & $22\%$ \\
Hard    & PPOLag   &  456 & $356.6$ & $72\%$ & $60\%$ \\
Hard    & PPOLag   &  789 & $57.2$  & $16\%$ & $14\%$ \\
Hard    & PPOLag   & 1337 & $189.5$ & $40\%$ & $34\%$ \\
Hard    & PPOLag   & 2024 & $202.8$ & $52\%$ & $44\%$ \\
Hard    & PPOLag   & 2718 & $103.6$ & $34\%$ & $24\%$ \\
Hard    & PPOLag   & 3141 & $103.1$ & $30\%$ & $26\%$ \\
Hard    & PPOLag   & 4242 & $138.7$ & $42\%$ & $34\%$ \\
Hard    & PPOLag   & 5555 & $145.1$ & $42\%$ & $32\%$ \\
Hard    & VLM+Conf &   42 & $131.0$ & $26\%$ & $24\%$ \\
Hard    & VLM+Conf &  123 & $39.5$  & $16\%$ & $10\%$ \\
Hard    & VLM+Conf &  456 & $97.9$  & $20\%$ & $18\%$ \\
Hard    & VLM+Conf &  789 & $243.0$ & $50\%$ & $40\%$ \\
Hard    & VLM+Conf & 1337 & $56.5$  & $16\%$ & $12\%$ \\
Hard    & VLM+Conf & 2024 & $100.1$ & $28\%$ & $24\%$ \\
Hard    & VLM+Conf & 2718 & $128.2$ & $28\%$ & $20\%$ \\
Hard    & VLM+Conf & 3141 & $96.9$  & $38\%$ & $20\%$ \\
Hard    & VLM+Conf & 4242 & $130.3$ & $28\%$ & $20\%$ \\
Hard    & VLM+Conf & 5555 & $27.1$  & $10\%$ & $6\%$  \\
\midrule
\textbf{Hard pooled (PPOLag, 10)}   & & & $149.8$ & $39.2\%$ & $\mathbf{31.6\%}$ \\
\textbf{Hard pooled (VLM+Conf, 10)} & & & $105.1$ & $26.0\%$ & $\mathbf{19.4\%}$ \\
\bottomrule
\end{tabular}
\end{table}

\subsection{Training curves}
\label{app:md-curves}

\begin{figure}[h]
  \centering
  \includegraphics[width=0.99\linewidth]{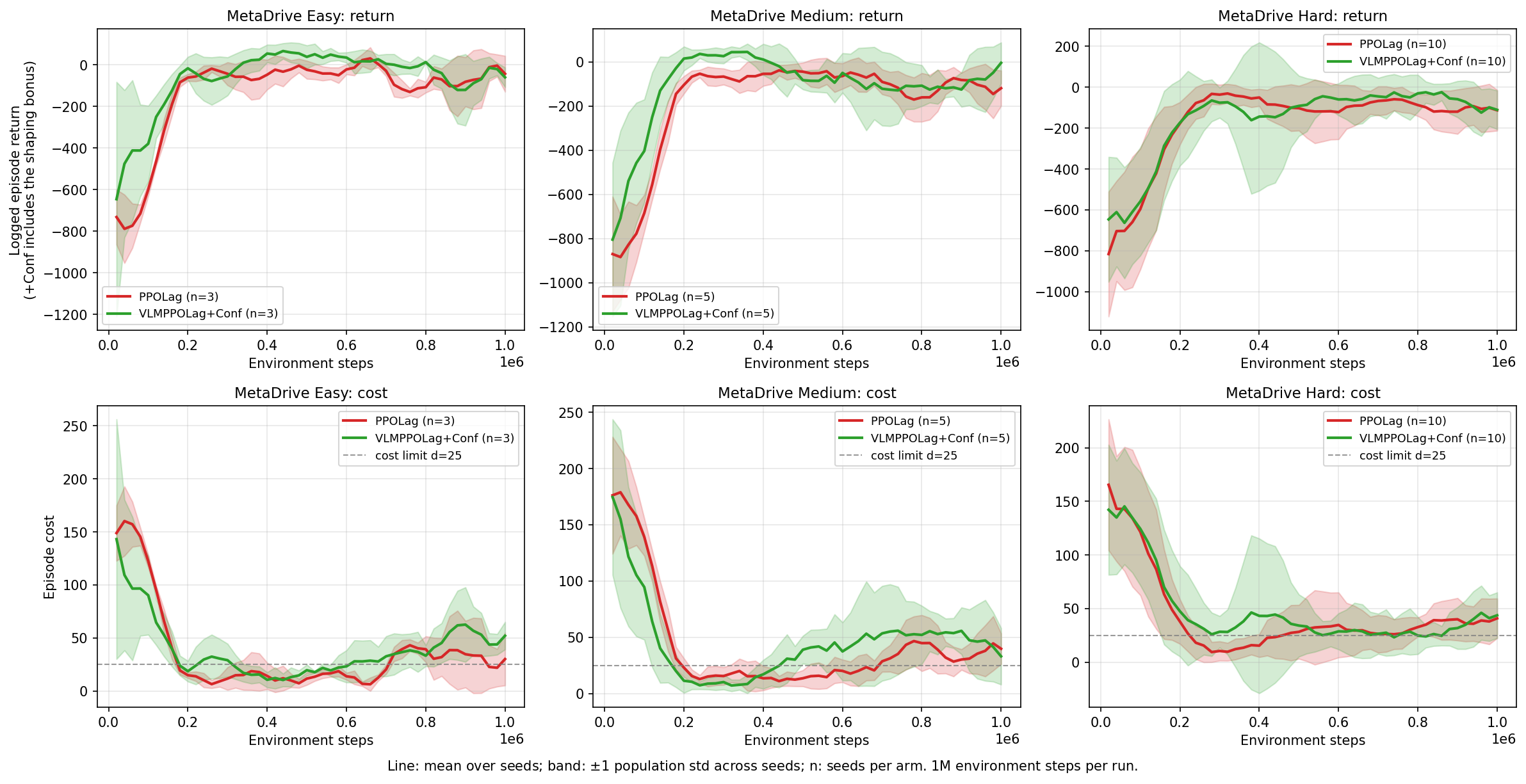}
  \caption{MetaDrive training curves for Easy, Medium and Hard. The top row
    reports augmented training return, and the bottom row reports environment
    cost with limit $\dlim{=}25$. Easy, Medium and Hard use three, five and ten
    seeds per method. Lines show seed means and shading shows one population
    standard deviation. Return curves are not comparable because
    VLMPPOLag$+$Conf includes the gated CLIP reward bonus.}
  \label{fig:appendix-curves-md}
\end{figure}

\Cref{fig:appendix-curves-md} reports the training curves for the runs in
\Cref{tab:md-perseed}. Returns are not comparable across methods because
VLMPPOLag$+$Conf includes the gated CLIP bonus. For one Hard run, the
unweighted mean VLM reward is $0.618$, which bounds the bonus by approximately
$62$ units over a $1000$-step episode. Its augmented return is $-151.5$, while
the matched PPOLag run records an environment return of $-30.9$.

Training costs are comparable because neither method adds a VLM term to the
environment cost. Final-epoch mean costs are $30.2\pm24.8$ for PPOLag and
$52.2\pm12.9$ for VLMPPOLag$+$Conf on Easy, $39.7\pm13.5$ and $33.0\pm25.1$
on Medium, and $40.9\pm18.5$ and $43.8\pm21.3$ on Hard. All means exceed the
budget of $25$. Training uses stochastic actions and random traffic, while
evaluation uses deterministic actions with fixed traffic. Training and
evaluation costs therefore need not follow the same ordering.

\subsection{Evaluation summary}
\label{app:md-fixed-eval}

\begin{figure}[h]
  \centering
  \includegraphics[width=0.99\linewidth]{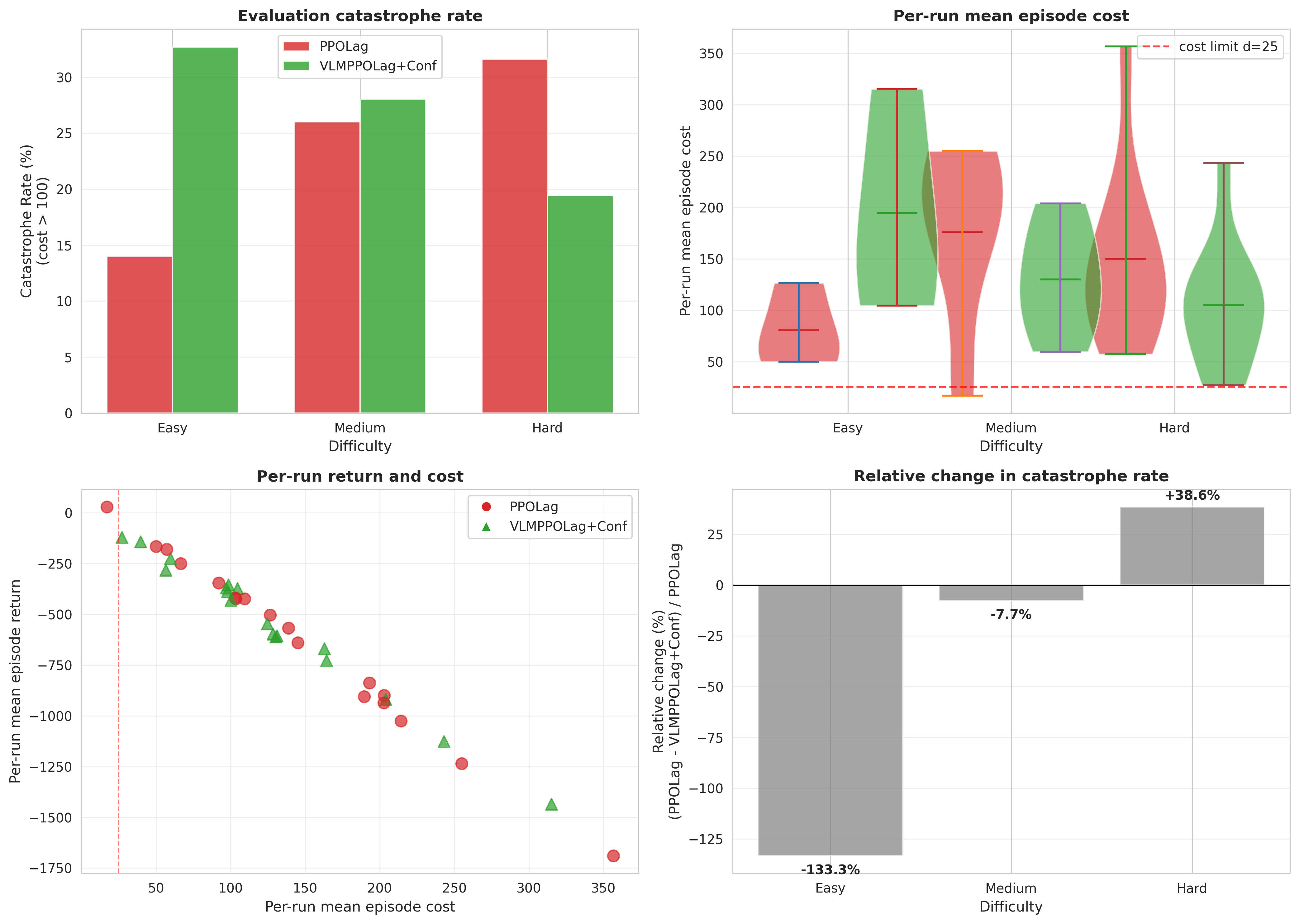}
  \caption{MetaDrive evaluation over $50$ deterministic episodes per run.
    Evaluation scenarios are drawn from the training pool. The top-left panel
    reports catastrophe rate, and the top-right panel shows per-seed mean cost.
    The bottom-left panel compares per-seed mean return and cost. The
    bottom-right panel reports relative catastrophe-rate change, with positive
    values favouring VLMPPOLag$+$Conf. Easy, Medium and Hard use three, five and
    ten seeds per method.}
  \label{fig:appendix-md-fixed-eval}
\end{figure}

\Cref{fig:appendix-md-fixed-eval} summarizes the evaluation results. On Easy,
catastrophe increases from $14.0\%$ to $32.7\%$. Medium changes from $26.0\%$
to $28.0\%$, with a seed-level interval spanning zero. On Hard, catastrophe
decreases from $31.6\%$ to $19.4\%$ and violation decreases from $39.2\%$ to
$26.0\%$.

Per-seed mean return and cost are strongly negatively correlated
($r{=}-0.995$ over $36$ runs), so lower cost may reflect reduced movement.
On Hard, $28.6\%$ of VLMPPOLag$+$Conf episodes have zero cost and return below
$20$, compared with $5.2\%$ for PPOLag. These episodes concentrate in two
training seeds. Passive behaviour may therefore explain part of the Hard
difference. The relative catastrophe-rate changes are $-133.3\%$ on Easy,
$-7.7\%$ on Medium and $38.6\%$ on Hard.

\subsection{Noise floor of the MetaDrive evaluation protocol}
\label{app:md-calib}

A separate set of MetaDrive runs was intended to test gate calibration, but
calibration was not applied in this environment. These runs therefore use the
same prior-symmetric gate as the reported VLMPPOLag$+$Conf runs. Their
comparison cannot estimate a calibration effect. It instead provides an
independent estimate of replication variability.

The two repetitions differ in pooled catastrophe rate by $-7.5$, $-6.8$ and
$8.2$ percentage points on Easy, Medium and Hard. Exact two-sided seed-level
permutation tests give $p{=}0.54$, $0.43$ and $0.29$. Across both repetitions,
seed-level catastrophe rates span $8$--$50\%$ on Easy, $6$--$44\%$ on Medium
and $0$--$48\%$ on Hard. Their population standard deviations are $11.9$,
$11.6$ and $13.1$ percentage points.

The $12.2$ percentage-point Hard difference between VLMPPOLag$+$Conf and
PPOLag is comparable in magnitude to this replication variability. Its
evidence comes from the seed-level interval over ten runs per method, not from
the point estimate alone. The repeated Hard comparison shares only four seeds
and therefore provides a less precise reference.

\begin{table}[h]
\centering
\small
\caption{Replication variability under the prior-symmetric MetaDrive
configuration. Arm A contains the reported VLMPPOLag$+$Conf runs, while Arm B
is an independent repetition with five seeds per difficulty. Rates pool $50$
deterministic episodes per run. $\Delta$ is Arm B minus Arm A. Seed range spans
both arms, and $p$ is an exact two-sided seed-level permutation test of the
catastrophe rate.}
\label{tab:md-noise-floor}
\setlength{\tabcolsep}{4pt}
\begin{tabular}{@{}lccccccc@{}}
\toprule
& \multicolumn{5}{c}{\textbf{Cat.\,\%}} & \multicolumn{2}{c}{\textbf{Viol.\,\%}} \\
\cmidrule(lr){2-6}\cmidrule(lr){7-8}
\textbf{Difficulty} & \textbf{Arm A ($n$)} & \textbf{Arm B ($n$)} & $\Delta$ & \textbf{Seed range} & $p$ & \textbf{Arm A} & \textbf{Arm B} \\
\midrule
Easy   & 32.7\% (3)  & 25.2\% (5) & $-7.5$\,pp & 8--50\% & 0.54 & 38.7\% & 30.4\% \\
Medium & 28.0\% (5)  & 21.2\% (5) & $-6.8$\,pp & 6--44\% & 0.43 & 36.0\% & 26.8\% \\
Hard   & 19.4\% (10) & 27.6\% (5) & $+8.2$\,pp & 0--48\% & 0.29 & 26.0\% & 36.8\% \\
\bottomrule
\end{tabular}
\end{table}

\subsection{Per-difficulty outcomes}
\label{app:md-difficulty}

All intervals below are seed-level bootstrap intervals for the difference
VLMPPOLag$+$Conf minus PPOLag. Evaluation returns exclude the VLM shaping
bonus. Since the VLM cost signal is disabled, the VLM affects training mainly
through reward shaping. The multiplier contribution is negligible
(\S\ref{sec:results-eta2}).

\paragraph{Easy.}
With traffic density $0.1$ and three seeds per method, catastrophe increases
from $14.0\%$ to $32.7\%$. The difference is $18.7$ percentage points with a
bootstrap interval of $[5.3,35.3]$. Violation increases from $18.0\%$ to
$38.7\%$, and mean return decreases from $-307$ to $-845$. Every seed favours
PPOLag on catastrophe rate, violation rate and mean cost. This is an important
adverse result, but the sample is too small for a conventional exact test to
reach $p<0.05$. We have not identified its cause.

\paragraph{Medium.}
With density $0.2$ and five seeds per method, catastrophe changes from
$26.0\%$ to $28.0\%$. The difference is $2.0$ percentage points with an
interval of $[-8.8,14.0]$. Violation changes from $32.8\%$ to $36.0\%$, while
mean return improves from $-794$ to $-543$. The evidence does not establish a
beneficial or harmful effect.

\paragraph{Hard.}
With density $0.35$ and ten seeds per method, catastrophe decreases from
$31.6\%$ to $19.4\%$. The difference is $-12.2$ percentage points with an
interval of $[-21.8,-3.2]$. Violation decreases from $39.2\%$ to $26.0\%$,
and mean return improves from $-653$ to $-468$. Nine of ten matched seeds have
a lower catastrophe rate under VLMPPOLag$+$Conf.

We do not attribute the Hard result to collision anticipation or the
multiplier update. The diagnostic gate medians are $0.90$--$0.98$ on Hard and
$0.86$--$0.98$ across all MetaDrive settings. These high values indicate that
the gate is nearly non-selective in the sampled videos. Reward shaping was not
isolated, and less active behaviour may explain part of the lower cost. The
mechanism therefore remains unresolved.

\section{Cross-Environment Results}
\label{app:cross-env}
\label{app:when-helps}

The results vary across environments. In MetaDrive, VLMPPOLag$+$Conf reduces
catastrophe rate on Hard over ten seeds and shows no detectable difference on
Medium over five seeds. On Easy, catastrophe rate increases for all three
seeds. Its bootstrap interval excludes zero, but the sample is too small for a
conventional exact test to establish significance.

Bullet Safety-Gym shows no detectable difference. Across six runs per method
and $200$ episodes per run, catastrophe rates are $13.2\%$ for PPOLag and
$12.7\%$ for VLMPPOLag$+$Conf. The seed-level interval is
$[-3.1,1.9]$ percentage points. On FormulaOne-L2, no evaluated method completes
a route stage, and cost is related to distance travelled. We therefore do not
interpret its cost differences as evidence of safety.

The MetaDrive point estimates change from $18.7$ percentage points on Easy to
$2.0$ on Medium and $-12.2$ on Hard. These settings differ in both traffic
density and map size, which increases from three to seven blocks. The
experiment does not isolate either factor, so this ordering does not establish
when the method improves safety.


\section{Statistical Methodology}
\label{app:stats}

We report $95\%$ percentile bootstrap intervals
\citep{efron1986bootstrap}. Per-run return and cost intervals use $10{,}000$
episode-level resamples. Rate-difference intervals resample training runs
independently within each method. MetaDrive intervals use training seeds as
clusters with $10{,}000$ resamples. Bullet intervals use the six runs from the
two training horizons. Lead-lag AUC intervals resample episodes $2{,}000$
times. Difference-of-means intervals use $10{,}000$ seed-level resamples
within each method.

FormulaOne comparisons of run-level means use exact one-sided permutation
tests. With three runs per method, the smallest attainable $p$-value is
$1/20=0.05$. Paired gate comparisons use exact sign-flip tests. Where reported,
Welch's $t$-test allows unequal variances between methods. The small number of
training runs limits the precision and power of these tests. We therefore
interpret results using effect sizes and confidence intervals, while avoiding
binary conclusions based only on a significance threshold
\citep{henderson2018matters,agarwal2021deep}.

Training results are reported as the mean and population standard deviation
over seeds. They use the final epoch unless a table states a different
aggregation. Evaluation rates use strict thresholds, with a violation defined
by $\Jc>\dlim$ and a catastrophe defined by $\Jc>4\dlim$.

\section{Reproducibility Checklist}
\label{app:reproducibility}

\paragraph{Artifacts.}
We will release the run configurations, prompt sets, training logs, evaluation
data and experiment metadata upon acceptance. These materials cover every
reported environment and baseline. The computational requirements are given
in App.~\ref{app:compute}.

\paragraph{Training seeds.}
The three-seed FormulaOne grid and all Bullet runs use
$\{42,123,456\}$. The calibrated FormulaOne cells use
$\{42,123,456,789,1024\}$, while the matched FormulaOne-L2 PPOLag control uses
$\{789,1024\}$. MetaDrive Easy uses $\{42,123,456\}$, and Medium adds $789$
and $2024$. Hard uses those five seeds and
$\{1337,2718,3141,4242,5555\}$. The independent MetaDrive repetitions use
$\{42,123,456,789,1024\}$ at every difficulty.

\paragraph{Evaluation conditions.}
MetaDrive uses $50$ deterministic episodes per run with seeds
$10000$--$10049$, fixed traffic and a fresh simulator instance for every
episode. These seeds map into the training scenario pool, so the results are
in-distribution. Bullet uses $200$ episodes per run. Its reset states cannot be
reproduced from supplied seeds, so episodes are random draws. FormulaOne
environment-only evaluation uses seeds $10000$--$10049$, while task metrics
use $30000$--$30049$.

\paragraph{Expected variability.}
Identically configured FormulaOne reruns can diverge substantially. Across
five paired reruns on L1, return differs by as much as $47\%$ over their shared
epochs, and four pairs show more than a twofold cost difference in at least one
epoch. Two independent calibrated FormulaOne-L2 families also differ in
outcome. MetaDrive repetitions show comparable seed and run variation, while
Bullet evaluation cannot reproduce episode initial conditions. Reproduction
should therefore target seed-level means and their reported intervals, not an
individual training trajectory.

\section{Baseline Descriptions}
\label{app:baselines}

\paragraph{Standard reinforcement-learning baselines.}
PPO~\citep{schulman2017proximal} is the unconstrained reference. CPO
\citep{achiam2017constrained} and PPOLag~\citep{ray2019benchmarking} enforce
the cost constraint without a VLM. The unmatched PPOLag baseline uses ten
update iterations and $\lambda_0{=}0$, while the two-seed matched control uses
forty iterations and $\lambda_0{=}10^{-3}$. CPPOPID
\citep{stooke2020responsive} replaces the standard Lagrangian update with a
PID controller. FOCOPS, CUP and P3O provide additional VLM-free constrained
baselines.

\paragraph{Semantic-reward baselines.}
PPO-CLG and CPO-CLG add a rescaled difference between positive and negative
CLIP similarities to the environment reward, following the semantic component
of VLM-RL~\citep{huang2024vlmrl}. They do not reproduce its full hierarchical
reward and do not use a VLM signal in the multiplier. CPO-Coupled instead
normalizes all prompt logits jointly. CPO-Decoupled uses separate positive and
negative prompt groups as defined in \eqnref{eq:decoupled}. These two CPO
variants differ only in the scoring construction, although several runs ended
before the full training horizon. CPO-Decoupled$+$Conf adds confidence gating.

\paragraph{Proposed-method comparisons.}
PPOLag-Decoupled combines PPO-Lagrangian with the decoupled CLIP reward and
sets $\eta_2=0$. It provides the direct comparison for the VLM term in the
multiplier when paired with ungated VLMPPOLag. VLMPPOLag uses the same
decoupled reward with the augmented multiplier update in
\eqnref{eq:vlm-lagrange}. VLMPPOLag$+$Conf adds confidence gating. The
three-seed gated and ungated runs differ only in gating and score every frame.
The five-seed gated runs use the calibrated gate and score every fourth frame,
so they do not isolate gating from the scoring period.

\paragraph{Representation substitutions.}
PPOLag-RND replaces the VLM signal with Random Network Distillation novelty
and removes semantic reward shaping. Qwen2-VL$+$Conf replaces CLIP with
Qwen2-VL-7B and queries the model every eight steps. These substitutions test
alternative signals and are not configuration-matched controls.

\section{Per-seed FormulaOne Results}
\label{app:full-tables}

\Cref{tab:full-results} reports final-epoch training metrics for the
three-seed FormulaOne grid over seeds $\{42,123,456\}$. It also includes
PPOLag-RND on L1 and L2 and Qwen2-VL$+$Conf on L2. The
VLMPPOLag$+$Conf rows use the prior-symmetric gate and score every frame. They
are distinct from the calibrated five-seed results in
\Cref{tab:phaseB-conf,tab:conf-families}. The matched PPOLag, CPPOPID,
CPO-Decoupled$+$Conf and separate $\eta_2{=}0$ runs are reported elsewhere.

\begin{table}[h]
\centering
\tiny
\caption{Per-seed final-epoch FormulaOne training metrics. The PPOLag rows are
the unmatched baseline. VLMPPOLag$+$Conf denotes the three-seed
prior-symmetric runs with $k_{\text{clip}}{=}1$. CLG and VLM-shaped methods
report augmented return, while PPO, CPO, PPOLag and PPOLag-RND report
environment return. Their $J_R$ values are therefore not comparable. $J_C$
uses a common scale, but does not establish safety because no method completes
the route and cost depends on movement. Symbols mark incomplete runs described
below.}
\label{tab:full-results}
\begin{minipage}[t]{0.48\linewidth}
\centering
\begin{tabular}{@{}llrrr@{}}
\toprule
\textbf{Method} & \textbf{Level} & \textbf{Seed} & $J_R$ & $J_C$ \\
\midrule
\multicolumn{5}{c}{\textit{Baselines (no VLM)}} \\
\midrule
PPO & L0 & 42  & 2.01 & 0.00 \\
PPO & L0 & 123 & 0.91 & 0.00 \\
PPO & L0 & 456 & 1.95 & 0.00 \\
PPO & L1 & 42  & 1.65 & 273.86 \\
PPO & L1 & 123 & 1.49 & 186.59 \\
PPO & L1 & 456 & 1.71 & 190.58 \\
PPO & L2 & 42  & 1.52 & 286.67 \\
PPO & L2 & 123 & 1.20 & 240.09 \\
PPO & L2 & 456 & 1.28 & 280.74 \\
\midrule
CPO & L0 & 42  & 1.42 & 0.00 \\
CPO & L0 & 123 & 1.42 & 0.00 \\
CPO & L0 & 456 & 2.78 & 0.00 \\
CPO & L1 & 42  & 0.20 & 28.42 \\
CPO & L1 & 123 & 0.38 & 54.23 \\
CPO & L1 & 456 & 0.38 & 24.35 \\
CPO & L2 & 42  & 0.28 & 30.58 \\
CPO & L2 & 123 & 0.72 & 58.38 \\
CPO & L2 & 456 & 0.05 & 19.28 \\
\midrule
PPOLag & L0 & 42  & 2.01 & 0.00 \\
PPOLag & L0 & 123 & 0.91 & 0.00 \\
PPOLag & L0 & 456 & 1.95 & 0.00 \\
PPOLag & L1 & 42  & 0.34 & 23.59 \\
PPOLag & L1 & 123 & 0.65 & 42.67 \\
PPOLag & L1 & 456 & 1.44 & 137.31 \\
PPOLag & L2 & 42  & 1.01 & 43.20 \\
PPOLag & L2 & 123 & 0.55 & 19.72 \\
PPOLag & L2 & 456 & 0.67 & 104.38 \\
\midrule
\multicolumn{5}{c}{\textit{CLG baselines}} \\
\midrule
PPO-CLG & L0 & 42  & 51.93 & 0.00 \\
PPO-CLG & L0 & 123 & 52.41 & 0.00 \\
PPO-CLG & L0 & 456 & 51.41 & 0.00 \\
PPO-CLG & L1 & 42  & 51.58 & 165.19 \\
PPO-CLG & L1 & 123 & 51.61 & 81.14 \\
PPO-CLG & L1 & 456 & 51.91 & 154.38 \\
PPO-CLG & L2 & 42  & 51.53 & 155.93 \\
PPO-CLG & L2 & 123 & 51.26 & 107.44 \\
PPO-CLG & L2 & 456 & 51.26 & 206.34 \\
\midrule
CPO-CLG & L0 & 42  & 51.50 & 0.00 \\
CPO-CLG & L0 & 123 & 52.12 & 0.00 \\
CPO-CLG & L0 & 456 & 51.34 & 0.00 \\
CPO-CLG & L1 & 42  & 50.34 & 40.75 \\
CPO-CLG & L1 & 123 & 51.18 & 36.42 \\
CPO-CLG & L1 & 456 & 50.65 & 21.12 \\
CPO-CLG & L2 & 42  & 50.73 & 38.05 \\
CPO-CLG & L2 & 123 & 50.91 & 25.63 \\
CPO-CLG & L2 & 456 & 51.00 & 38.02 \\
\bottomrule
\end{tabular}
\end{minipage}\hfill
\begin{minipage}[t]{0.48\linewidth}
\centering
\begin{tabular}{@{}llrrr@{}}
\toprule
\textbf{Method} & \textbf{Level} & \textbf{Seed} & $J_R$ & $J_C$ \\
\midrule
\multicolumn{5}{c}{\textit{CMDP $+$ VLM variants}} \\
\midrule
CPO-Coupled & L0 & 42  & 20.48 & 0.00 \\
CPO-Coupled & L0 & 123 & 21.29 & 0.00 \\
CPO-Coupled & L0 & 456 & 21.80 & 0.00 \\
CPO-Coupled & L1 & 42  & 21.43 & 23.45 \\
CPO-Coupled & L1 & 123$^{\dagger}$ & 20.05 & 33.20 \\
CPO-Coupled & L1 & 456 & 21.46 & 30.64 \\
CPO-Coupled & L2 & 42$^{\dagger}$  & 21.42 & 38.41 \\
CPO-Coupled & L2 & 123 & 21.91 & 30.42 \\
CPO-Coupled & L2 & 456$^{\dagger}$ & 21.53 & 28.25 \\
\midrule
CPO-Decoupled & L0 & 42  & 64.29 & 0.00 \\
CPO-Decoupled & L0 & 123 & 64.42 & 0.00 \\
CPO-Decoupled & L0 & 456 & 63.73 & 0.00 \\
CPO-Decoupled & L1 & 42  & 63.53 & 63.68 \\
CPO-Decoupled & L1 & 123 & 63.72 & 32.06 \\
CPO-Decoupled & L1 & 456 & 63.94 & 17.01 \\
CPO-Decoupled & L2 & 42$^{\dagger}$ & 64.11 & 38.68 \\
CPO-Decoupled & L2 & 123 & 63.82 & 38.61 \\
CPO-Decoupled & L2 & 456 & 63.70 & 15.37 \\
\midrule
PPOLag-Dec. & L0 & 42  & 64.49 & 0.00 \\
PPOLag-Dec. & L0 & 123 & 64.57 & 0.00 \\
PPOLag-Dec. & L0 & 456 & 63.73 & 0.00 \\
PPOLag-Dec. & L1 & 42  & 63.92 & 31.20 \\
PPOLag-Dec. & L1 & 123 & 64.06 & 33.94 \\
PPOLag-Dec. & L1 & 456 & 64.12 & 35.40 \\
PPOLag-Dec. & L2 & 42$^{\S}$ & 63.75 & 46.77 \\
PPOLag-Dec. & L2 & 123 & 64.13 & 32.37 \\
PPOLag-Dec. & L2 & 456 & 63.42 & 42.82 \\
\midrule
VLMPPOLag & L0 & 42  & 64.49 & 0.00 \\
VLMPPOLag & L0 & 123 & 64.57 & 0.00 \\
VLMPPOLag & L0 & 456 & 63.73 & 0.00 \\
VLMPPOLag & L1 & 42  & 63.85 & 44.38 \\
VLMPPOLag & L1 & 123 & 64.11 & 26.64 \\
VLMPPOLag & L1 & 456 & 64.30 & 27.29 \\
VLMPPOLag & L2 & 42  & 63.88 & 53.89 \\
VLMPPOLag & L2 & 123 & 64.03 & 23.37 \\
VLMPPOLag & L2 & 456 & 63.58 & 43.23 \\
\midrule
VLMPPOLag+Conf & L0 & 42  & 42.10 & 0.00 \\
VLMPPOLag+Conf & L0 & 123 & 50.78 & 0.00 \\
VLMPPOLag+Conf & L0 & 456 & 42.94 & 0.00 \\
VLMPPOLag+Conf & L1 & 42  & 50.19 & 15.64 \\
VLMPPOLag+Conf & L1 & 123 & 48.03 & 44.48 \\
VLMPPOLag+Conf & L1 & 456 & 47.91 & 22.79 \\
VLMPPOLag+Conf & L2 & 42  & 48.37 & 23.15 \\
VLMPPOLag+Conf & L2 & 123 & 50.74 & 44.21 \\
VLMPPOLag+Conf & L2 & 456 & 46.17 & 24.12 \\
\midrule
\multicolumn{5}{c}{\textit{Substitutions}} \\
\midrule
PPOLag-RND & L1 & 42  & 0.45 & 42.01 \\
PPOLag-RND & L1 & 123 & 0.47 & 58.15 \\
PPOLag-RND & L1 & 456 & 1.38 & 86.94 \\
PPOLag-RND & L2 & 42  & 0.17 & 43.56 \\
PPOLag-RND & L2 & 123 & 0.87 & 29.76 \\
PPOLag-RND & L2 & 456 & 0.25 & 63.71 \\
\midrule
Qwen2-VL+Conf & L2 & 42  & 8.03 & 24.90 \\
Qwen2-VL+Conf & L2 & 123 & 8.39 & 34.00 \\
Qwen2-VL+Conf & L2 & 456 & 8.31 & 77.67 \\
\bottomrule
\end{tabular}
\end{minipage}
\end{table}

$^{\dagger}$ These runs ended before $10^6$ steps. CPO-Coupled L1 seed $123$
completed two epochs. CPO-Coupled L2 seeds $42$ and $456$ completed $39$ and
$36$ epochs, while CPO-Decoupled L2 seed $42$ completed $25$. Their entries
use the final available epoch. None retained a trained checkpoint, so they are
excluded from policy evaluation.

$^{\S}$ The PPOLag-Decoupled L2 run at seed $42$ completed $39$ epochs but
retained only its initial policy. Its training entry uses the final logged
epoch, while evaluation summaries exclude the untrained policy. Including this
run would change the L2 training mean cost from $37.6$ over two trained runs to
$40.7$ over all three runs.

\section{RND and Qwen2-VL Substitutions}
\label{app:cr-extras}

We evaluate two representation substitutions on FormulaOne. PPOLag-RND
replaces the VLM danger score with Random Network Distillation novelty.
Qwen2-VL$+$Conf replaces CLIP with Qwen2-VL-7B. Neither experiment is a
matched ablation. At the final epoch, every RND run exceeds the cost budget
at both levels, while we detect no cost difference between Qwen2-VL and CLIP
on L2. Per-seed results appear in \cref{tab:full-results}.

\subsection{RND: the novelty cost stays small throughout training}
\label{app:rnd-details}

We replace $\cvlm$ with Random Network Distillation novelty
$\nu_t$~\cite{burda2018rnd}. A fixed, randomly initialised target network maps
the $44$-dimensional proprioceptive observation to a $32$-dimensional
representation. A predictor learns the same mapping from observations
normalised by their running mean and standard deviation. Both networks use
two hidden layers of $64$ ReLU units. The predictor uses Adam with a learning
rate of $10^{-4}$. The novelty cost is
\[
\nu_t = 1-\exp\!\left(-\max(z_t,0)\right),
\]
where $z_t$ is the squared prediction error divided by its running standard
deviation.

The environment cost remains the constrained quantity. Only the epoch mean
novelty $\bar\nu$ enters the multiplier update through
$\eta_2(\bar\nu-\tau)$. The RND runs use the FormulaOne multiplier settings
$\eta_1{=}0.035$, $\eta_2{=}0.01$, $\dlim{=}25$, and $\tau{=}0.5$. They omit
semantic reward shaping and the confidence gate. The comparison with
VLMPPOLag$+$Conf consequently changes three components and is not a matched
ablation.

The novelty signal remains small throughout training
(\cref{tab:rnd-novelty}). Across the six runs, $\bar\nu$ ranges from $0.007$
to $0.030$. By comparison, epoch-mean danger scores range from $0.607$ to
$0.629$ for decoupled CLIP on FormulaOne and from $0.577$ to $0.648$ for
Qwen2-VL. One possible explanation is rapid predictor convergence because
the proprioceptive observations have limited variation. We did not measure
the association between step-level novelty and environment cost, so this
interpretation remains untested.

The novelty term has little effect on the multiplier update. Its magnitude is
between $4.7{\times}10^{-3}$ and $4.9{\times}10^{-3}$, while the smallest
observed $|\Jc-\dlim|$ is $0.41$. It never changes the sign of the update in
any of the $50$ epochs. Because $\bar\nu<\tau$, it relaxes the effective cost
budget by at most $4.9{\times}10^{-3}$ cost units.

\begin{table}[h]
\centering
\small
\caption{Epoch-mean RND novelty cost $\bar\nu$ at four FormulaOne training
checkpoints. All six runs remain within $[0.0097,0.0247]$ at these checkpoints
and within $[0.009,0.029]$ after $100$k steps.}
\label{tab:rnd-novelty}
\begin{tabular}{@{}lccccc@{}}
\toprule
Level & Seed & $100$k & $240$k & $500$k & $1$M \\
\midrule
L1 & 42  & 0.0097 & 0.0107 & 0.0132 & 0.0101 \\
L1 & 123 & 0.0103 & 0.0133 & 0.0138 & 0.0127 \\
L1 & 456 & 0.0157 & 0.0199 & 0.0212 & 0.0224 \\
L2 & 42  & 0.0128 & 0.0121 & 0.0126 & 0.0106 \\
L2 & 123 & 0.0140 & 0.0126 & 0.0159 & 0.0176 \\
L2 & 456 & 0.0230 & 0.0247 & 0.0192 & 0.0157 \\
\bottomrule
\end{tabular}
\end{table}

\subsection{Qwen2-VL substitution: implementation and timing}
\label{app:qwen-details}

Qwen2-VL-7B replaces the CLIP scorer. For each frame $o_t$, the model answers
whether the image shows either ``safe driving conditions'' or ``driving danger
or imminent collision.'' The probability assigned to a yes answer is computed
from the logits of common yes and no token variants,
\begin{equation*}
P(\mathrm{yes}\mid o,d)
=
\sigma\!\left(
\underset{i\in\mathcal{Y}}{\mathrm{logsumexp}}\,\ell_i
-
\underset{j\in\mathcal{N}}{\mathrm{logsumexp}}\,\ell_j
\right),
\end{equation*}
where $d$ is the descriptor and $\mathcal{Y}$ and $\mathcal{N}$ contain the
single-token yes and no variants. We define $\rvlm$ using the safe descriptor
and $\cvlm$ using the danger descriptor. Each scoring call therefore requires
two model evaluations.

The confidence gate is
\[
\kappa_t
=
\left|
\frac{2\rvlm}{\rvlm+\cvlm}-1
\right|.
\]
This function has no learned scale or centre. The Qwen2-VL comparison
therefore changes both the visual-language backbone and the confidence
function.

The model is evaluated in half precision on one A100 GPU. The five runs with
timing measurements require $107.2$ to $113.0$\,ms per scoring call. All
Qwen2-VL runs query the model every eight control steps, giving an amortised
cost of about $14$\,ms per step. A queried step still exceeds the $40$\,ms
control period, so this measurement characterises training throughput and
does not establish real-time operation.

The epoch-mean danger score remains between $0.577$ and $0.648$ across all
three seeds. Final costs nevertheless range from $24.9$ to $77.7$. A stable
mean score does not establish calibration or explain this variation. The VLM
cost is not added to the environment cost, and its multiplier term never
changes the sign of the update in any of the $50$ epochs. Its direct influence
is therefore negligible. The Qwen2-VL signal primarily affects the policy
through the confidence-weighted reward bonus.

\begin{table}[h]
\centering
\small
\caption{Epoch-mean Qwen2-VL danger score at four FormulaOne-L2 training
checkpoints. Across all epochs, the three seeds remain within
$[0.577,0.648]$, compared with $[0.607,0.629]$ for decoupled CLIP scoring on
FormulaOne.}
\label{tab:qwen-vlmc}
\begin{tabular}{@{}lcccc@{}}
\toprule
Seed & $100$k & $240$k & $500$k & $1$M \\
\midrule
42  & 0.620 & 0.621 & 0.610 & 0.637 \\
123 & 0.610 & 0.619 & 0.601 & 0.616 \\
456 & 0.618 & 0.640 & 0.641 & 0.630 \\
\bottomrule
\end{tabular}
\end{table}

\clearpage
\subsection{Statistical tests for the substitution comparisons}
\label{app:cr-stats}

We compare each substitution with the CLIP$+$Conf arm on FormulaOne-L2. All
arms use seeds $\{42,123,456\}$, and the analysis uses the final epoch at
$10^6$ steps. We report a two-sided Welch test, a percentile-bootstrap $95\%$
confidence interval for the difference in means, and an exact one-sided
permutation test in the prespecified direction.

Qwen2-VL$+$Conf has a mean cost of $45.5$, compared with $30.5$ for
CLIP$+$Conf. The confidence interval contains zero, and neither test detects a
difference (\cref{tab:cr-stats}). This contrast does not isolate the backbone
because the arms also use different query periods and confidence functions.
PPOLag-RND has a mean cost of $45.7$. Its difference from CLIP$+$Conf is also
uncertain with three seeds. Comparisons with the five-seed calibrated family
depend on the aggregation window. The last-10-epoch comparison gives a
difference of $-24.2$ for CLIP$+$Conf minus PPOLag-RND, with Welch
$p{=}0.034$ and permutation $p{=}1/56$. At the final epoch, the difference is
$-23.3$, with Welch $p{=}0.13$ and permutation $p{=}2/56$
(\cref{tab:perm-l2}). Because RND also removes semantic reward shaping and the
confidence gate, neither comparison identifies the source of lower
FormulaOne-L2 cost.

The return comparisons do not measure task performance. Qwen2-VL$+$Conf and
CLIP$+$Conf report shaped return, while PPOLag-RND reports environment return.
The lower shaped return under Qwen2-VL is consistent with its smaller positive
score, whose epoch mean is $0.34$ compared with $0.63$ under CLIP. The gate is
not logged, so its contribution cannot be separated from this score.

\begin{table}[!t]
\centering
\small
\caption{Final-epoch comparisons on FormulaOne-L2 with three seeds per arm.
$\Delta$ is the difference between the two seed means. Confidence intervals use
$10^4$ percentile-bootstrap resamples. We also report two-sided Welch tests and
exact one-sided permutation tests in the prespecified direction. The smallest
attainable permutation $p$-value is $0.05$. $^{\dagger}$Returns are not directly
comparable because the Qwen2-VL and CLIP arms include the shaped reward, while
PPOLag-RND reports environment return.}
\label{tab:cr-stats}
\setlength{\tabcolsep}{4pt}
\resizebox{\textwidth}{!}{%
\begin{tabular}{@{}llccccccc@{}}
\toprule
Arm$_1$ & Arm$_2$ & Metric & mean$_1$ & mean$_2$ & $\Delta$ & Bootstrap $95\%$ CI on $\Delta$ & Welch $p$ & Perm.\ $p$ ($H_1$ dir.)\\
\midrule
Qwen2-VL$+$Conf & CLIP$+$Conf & $\Jc$                     & $45.52$ & $30.49$ & $+15.03$ & $[-9.58, +46.85]$  & $0.46$                & $0.80$ ($<$) \\
CLIP$+$Conf     & PPOLag-RND  & $\Jc$                     & $30.49$ & $45.68$ & $-15.18$ & $[-33.52, +3.15]$  & $0.28$                & $0.15$ ($<$) \\
\midrule
Qwen2-VL$+$Conf & CLIP$+$Conf & logged $\Jr^{\dagger}$    & $8.24$  & $48.43$ & $-40.19$ & $[-42.47, -37.96]$ & $1.0\!\times\!10^{-3}$ & $0.05$ ($<$) \\
CLIP$+$Conf     & PPOLag-RND  & logged $\Jr^{\dagger}$    & $48.43$ & $0.43$  & $+48.00$ & $[+45.92, +50.31]$ & $5.6\!\times\!10^{-4}$ & $0.05$ ($>$) \\
\bottomrule
\end{tabular}%
}

\end{table}

\section{Multiplier Learning-Rate Sweep and CLIP Backbone Size}
\label{app:extended-robustness}

We examine two sources of sensitivity. First, we vary the multiplier learning
rate and the coefficient of the VLM term. Second, we replace CLIP ViT-B/32
with ViT-L/14 while keeping the CLIP scoring rule fixed.

\subsection{Multiplier Learning-Rate Sweep}
\label{app:two-mult-ablation}

We use this sweep to test whether the VLM coefficient $\eta_2$ produces an
effect that could be explained by changing the multiplier learning rate
$\eta_1$. Both coefficients act on one multiplier. We did not evaluate
separate multipliers for environment cost and the VLM signal.

In the multiplier update, $\eta_1$ is the Adam learning rate and $\eta_2$
weights the VLM term inside
$(\Jc-\dlim)+\eta_2(\meancvlm-\tau)$. Across $8{,}572$ epochs from
$189$ training records with $\eta_2>0$, this term never reverses the sign of
a nonzero update. Removing it in a replay changes $\lambda$ by at most
$3.6{\times}10^{-4}$ (\S\ref{sec:results-eta2}). Within the sweep below, its
magnitude is at most $3.9{\times}10^{-3}$, while the smallest observed
$|\Jc-\dlim|$ is $0.02$.

We evaluate $\eta_1\in\{0.035,0.07\}$ and
$\eta_2\in\{0,0.01,0.03\}$ on FormulaOne-L2 using three seeds and
$10^6$ steps (\cref{tab:two-mult}). All VLM arms use decoupled CLIP scoring
without a confidence gate and set $\tau{=}0.5$. Query periods differ across
cells. The $\eta_1{=}0.035$ cells with $\eta_2{=}0$ and $0.01$ use $k{=}1$,
while the $\eta_2{=}0.03$ cell uses $k{=}4$. A matched $k{=}4$ control with
$\eta_2{=}0$ obtains $\Jr{=}63.8$ and $\Jc{=}27.5$. The
$\eta_1{=}0.07$, $\eta_2{=}0$ cell has no VLM and reports environment return.
All other cells report shaped return. We therefore do not compare return
across these reward definitions.

\begin{table}[h]
\centering
\caption{Multiplier sweep on FormulaOne-L2 with three seeds per cell. Each
cell reports final-epoch mean return and cost, followed by the number of seeds
over the cost budget. Bold marks a mean cost within the budget.
$^\ddagger$The VLM is queried every four steps. Unmarked cells in the first
row query every step. $^\dagger$This cell has no VLM. The cells are not matched
controls, so the table is interpreted only as a hyperparameter sweep.}
\label{tab:two-mult}
\small
\setlength{\tabcolsep}{6pt}
\begin{tabular}{@{}lccc@{}}
\toprule
 & $\eta_2{=}0$ & $\eta_2{=}0.01$ & $\eta_2{=}0.03$ \\
\midrule
$\eta_1{=}0.035$ (default)
  & $63.8\,/\,40.7$ (3/3)$^{\S}$
  & $63.8\,/\,40.2$ (2/3)
  & $63.8\,/\,41.4$ (2/3)$^\ddagger$ \\
$\eta_1{=}0.07$ (2$\times$ default)
  & $0.2^{\,\dagger}\,/\,33.0$ (3/3)
  & $63.9\,/\,\mathbf{20.1}$ (2/3)$^\ddagger$
  & $63.8\,/\,29.8$ (2/3)$^\ddagger$ \\
\bottomrule
\multicolumn{4}{p{0.92\linewidth}}{\footnotesize $^\dagger$~This VLM-free cell reports environment return, which is not comparable with the shaped returns in the other cells. $^{\S}$~One run stopped at $7.8{\times}10^{5}$ steps. Excluding it gives $\Jc{=}37.6$ across the remaining two seeds.}
\end{tabular}
\end{table}

The sweep does not show that increasing $\eta_2$ lowers cost. At
$\eta_1{=}0.035$, the two cells evaluated every step have nearly identical
costs of $40.7$ and $40.2$. The $\eta_2{=}0.03$ cell uses $k{=}4$ and cannot
be compared directly with them. Its matched $\eta_2{=}0$ control has a lower
cost of $27.5$ (\S\ref{sec:results-eta2}). At $\eta_1{=}0.07$, the
$\eta_2{=}0$ cell has no VLM reward, so its cost is not a controlled reference
for the other cells. Across the three $k{=}4$ cells with $\eta_2>0$, the VLM
term never changes the sign of the multiplier update.

At an interior stationary point, the effective budget is
$J_C^\star=\dlim+\eta_2(\tau-\meancvlm)$. The largest observed shift in this
sweep is $3.9{\times}10^{-3}$ cost units. Thus, the VLM term has a negligible
direct effect on the multiplier, while the unmatched grid prevents a causal
comparison of final costs.

The twelve additional runs require $242$ GPU-hours on single A100 GPUs.
Individual runs take $19.0$ to $21.2$ hours, with a median of $20.4$ hours.
Restarted runs add $73$ GPU-hours.

\subsection{CLIP Backbone Size}
\label{app:clip-capacity}

Qwen2-VL changes both the model and scoring rule. To examine CLIP backbone
size more directly, we replace ViT-B/32 with ViT-L/14 while retaining
decoupled cosine scoring. The frozen ViT-B/32 and ViT-L/14 models contain
$151$ and $428$ million parameters, respectively.

We begin with the calibrated FormulaOne-L2 VLMPPOLag$+$Conf configuration.
For ViT-L/14, we recompute the text features and recalibrate the confidence
gate using $5000$ random-policy frames. ViT-B/32 uses the original
$500$-frame calibration buffer. All remaining settings are fixed, including
$\lambda_r{=}0.1$, $\lambda_c{=}0.5$, $\eta_2{=}0.01$, $\tau{=}0.5$, and
$k{=}4$. ViT-L/14 uses seeds $\{42,123,456\}$, while the ViT-B/32 reference
uses the five calibrated family-A seeds.

\begin{table}[h]
\centering
\caption{CLIP backbone comparison on FormulaOne-L2. Values are last-10-epoch
means $\pm$ population standard deviations. Both frozen backbones use
decoupled scoring and a separately calibrated confidence gate. Returns include
the shaped reward. Viol. counts seeds whose mean cost exceeds $\dlim{=}25$.
The ViT-B/32 and ViT-L/14 rows use five and three seeds, respectively.}
\label{tab:clip-capacity}
\small
\setlength{\tabcolsep}{6pt}
\begin{tabular}{@{}lcccc@{}}
\toprule
\textbf{Backbone} & \textbf{Params} & $\Jr$ & $\Jc$ & \textbf{Viol.} \\
\midrule
CLIP ViT-B/32 (calibrated, family~A) & 151\,M & $31.8\pm 12.2$ & $\textcolor{blue}{22.5}\pm 5.9$ & 1/5 \\
CLIP ViT-L/14 & 428\,M & $16.1\pm 1.7$ & $26.9\pm 5.7$ & 1/3 \\
\bottomrule
\multicolumn{5}{p{0.8\linewidth}}{\footnotesize One ViT-B/32 scoring call takes
$7.11$\,ms on an A100. ViT-L/14 latency was not benchmarked.}
\end{tabular}
\end{table}

ViT-L/14 has a last-10-epoch mean cost of $26.9$, which lies within the
bootstrap $95\%$ confidence interval $[17.9,27.9]$ for the ViT-B/32 mean.
This overlap does not establish equivalent safety performance, especially
because FormulaOne cost tracks distance travelled (\S\ref{app:task-metrics}).

The shaped return falls from $31.8$ with ViT-B/32 to $16.1$ with ViT-L/14.
This difference does not measure task performance. The mean positive score is
similar across the two backbones at $0.635$ and $0.617$, so the lower return
may reflect a smaller confidence-weighted bonus under ViT-L/14. We cannot test
this explanation because the gate value was not recorded. The fitted scale
ranges from $75$ to $176$ for ViT-L/14 and from $109$ to $562$ for ViT-B/32.
The corresponding centres range from $0.006$ to $0.010$ and from $0.014$ to
$0.029$. The backbone comparison therefore shows sensitivity in the shaped
return but does not isolate its source.

\subsection{Limitations}
\label{app:limitations}

Our conclusions are limited in five ways.
\begin{enumerate}[leftmargin=20pt,itemsep=1pt,topsep=2pt]
\item MetaDrive evaluation uses scenarios drawn from the training pool, so it
does not measure transfer to unseen scenarios (\S\ref{sec:setup}).
\item The reduction appears only on MetaDrive Hard, which combines the densest
traffic with the largest map. We detect no benefit on Medium or Bullet, while
performance worsens on Easy (\cref{tab:generalisation}).
\item FormulaOne-L2 remains unsolved, and its cost largely tracks distance
travelled. We therefore use this environment only for mechanistic analyses
(\S\ref{sec:results-main}).
\item Reward shaping is not isolated. The comparison with PPOLag changes the
bonus, confidence gate, and multiplier term together. RND and Qwen2-VL are also
unmatched substitutions. Most comparisons use three to five seeds, and an
exact test with three seeds per group cannot attain $p<0.05$
(App.~\ref{app:perm-tests}).
\item Prompts are written separately for each environment. The environment and
VLM signals also share one multiplier~\cite{altman1999constrained}. Learning
prompts from observed costs~\cite{ma2024eureka} and assigning a separate
multiplier to the VLM signal remain open directions.
\end{enumerate}

\end{document}